\documentclass[11pt]{article}
\usepackage[
  margin=1.5cm,
  includefoot,
  footskip=30pt,
]{geometry}
\counterwithin{figure}{section}
\usepackage{tikz}
\usetikzlibrary{matrix, arrows}
\usepackage{amsmath,amssymb}
\let\checkmark\checkmarka
\usepackage{dingbat}
\usepackage{amsthm}
\usepackage{mathtools}
\usepackage{xspace}
\usepackage[noend]{algorithmic}
\usepackage[ruled,vlined]{algorithm2e}
\usepackage{url}
\usepackage{makeidx}
\usepackage{enumerate}
\usepackage{epstopdf}
\usepackage{booktabs}
\usepackage{color}
\usepackage[utf8]{inputenc}
\usepackage{thm-restate}
\usepackage{scalerel,stackengine}
\usepackage[shortlabels]{enumitem}
\usepackage{xr}
\usepackage{fancyvrb}
\usepackage{xcolor}
\usepackage{bold-extra}
\usepackage[width=474.18663pt]{caption}
\usepackage{subfigure}
\usepackage[most]{tcolorbox}
\usepackage{fvextra}
\usepackage[frozencache=true, finalizecache=false, cachedir=./minted-cache]{minted} 
\usepackage{float}
\usepackage{alltt}
\usepackage{soul}
\usepackage{fancyvrb}
\usepackage{multirow}
\usepackage[final]{hyperref}
\usepackage[bottom]{footmisc}
\usepackage{natbib}
\usepackage{linguex}
\usepackage{ulem}
\usepackage[T1]{fontenc}
\usepackage[titles,subfigure]{tocloft}

\usepackage[all]{nowidow}

\usepackage{listings}
\usepackage{tikz}
\usetikzlibrary{shapes,calc,positioning}

\tcbset{
  aibox/.style={
    top=10pt,
    left=5pt,
    colback=white,
    colframe=black,
    colbacktitle=black,
    enhanced,
    center,
    attach boxed title to top left={yshift=-0.1in,xshift=0.15in},
    boxed title style={boxrule=0pt,colframe=white,},
  }
}
\newtcolorbox{AIbox}[3][]{aibox,title=#2,#1,width=#3}

\definecolor{aigold}{RGB}{244,210, 1} 
\definecolor{aigreen}{RGB}{210,244,211} 
\definecolor{highlightpurple}{RGB}{0,112,192} 
\definecolor{checkgreen}{RGB}{0,194,6}

\sethlcolor{aigreen}

\definecolor{aired}{RGB}{255,180,181}

\newtcbox{\mybox}[1][green]{on line,
arc=0pt,outer arc=0pt,colback=#1!10!white,colframe=#1!50!black,
boxsep=0pt,left=0pt,right=0pt,top=0pt,bottom=0pt,
boxrule=0pt,bottomrule=0pt,toprule=0pt}

\usepackage{xurl}

\usepackage{scalerel}
\def\tortoise{\includegraphics[height=0.8em]{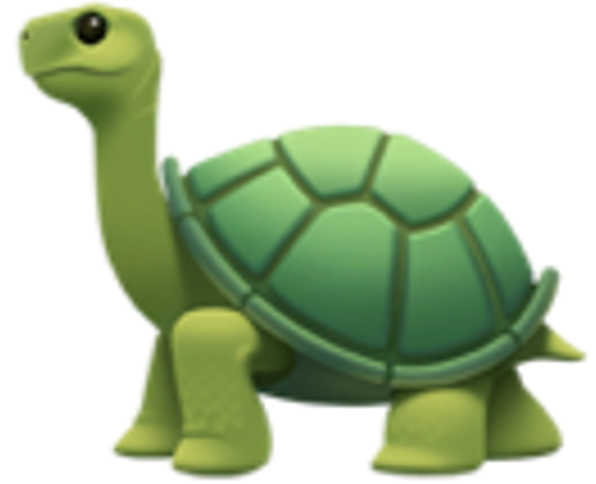}{}}
\def\redx{\includegraphics[height=0.8em]{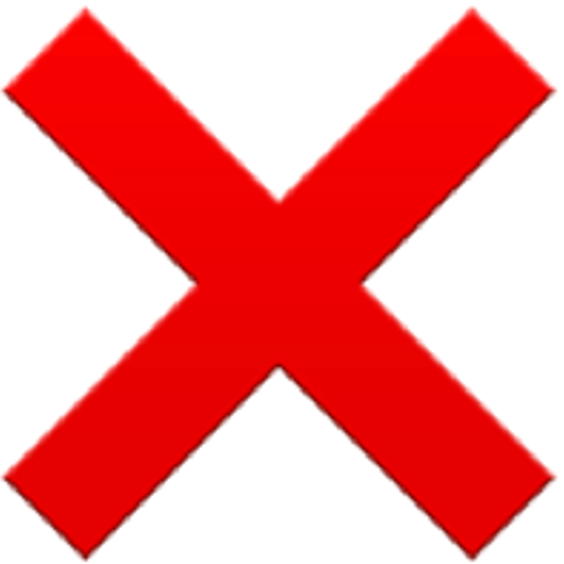}{}}

\usepackage{array}
\newcolumntype{L}[1]{>{\raggedright\let\newline\\\arraybackslash\hspace{0pt}}m{#1}}
\newcolumntype{C}[1]{>{\centering\let\newline\\\arraybackslash\hspace{0pt}}m{#1}}
\newcolumntype{R}[1]{>{\raggedleft\let\newline\\\arraybackslash\hspace{0pt}}m{#1}}

\begin{document}

\newcommand{\DV}{GPT-4\xspace}

\title{
\textbf{Embers of Autoregression: Understanding Large Language Models Through the Problem They are Trained to Solve}}

\author{R. Thomas McCoy
\and Shunyu Yao
\and Dan Friedman
\and Matthew Hardy
\and Thomas L. Griffiths
}

\date{Princeton University}

\maketitle

\renewcommand{\abstractname}{}  
\abstract{
\noindent
\textbf{One-sentence summary:} 

\noindent
To understand what language models are, we must understand what we have trained them to be.

\bigskip
\noindent
\textbf{Abstract:}

\noindent
The widespread adoption of large language models (LLMs) makes it important to recognize their strengths and limitations. We argue that in order to develop a holistic understanding of these systems we need to consider the problem that they were trained to solve: next-word prediction over Internet text.
By recognizing the pressures that this task exerts we can make predictions about the strategies that LLMs will adopt, allowing us to reason about when they will succeed or fail.
This approach---which we call the teleological approach---leads us to identify three factors that we hypothesize will influence LLM accuracy:
the probability of the task to be performed, the probability of the target output, and the probability of the provided input.
We predict that LLMs will achieve higher accuracy when these probabilities are high than when they are low---even in deterministic settings where probability should not matter.
To test our predictions, we evaluate two LLMs (GPT-3.5 and GPT-4) on eleven tasks, and we find robust evidence that LLMs are influenced by probability in the ways that we have hypothesized. 
In many cases, the experiments reveal surprising failure modes. For instance, GPT-4's accuracy at decoding a simple cipher is 51\% when the output is a high-probability word sequence but only 13\% when it is low-probability.
These results show that AI practitioners should be careful about using LLMs in low-probability situations. 
More broadly, we conclude that we should not evaluate LLMs as if they are humans but should instead treat them as a distinct type of system---one that has been shaped by its own particular set of pressures.
}

\tableofcontents

\setlength{\Exlabelwidth}{0.7em}
\setlength{\SubExleftmargin}{1.35em}
\renewcommand{\firstrefdash}{} 

\section{Introduction}
    
Large language models (LLMs), such as GPT-4 \citep{openai2023gpt4}, PaLM \citep{chowdhery2022palm}, and Llama~2 \citep{touvron2023llama}, receive a piece of text as input and then generate additional text as output.
Virtually any task can be framed in the form of linguistic queries, so LLMs could be applied to virtually any task---from summarizing text to generating computer code. 
This flexibility is exciting: it led one recent paper to argue that LLMs display ``sparks of artificial general intelligence'' \citep{bubeck2023sparks}. However, it also hinders us from understanding LLMs holistically. Given that we can only run a finite number of tests, how can we understand a system whose potential scope is infinite? Answering this question requires some method for deciding which tests will be most informative about the general strengths and weaknesses of LLMs.

\begin{figure}[b!]
    \centering
    \includegraphics[width=0.7\textwidth]{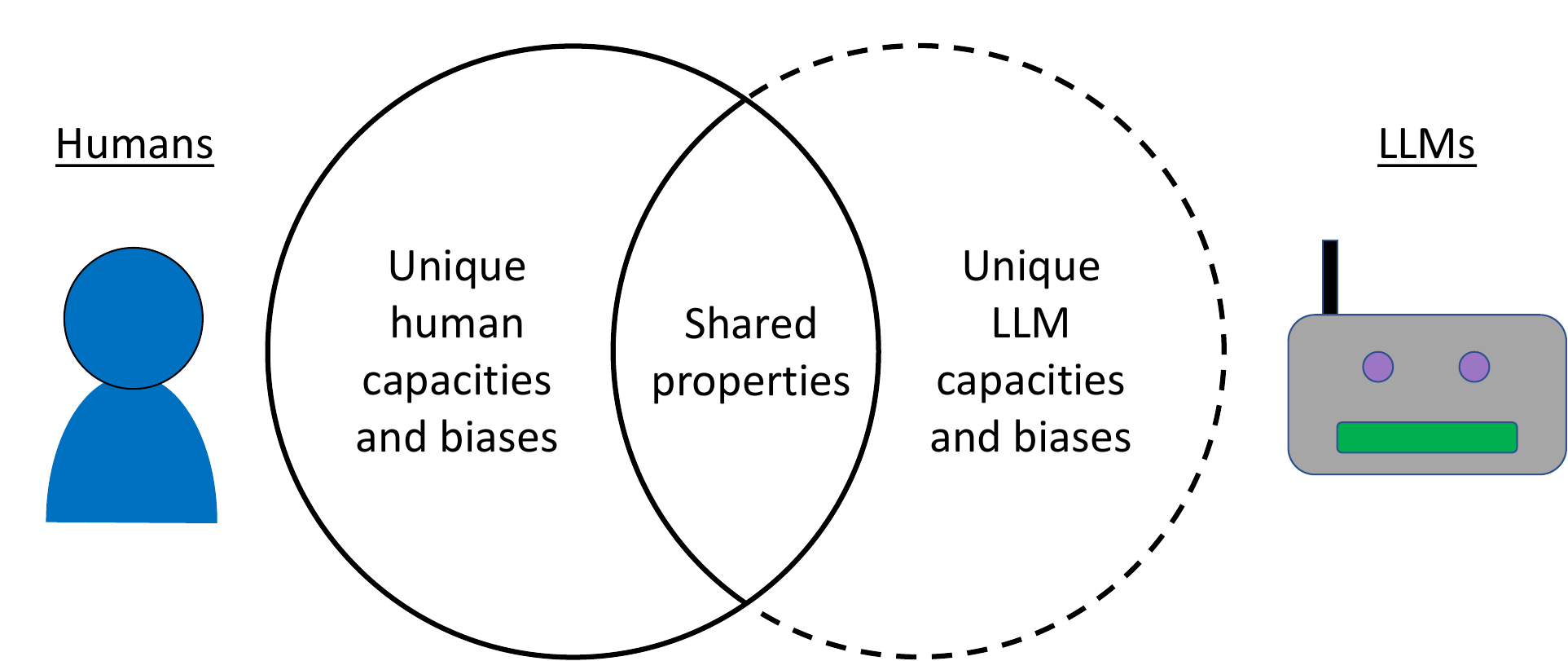}
    \caption{Humans and large language models (LLMs) have some shared properties and some properties that differ. If LLMs are analyzed using tests designed for humans, we risk identifying only the shared properties, missing the properties that are unique to LLMs (the dotted region of the diagram). We argue that in order to identify the properties in the dotted region we must approach LLMs on their own terms by considering the problem that they were trained to solve: next-word prediction over Internet text.}
    \label{fig:venn}
\end{figure}

One popular way to select evaluations is to use a human-centric approach: test for the abilities that are viewed as most important for characterizing human cognition.
For instance, some popular benchmarks target the ability to draw inferences from sentences \citep{dagan2005pascal,williams2018broad}, which is widely viewed as a central aspect of human reasoning.
In some cases, LLM evaluations are not only inspired by general human abilities but in fact use specific tests that were originally designed for humans: \citet{openai2023gpt4} evaluated GPT-4 on the SAT, the GRE, and other real-world exams, and many popular benchmarks such as ARC \citep{clark2018think} and MMLU \citep{hendrycks2020measuring} are curated from tests written for humans.
However, LLMs are not humans, so the types of tests that might be most informative about them may differ from the tests that are most informative about humans \citep{mitchell2023we}. In particular, a human-centric approach runs the risk of highlighting the strengths of these models---their overlap with human abilities---without revealing their idiosyncratic weaknesses (Figure~\ref{fig:venn}). How can we approach the problem of understanding a \textit{new} type of intelligence, evaluating it on its own terms?

We argue for an approach that has been extremely productive in cognitive science: understanding intelligent systems by understanding the sorts of problems that they developed to solve \citep{marr1982vision,shepard1987toward,anderson1990adaptive,griffiths2020understanding}. 
This approach---which we refer to as the \textbf{teleological approach}---focuses on the goals of the system and the environment in which it operates.
As an example from biology, some species of birds store seeds in the ground and then retrieve them months later. 
Finding stored seeds is much easier if one can remember where they are hidden, so we would expect birds that rely on seed caching to have evolved strong spatial memory abilities. Indeed, \citet{balda1989comparative} found that Clark's nutcrackers---which store thousands of seeds each year---perform better on spatial memory tasks than scrub jays---which, though they are related to Clark's nutcrackers, are much less reliant on seed caching. 
That is, the memory differences between these species can be explained by differences in the problems that they need to solve.

The crucial question to ask, then, is: What problem(s) do LLMs need to solve, and how do these pressures influence them? Here we focus on perhaps the most salient pressure that defines any machine learning system, namely the task that it was trained to perform. For the LLMs that have been the focus of recent attention in AI, this task is autoregression---next-word prediction \citep{elman1991distributed,bengio2000neural,radford2018improving}---performed over Internet text. 
As described in more detail in the next section, we argue that a full understanding of LLMs should be strongly informed by the fact that this is their primary training objective. We therefore hypothesize that, even when they are asked to perform tasks that seem very different from next-word prediction---e.g., solving math problems---their performance on such tasks is highly influenced by the autoregressive substrate from which all of their abilities emerge. 
That is, if we continue with the example of solving math problems, it is important to view the LLM not as a ``math problem solver'' but rather as a ``statistical next-word prediction system being used to solve math problems.''

\begin{figure}[t!]
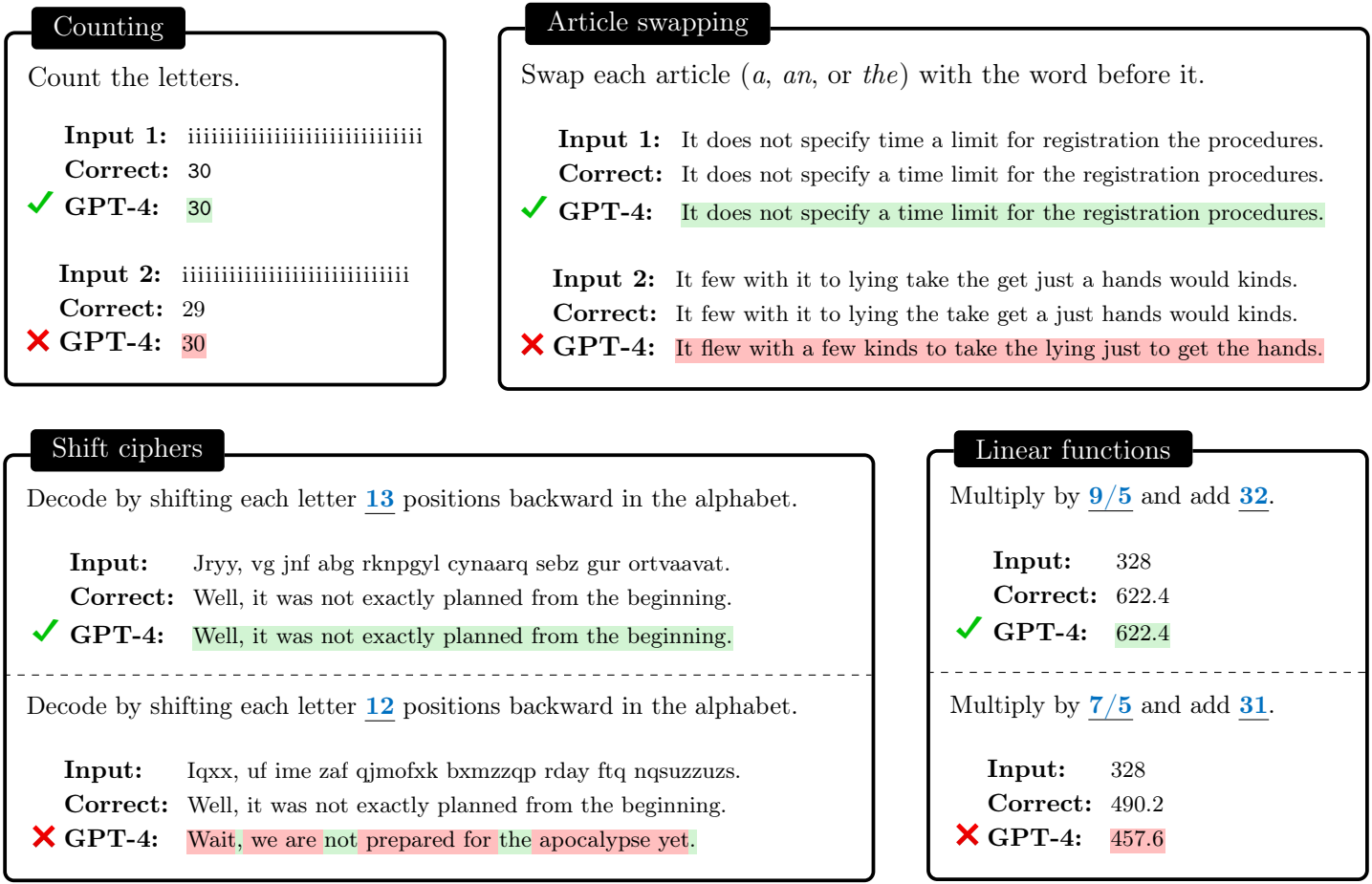


    \centering

    \begin{minipage}[l]{0.31\textwidth}
    \centering
    \begin{AIbox}{Counting}{175pt}
    \parbox[t]{\linewidth}{
    Count the letters.

    \bigskip

    \sodef\count{}{0.06em}{0.2em plus0.2em}{0.5em plus.1em minus.1em}
    \begin{tabular}{@{}c@{~}l@{~~}p{15.5cm}}
        & {\small \textbf{Input 1:}} & {\footnotesize \count{iiiiiiiiiiiiiiiiiiiiiiiiiiiiii}} \\
        & {\small \textbf{Correct:}} & \sethlcolor{white}{\footnotesize \texttt{\hl{30}}}\\
        \textcolor{checkgreen}{\textbf{\checkmark}} & {\small \textbf{GPT-4:}} & \sethlcolor{aigreen}{\footnotesize \texttt{\hl{30}}}
    \end{tabular}

    \bigskip

    \begin{tabular}{@{}c@{~}l@{~~}p{5cm}}
        & {\small \textbf{Input 2:}} & {\footnotesize \count{iiiiiiiiiiiiiiiiiiiiiiiiiiiii}} \\
        & {\small \textbf{Correct:}} & \sethlcolor{white}{\footnotesize {\hl{29}}}\\
        \redx{} & \textbf{GPT-4:} & \sethlcolor{aigreen}{\footnotesize {\sethlcolor{pink}\hl{30}}}
    \end{tabular}
    }
    \end{AIbox}
    \end{minipage}
    \hfill
    \begin{minipage}[l]{0.63\textwidth}
    \centering
    \begin{AIbox}{Article swapping}{345pt}
    \parbox[t]{\linewidth}{
    Swap each article (\textit{a}, \textit{an}, or \textit{the}) with the word before it.

    \bigskip

    \begin{tabular}{@{}c@{~}l@{~~}p{10.8cm}}
        & {\small \textbf{Input 1:}} & \sethlcolor{white}{\footnotesize {\hl{It does not specify time a limit for registration the procedures.}}} \\
        & {\small \textbf{Correct:}} & \sethlcolor{white}{\footnotesize {\hl{It does not specify a time limit for the registration procedures.}}}\\
        \textcolor{checkgreen}{\textbf{\checkmark}} & {\small \textbf{GPT-4:}} & \sethlcolor{aigreen}{\footnotesize {\hl{It does not specify a time limit for the registration procedures.}}}
    \end{tabular}

    \bigskip

    \begin{tabular}{@{}c@{~}l@{~~}p{10.8cm}}
        & {\small \textbf{Input 2:}} & \sethlcolor{white}{\footnotesize {\hl{It few with it to lying take the get just a hands would kinds.}}} \\
        & {\small \textbf{Correct:}} & \sethlcolor{white}{\footnotesize {\hl{It few with it to lying the take get a just hands would kinds.}}}\\
        \redx{} & \textbf{GPT-4:} & \sethlcolor{aigreen}{\footnotesize {\sethlcolor{pink}\hl{It flew with a few kinds to take the lying just to get the hands.}}}
    \end{tabular}
    }
    \end{AIbox}
    \end{minipage}

    \vspace{5mm}

    \begin{minipage}[l]{0.63\textwidth}
    \centering
    \begin{AIbox}{Shift ciphers}{345pt}
    \parbox[t]{\linewidth}{
    {
    {\small Decode by shifting each letter \uline{\textcolor{highlightpurple}{\textbf{13}}} positions backward in the alphabet.}
    }

    \bigskip
    
    \hspace*{-0.0cm}
    \begin{tabular}{@{}c@{~}l@{~~}p{9.8cm}}
        & {\small \textbf{Input:}} & \sethlcolor{white}{\footnotesize {\hl{Jryy, vg jnf abg rknpgyl cynaarq sebz gur ortvaavat.}}} \\
        & {\small \textbf{Correct:}} & \sethlcolor{white}{\footnotesize {\hl{Well, it was not exactly planned from the beginning.}}}\\
        \textcolor{checkgreen}{\textbf{\checkmark}} & {\small \textbf{GPT-4:}} & \sethlcolor{aigreen}{\footnotesize {\hl{Well, it was not exactly planned from the beginning.}}}
    \end{tabular}
    
    \tcbline

    {\small 
    Decode by shifting each letter \uline{\textcolor{highlightpurple}{\textbf{12}}} positions backward in the alphabet.}

    \bigskip

    \hspace*{-0.0cm}
    \begin{tabular}{@{}c@{~}l@{~~}p{9.8cm}}
        & {\small \textbf{Input:}} & \sethlcolor{white}{\footnotesize {\hl{Iqxx, uf ime zaf qjmofxk bxmzzqp rday ftq nqsuzzuzs.}}} \\
        & {\small \textbf{Correct:}} & \sethlcolor{white}{\footnotesize {\hl{Well, it was not exactly planned from the beginning.}}}\\
        \redx{} & {\small \textbf{GPT-4:}} & \sethlcolor{aigreen}\sethlcolor{pink}{\footnotesize {\hl{Wait}\sethlcolor{aigreen}\hl{,}\sethlcolor{pink}\hl{ we are  }\sethlcolor{aigreen}\hl{not}\sethlcolor{pink}\hl{ prepared for }\sethlcolor{aigreen}\hl{the}\sethlcolor{pink}\hl{ apocalypse yet}\sethlcolor{aigreen}\hl{.}}}
    \end{tabular}
    }

    \end{AIbox}
    \end{minipage}
    \hfill
    \begin{minipage}[l]{0.31\textwidth}
    \centering
    \begin{AIbox}{Linear functions}{175pt}
    \parbox[t]{\linewidth}{
    {
    {\small Multiply by \uline{\textcolor{highlightpurple}{\textbf{9/5}}} and add \uline{\textcolor{highlightpurple}{\textbf{32}}}.}
    }

    \bigskip
    
    \hspace*{-0.0cm}
    \begin{tabular}{@{}c@{~}l@{~~}p{9.8cm}}
        & {\small \textbf{Input:}} & \sethlcolor{white}{\footnotesize {\hl{328}}} \\
        & {\small \textbf{Correct:}} & \sethlcolor{white}{\footnotesize {\hl{622.4}}}\\
        \textcolor{checkgreen}{\textbf{\checkmark}} & {\small \textbf{GPT-4:}} & \sethlcolor{aigreen}{\footnotesize {\hl{622.4}}}
    \end{tabular}
    
    \tcbline

    {\small 
    Multiply by \uline{\textcolor{highlightpurple}{\textbf{7/5}}} and add \uline{\textcolor{highlightpurple}{\textbf{31}}}.}

    \bigskip

    \hspace*{-0.0cm}
    \begin{tabular}{@{}c@{~}l@{~~}p{9.8cm}}
        & {\small \textbf{Input:}} & \sethlcolor{white}{\footnotesize {\hl{328}}} \\
        & {\small \textbf{Correct:}} & \sethlcolor{white}{\footnotesize {\hl{490.2}}}\\
        \redx{} & {\small \textbf{GPT-4:}} & \sethlcolor{pink}{\footnotesize {\hl{457.6}}}
    \end{tabular}
    }
    \end{AIbox}
    \end{minipage}

    \caption{GPT-4 struggles on some seemingly simple tasks such as counting, article swapping, shift ciphers, and linear functions. Later in the paper, we explain the contrasts that are illustrated here. In the counting and article swapping examples, GPT-4 fails in the cases where the correct output is a low-probability piece of text (for the counting example, we refer to 29 as low-probability because it occurs much less frequently in natural corpora than 30 does). In the shift cipher and linear function examples, GPT-4 performs well on the task variants that are common in Internet text but poorly on the variants that are rare (note that the shift cipher with a shift of 13 is over 100 times more common in Internet text than the shift cipher with a shift of 12; and the linear function $f(x) = (9/5)x + 32$ is common because it is the Celsius-to-Fahrenheit conversion, while the other linear function has no special significance). For the sake of brevity, this figure does not show the full prompts we used; see later in the paper for the complete prompts. The GPT-4 predictions were obtained using \texttt{gpt-4-0613} on the OpenAI API; other model versions (e.g., the online chat interface) may give different predictions.}
    \label{fig:failure_examples}
    \end{figure}

A mismatch between the problem that a system developed to solve and the task that it is being given can have significant consequences. Just as the human drive to obtain sweet and fatty foods can be maladaptive in a world where those foods are easily available, the autoregressive tendencies of LLMs can cause problems when they are given a task that is not next-word prediction. In this paper, we exhibit what might seem like surprising failures of LLMs on tasks that are straightforward for humans to perform, such as using simple ciphers, constructing acronyms, calculating linear functions, and counting; see Figure~\ref{fig:failure_examples}. 
These failures can be understood directly in terms of a conflict between next-word prediction and the task at hand. Being able to reason about when  LLMs will fail at a task is critical as these systems become more widely deployed.

Based on an analysis of the problem that LLMs are trained to solve (statistical next-word prediction), we make three predictions about how LLMs will be influenced by their origin in this task---the \textbf{embers of autoregression} that appear in these systems even as they might show sparks of artificial general intelligence. 
These predictions are listed in Table~\ref{tab:listofembers}. For example, we predict that, even when the task is a deterministic one that does not depend on probability, LLMs will achieve higher accuracy when the answer is high-probability than when it is low-probability. 
In the next section, we discuss these hypotheses in more detail. 

We then test these hypotheses through an extensive set of experiments. 
Since our goal is to identify sources of difficulty for LLMs, our experiments need to take place in settings where LLMs make at least some errors. 
To create such settings, we use an adversarial strategy that follows straightforwardly from our hypotheses: we use tasks that push LLMs into low-probability situations where we expect they will perform poorly. For instance, one task we use is decoding simple ciphers, since text written in a cipher is a low-probability subspace of the broader space of possible types of text. Table~\ref{tab:tasks_all_examples} on page \pageref{tab:tasks_all_examples} gives the full list of tasks that we use.
As long as models indeed have a non-negligible error rate on these tasks, we can then investigate which factors cause this error rate to increase or decrease---focusing on the factors that we have hypothesized will affect the error rate.

Using this strategy, we find robust evidence for the effects described in Table~\ref{tab:listofembers}: as task probability or example probability varies, LLM accuracy can indeed vary substantially in the ways that we have hypothesized.
Beyond the three effects that we focus on, we also use some additional experiments and a literature review to discuss how the teleological approach provides a unifying perspective on a variety of effects that have been discussed in prior work. Combining these additional effects with the three that we focus on yields the following complete list:

\ex. \textbf{Embers of autoregression:} \label{ex:emberslist}
    \a. Sensitivity to task frequency
    \b. Sensitivity to output probability
    \c. Sensitivity to input probability
    \d. Lack of embodiment
    \d. Sensitivity to wording
    \d. Difficulty on meaning-dependent tasks
    \d. Inability to modify text that has already been produced
    \d. Societal biases and spurious correlations
    \d. Idiosyncratic memorization
    \d. Sensitivity to tokenization
    \d. Limited compositionality and systematicity

\noindent
Overall, our findings illustrate how we can understand LLMs more clearly if we recognize the pressures that have shaped them. Our code, our stimuli, and the responses we gathered in our experiments are all on GitHub.\footnote{\url{https://github.com/tommccoy1/embers-of-autoregression}}

\begin{table*}[]
    \centering
    \resizebox{\textwidth}{!}{
    \begin{tabular}{p{2.5cm}p{8cm}p{8cm}} \toprule
        \raggedright Ember of autoregression & Definition & Example \\ \midrule
        \raggedright 
        \textbf{Sensitivity to task frequency} & LLMs perform better on tasks that are frequent than ones that are rare, even when the tasks have an equivalent level of complexity. &  When asked to translate English sentences into Pig Latin, GPT-4 gets 42\% accuracy when using the most common variant of Pig Latin but only 23\% accuracy when using a rare variant. \vspace{3mm}\\ 
        \raggedright 
        \textbf{Sensitivity to output probability} & LLMs achieve higher accuracy when the correct answer is high-probability text than when it is low-probability text, even when the task is deterministic. & When asked to reverse a sequence of words, GPT-4 gets 97\% accuracy when the answer is a high-probability sentence yet 53\% accuracy when the output is low probability. \vspace{3mm}\\
        \raggedright \textbf{Sensitivity to input probability} & Even when the task is deterministic, LLMs sometimes achieve higher accuracy when the input text is high-probability than when it is low-probability, but input probability is less influential than output probability. & When asked to encode sentences in a simple cipher (rot-13), GPT-4 gets 21\% accuracy when the input is a high-probability sentence yet 11\% accuracy when the input is low probability. \\ 
        \bottomrule
    \end{tabular}
    }
    \caption{Effects on the performance of large language models that are attributable to the fact that they are statistical next-word prediction systems.} 
    \label{tab:listofembers}
\end{table*}

\section{A teleological approach to understanding LLMs}

To understand an information-processing system such as an LLM, the approach that we are arguing for is to first characterize the problem that the system solves and to then use this characterization as a source of hypotheses about the system's capacities and biases. We refer to this as the teleological approach because it focuses on explaining the behavior of the system in terms of its goal (\textit{telos} in Greek) \citep[e.g.,][]{lombrozo2006functional}. Teleological explanation is a common strategy in making sense of biological systems, manifest in approaches such as computational-level \citep{marr1982vision} and rational \citep{anderson1990adaptive} analysis in cognitive science, and adaptationist explanations in evolutionary biology \citep{godfrey2001three,shettleworth2010cognition}. Just as teleological explanations can be incomplete or misleading in these settings \citep{gould1979spandrels,lombrozo2019mechanistic} we do not anticipate that all the properties of LLMs can be understood via their goals. However, we believe that this is a useful lens through which to study these systems and generate predictions about their behavior.
In Section~\ref{sec:llmproblem} we describe more clearly the goals that LLMs are trained to accomplish. In Section \ref{sec:hypotheses}, we then use this analysis to generate hypotheses about LLM behavior. Finally, Section \ref{sec:notarguing} clarifies what arguments we are and are not intending to make in pursuing this approach.

\subsection{What problem do LLMs solve?}\label{sec:llmproblem}

\citet[][p.~29]{anderson1990adaptive} argued that the problem solved by a cognitive system can be characterized by three factors: the system's goal, the environment in which the system pursues this goal, and the computational limitations that constrain the system. 
In the context of machine learning, those three factors correspond to the following aspects of a model:

\begin{enumerate}
    \item \textbf{Training task:} What goal is the system trained to accomplish?
    \item \textbf{Training distribution:} What types of examples is the system trained on?
    \item \textbf{Model architecture:} What computational tools does the system have access to?
\end{enumerate}
For standard LLMs, the training task is next-word prediction, the training distribution is a distribution over Internet text,\footnote{We use \textit{Internet text} as an umbrella term covering the full range of diverse document types that appear in LLM pre-training datasets. Such documents include the types of data that are traditionally referred to as text (e.g., news articles), as well as other types of data such as computer code.} and the model architecture is a neural network (specifically, a Transformer: \citealp{vaswani2017attention}).
Thus, the problem that an LLM must solve is: \textit{how to perform next-word prediction on samples of Internet text, given the mechanisms available in a neural network}.

Note that many recent LLMs are not solely trained on next-word prediction but also go through
a second training phase based on instruction tuning \citep{ouyang2022training}, which is intended to align model behavior with human preferences. 
A complete teleological analysis should therefore consider instruction tuning as well as next-word prediction. 
However, in this paper, we only analyze next-word prediction, leaving instruction tuning for future work.
We start with next-word prediction because,
during LLM training, the next-word prediction phase is much longer than the instruction tuning phase. 
We therefore conjecture that next-word prediction plays a greater role in shaping the basic nature of LLMs than instruction tuning does, 
making analysis of next-word prediction a reasonable starting point for the goal of understanding LLMs.

Importantly, the tasks for which LLMs are used often differ from the problem they were trained to solve. For example, even though they were trained for next-word prediction, LLMs are sometimes asked to translate sentences or develop Python code. Why does this discrepancy matter? When a system is adapted for one purpose but then co-opted for a different purpose, the original purpose may influence the system's nature in ways that would not make sense if only the new purpose were considered. As an example from biology, the basic mammalian body plan evolved for a life on four legs, but humans recently evolved to instead walk on two legs. 
Biologists have argued that the friction between our quadrupedal roots and our bipedal lifestyle contributes to several musculoskeletal ailments that are common in humans \citep{latimer2005perils,held2009quirks}. For instance, having our spines be perpendicular to the ground rather than parallel with it causes strains that make humans susceptible to lower back pain \citep{krogman1951scars,plomp2015ancestral}.
This bipedalism example can be viewed as friction arising from a shift in the body's function or ``task'' (four-legged life vs.\ two-legged life); mismatches can also arise with the other two factors of environment (training distribution) and architecture. An example of a shift in environment is the dietary mismatch highlighted in the introduction: humans developed a drive towards consuming fatty foods because we evolved in a world where such foods were scarce and their calories were valuable, but this taste can now be harmful in a world with readily-available donuts and ice cream \citep{buss1998adaptations}. Finally, a difficulty arising from our ``architecture'' is the blind spot in human vision: due to the way in which the optic nerve exits the retina, there is a point in the retina without photoreceptors, making us unable to see a specific region in our visual field \citep{wandell1995foundations}. We anticipate that goals, training distribution, and model architecture create opportunities for similar mismatches in machines.

\subsection{Hypothesized embers of autoregression}\label{sec:hypotheses}

Do LLMs face anything analogous to humans' back pain, love for fatty foods, and blind spots---any quirks attributable to a mismatch between what they ``adapted'' to do and what they are used for? To answer this question, we must consider the types of solutions that would arise from the pressures that shape LLMs. First, the fact that LLMs are neural networks makes them statistical systems. The other two factors---the task of next-word prediction and the training distribution of Internet text---dictate \textit{which} statistics they are sensitive to: the statistics of word sequences in Internet text. Building on this analysis, we make the following three predictions about LLM performance (some additional predictions are listed in \ref{ex:emberslist} on page \pageref{ex:emberslist}):
\begin{itemize}
    \item Even when there is no difference in the complexity of the tasks, LLMs will perform worse on rare tasks than on frequent tasks. 
    \item Even when asked to perform a deterministic task, LLMs will be sensitive to the probability of the answers they must produce.
    \item Even when asked to perform a deterministic task, LLMs will be sensitive to the probability of the inputs they are given, but to a lesser degree than they are sensitive to the probability of the  answers they must produce.
\end{itemize}
A crucial aspect of these predictions is the first clause of each one. For instance, it is clear that probability is important for 
next-word prediction, but we are predicting that probability will influence LLMs even in tasks that are not inherently probabilistic, such as forming an acronym from a sequence of words. 
In later sections, we will develop these hypotheses in more detail and will provide extensive evidence supporting them.

\subsection{What we are not arguing}
\label{sec:notarguing}

We are arguing that certain types of tasks and examples will be harder for LLMs than others. We are not arguing that LLMs are incapable of handling these tasks and examples. That is, our core predictions are of the form ``for LLMs, X is harder than Y'' rather than ``LLMs can't do X.''

This distinction is important for explaining how we set up our experiments. For most experiments, we test LLMs using basic prompting (i.e., simply asking them to provide an answer to a query), even though LLM accuracy can be substantially increased by using more sophisticated inference techniques such as step-by-step prompting \citep{nye2021show,kojima2022large}, chain-of-thought prompting \citep{wei2022chain}, or the tree of thoughts framework \citep{yao2023tree}.
Our decision to use basic prompting would be problematic if we were claiming that LLMs cannot perform our tasks: such claims would only be true if LLMs fail under all prompting approaches, so investigations of such claims should use the prompting approach that yields the strongest performance.
However, since this is not the type of claim we are making, it is not necessary to use the strongest available prompting paradigms. Instead, because our claims are comparative in nature (``for LLMs, X is harder than Y''), what is important is ensuring that the two conditions (X and Y) are evaluated in the same way as each other. Since any prompting approach could work for this purpose, we chose the approach that is the fastest and most straightforward to run, namely basic prompting---though see Section~\ref{sec:promptvariations} for experiments with other prompting techniques.

Though the choice of prompting technique does not matter for our purposes, the fact that LLMs are so sensitive to the prompting approach supports our broader claim that, in order to understand LLMs, we must consider their autoregressive roots. That is, because LLMs are trained to make probabilistic predictions about word sequences, we would expect them to be highly sensitive to the particular word sequences that they interact with, both regarding the wording of the prompt (a point elaborated on in Section~\ref{sec:wording}) and regarding the types of word sequences that they are encouraged to produce (the factor that is targeted by several prominent prompting techniques such as chain-of-thought prompting: \citealt{wei2022chain}). 
The fact that prompting techniques indeed influence performance substantially can therefore be viewed as another ember of autoregression.

In addition, we are not claiming that researchers have ignored the autoregressive origins of LLMs. Indeed, in the ``Sparks of artificial general intelligence'' paper \citep{bubeck2023sparks}, Section 8 is titled ``Limitations of autoregressive architecture highlighted by GPT-4,'' and it considers failures of planning in arithmetic and text generation that result from only predicting the next word in a sequence. What we are claiming is that this aspect of LLMs has been neglected in constructing effective evaluations of their capacities. As highlighted in Figure~\ref{fig:venn}, much of the literature evaluating LLMs has started with tasks that are viewed as important indicators of human abilities
and then assessed how well LLMs can do them. 
By instead starting with tasks that we anticipate will be challenging for systems that are focused on next-word prediction, we get a more balanced view of what kinds of tasks are easy or hard for these systems. 

\section{Motivating our predictions}

In Section~\ref{sec:hypotheses}, we listed several hypotheses about factors that cause difficulty for LLMs. 
Here we show how we arrived at these hypotheses by analyzing the problem that LLMs need to solve.
To make this discussion more concrete, we will provide brief examples from one particular domain, namely shift ciphers (defined below).
Later in the paper, we will expand upon the shift cipher results (augmenting our initial observations with statistical tests and further analyses) and will also use an additional ten tasks to further test our predictions.

\subsection{Running example: Shift ciphers}\label{sec:describeshift}

In a shift cipher, a message is encoded by shifting each letter forward in the alphabet a certain number of positions. For example, with a shift of 1, \textit{How are you?}\ becomes \textit{Ipx bsf zpv?}\ because the letter after \textit{h} is \textit{i}, the letter after \textit{o} is \textit{p}, etc.
One particularly prominent case is \textbf{rot-13} (short for ``rotate by 13''), the shift cipher with a shift of 13 positions. Rot-13 is commonly used in online forums as a spoiler-free way to share information. For example, in some puzzle-solving groups, community members write hints in rot-13 so that those who want to solve the puzzle without help will not read the hint inadvertently.

    \begin{figure}

    \centering

    \begin{minipage}[l]{0.64\textwidth}
    \centering
    \begin{AIbox}{Shift cipher: Task probability}{300pt}
    \parbox[t]{\linewidth}{
    {
    {\small {\bf Common task: Rot-13.} 
    Decode the message by shifting each letter \underline{\textcolor{highlightpurple}{\textbf{thirteen}}} positions backward in the alphabet.}
    }

    \bigskip
    
    \hspace*{-0.0cm}
    \begin{tabular}{@{}c@{~}l@{~~}p{9.8cm}}
        & {\small \textbf{Input:}} & \sethlcolor{white}{\footnotesize \texttt{\hl{Jryy, vs gurl qba'g pbzr, fb or vg.}}} \\
        & {\small \textbf{Correct:}} & \sethlcolor{white}{\footnotesize \texttt{\hl{Well, if they don't come, so be it.}}}\\
        \textcolor{checkgreen}{\textbf{\checkmark}} & {\small \textbf{GPT-4:}} & \sethlcolor{aigreen}{\footnotesize \texttt{\hl{Well, if they don't come, so be it.}}}
    \end{tabular}
    
    \tcbline

    {\small {\bf Uncommon task: Rot-2.} 
    Decode the message by shifting each letter \underline{\textcolor{highlightpurple}{\textbf{two}}} positions backward in the alphabet.}

    \bigskip

    \hspace*{-0.0cm}
    \begin{tabular}{@{}c@{~}l@{~~}p{9.8cm}}
        & {\small \textbf{Input:}} & \sethlcolor{white}{\footnotesize \texttt{\hl{Ygnn, kh vjga fqp'v eqog, uq dg kv.}}} \\
        & {\small \textbf{Correct:}} & \sethlcolor{white}{\footnotesize \texttt{\hl{Well, if they don't come, so be it.}}}\\
        \redx{} & {\small \textbf{GPT-4:}} & \sethlcolor{aigreen}{\footnotesize \texttt{\hl{Well, if }\sethlcolor{pink}\hl{there isn't cake}\sethlcolor{aigreen}\hl{, }\sethlcolor{pink}\hl{to}\sethlcolor{aigreen}\hl{ be it.}}}
    \end{tabular}
    }
    \end{AIbox}
    \end{minipage}
    \begin{minipage}[r]{0.35\textwidth}
    \centering
    \vspace{0.3cm}
    
    \includegraphics[scale=0.55]{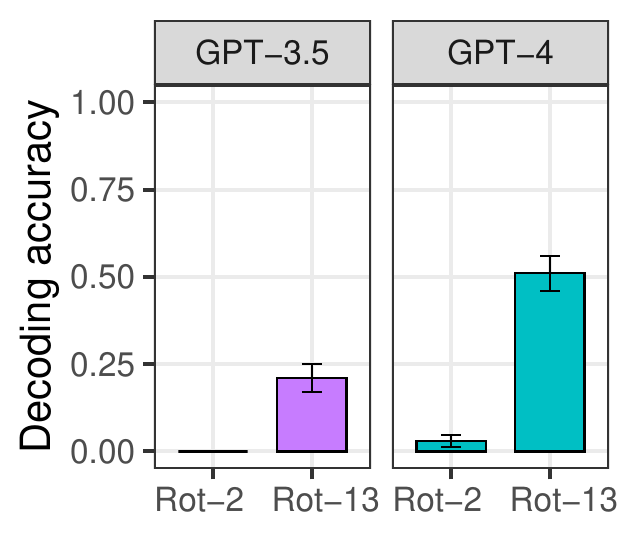}

    \textit{Note:} In Internet text, rot-13 is about 60 times more common than rot-2.
    \end{minipage}

    \vspace{0.5cm}
    \begin{minipage}[l]{0.64\textwidth}
    \centering
    \begin{AIbox}{Shift cipher: Output probability}{300pt}
    \parbox[t]{\linewidth}{

    {\small \underline{\textbf{Rot-13 decoding: Example of high-probability output}}}
    
    \begin{tabular}{@{}c@{~}l@{~~}p{9.8cm}}
        & {\small \textbf{Input:}} & \sethlcolor{white}{\footnotesize \texttt{\hl{Svefg, fur whfg cbfgrq gb ure Vafgntenz fgbel.}}} \\
        & {\small \textbf{Correct:}} & \sethlcolor{white}{\footnotesize \texttt{\hl{First, she just posted to her Instagram story.}}}\\
        \textcolor{checkgreen}{\textbf{\checkmark}} & {\small \textbf{GPT-4:}} & \sethlcolor{aigreen}{\footnotesize \texttt{\hl{First, she just posted to her Instagram story.}}}
    \end{tabular}

    \bigskip

    {\small \underline{\textbf{Rot-13 decoding: example of medium-probability output}}}

    \begin{tabular}{@{}c@{~}l@{~~}p{9.8cm}}
        & {\small \textbf{Input:}} & \sethlcolor{white}{\footnotesize \texttt{\hl{Fbeel, Naanguba jevgrf gb bhe Pbclevtug Hfref.}}} \\
        & {\small \textbf{Correct:}} & \sethlcolor{white}{\footnotesize \texttt{\hl{Sorry, Annathon writes to our Copyright Users.}}}\\
        \redx{} & \textbf{GPT-4:} & \sethlcolor{aigreen}{\footnotesize \texttt{\hl{Sorry,    }\sethlcolor{pink}\hl{Annabeth}\sethlcolor{aigreen}\hl{ writes to our }\sethlcolor{pink}\hl{Prophetic}\sethlcolor{aigreen}\hl{ Users.}}}
    \end{tabular}
    }
    \end{AIbox}
    \end{minipage}
    \begin{minipage}[r]{0.35\textwidth}
    \centering
    \vspace{0.7cm}
    
    \includegraphics[scale=0.55]{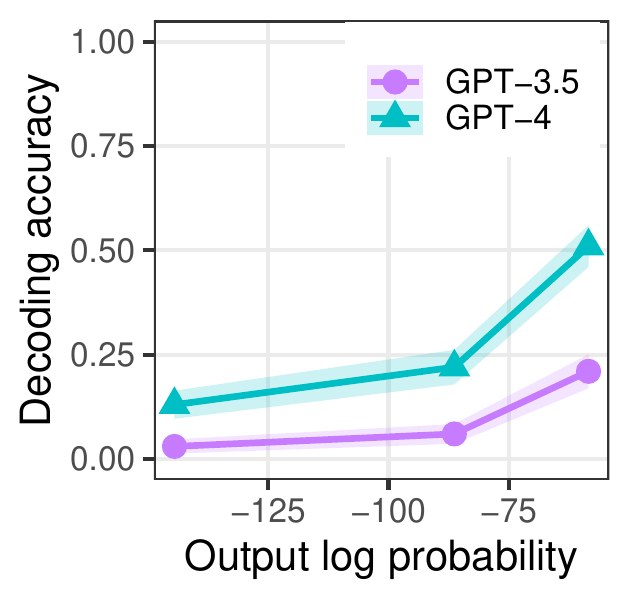}
    \end{minipage}

    \vspace{0.3cm}
    \begin{minipage}[l]{0.64\textwidth}
    \centering
    \begin{AIbox}{Shift cipher: Input probability}{300pt}
    \parbox[t]{\linewidth}{

    {\small \underline{\textbf{Rot-13 encoding: example of high-probability input}}}

    \begin{tabular}{@{}c@{~}l@{~~}p{9.8cm}}
        & {\footnotesize \textbf{Input:}} & \sethlcolor{white}{\footnotesize \hl{\texttt{In a word, everything has been complicated there.}}} \\
        & {\footnotesize \textbf{Correct:}} & \sethlcolor{white}{\footnotesize \hl{\texttt{Va n jbeq, rirelguvat unf orra pbzcyvpngrq gurer.}}}\\
        \textcolor{checkgreen}{\textbf{\checkmark}} & \textbf{GPT-4:} & \sethlcolor{aigreen}{\footnotesize \hl{\texttt{Va n jbeq, rirelguvat unf orra pbzcyvpngrq gurer.}}}
    \end{tabular}

    \bigskip

    {\small \underline{\textbf{Rot-13 encoding: example of medium-probability input}}}

    \begin{tabular}{@{}c@{~}l@{~~}p{9.8cm}}
        & {\small \textbf{Input:}} & \sethlcolor{white}{\footnotesize \hl{\texttt{In a word, governance has been frustrating daily.}}} \\
        & {\small \textbf{Correct:}} & \sethlcolor{white}{\footnotesize 
        \hl{\texttt{Va n jbeq, tbireanapr unf orra sehfgengvat qnvyl.}}}\\
        \redx{} & \textbf{GPT-4:} & \sethlcolor{aigreen}{\footnotesize \hl{\texttt{Va n jbeq,}    }\sethlcolor{pink}\hl{\texttt{tbinapr}}\sethlcolor{aigreen}\hl{\texttt{ unf orra }}\sethlcolor{pink}\hl{\texttt{sehfgevat qnlyl}}\sethlcolor{aigreen}\hl{\texttt{.}}}
    \end{tabular}
    }
    \end{AIbox}
    \end{minipage}
    \begin{minipage}[r]{0.35\textwidth}
    \centering
    \vspace{0.7cm}
    
    \includegraphics[scale=0.55]{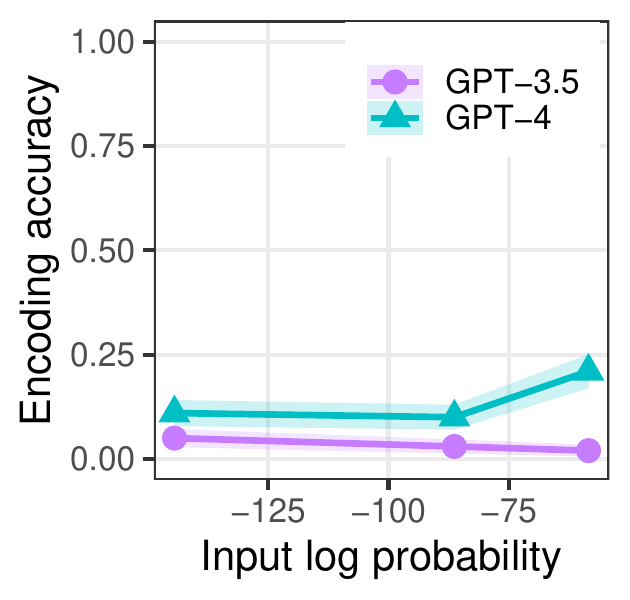}
    \end{minipage}

    \caption{When processing shift ciphers, LLMs are highly sensitive to probability, even though shift ciphers are deterministic. \textbf{Top:} Effect of task probability. GPT-3.5 and GPT-4 both score better on a frequently-occurring shift cipher (rot-13) than on a rare one (rot-2). \textbf{Middle:} Effect of output probability. When decoding text written in the rot-13 cipher, both models score better when the answer is high-probability than when it is low-probability. \textbf{Bottom:} Effect of input probability. As predicted, input probability does not show as strong an effect as output probability. All error bars and intervals show one standard error.}
    \label{fig:initial_shift_results}
    \end{figure}

\subsection{Sensitivity to task probability}\label{sec:rot13taskprob} 

Task probability is the probability that a task will be illustrated in a randomly-selected sample of text.
This probability determines how many examples of the task will have been encountered by an LLM during training. 
Neural networks generally perform better on a task when they have been trained on more examples of the task, so we expect that LLMs will perform better on high-probability tasks than low-probability tasks.

This prediction is borne out for shift ciphers. As mentioned above, rot-13 is a commonly-used shift cipher. In contrast, rot-2---which uses a shift of 2 rather than 13---is no more complex than rot-13 but is used rarely. We asked GPT-3.5 and GPT-4 to decode messages written in each cipher. Both models performed much better on the frequent task of rot-13 than on the rare task of rot-2 (Figure~\ref{fig:initial_shift_results}, top), supporting the conclusion that LLMs are sensitive to task probability.

\subsection{Sensitivity to output probability}\label{sec:outputprobmotivation}

LLMs are statistical systems. We therefore expect that their predictions will be influenced by probability, even in deterministic situations where probability should be irrelevant.
Specifically, we predict that LLMs will be biased toward producing high-probability sequences of words, meaning that their performance will get worse when the correct output is in fact low-probability.

To motivate this point more formally, we can first frame the LLM's task as finding the output that maximizes $P(\text{output} | \text{input})$, where the input is the start of a word sequence, and the output is the sequence's continuation. By Bayes' rule, this problem is equivalent\footnote{We do not claim that LLMs actually compute a likelihood and prior. Rather, the problem that they must solve can be framed in these terms, such that we expect their behavior to display phenomena that would arise in a system which performed this decomposition, even though the LLM might arrive at this behavior in a way that does not explicitly involve such a decomposition. That is, we are analyzing LLMs at \citeauthor{marr1982vision}'s (\citeyear{marr1982vision}) computational level.} to maximizing $P(\text{input} | \text{output}) P(\text{output})$.
Therefore, as long as there are multiple outputs for which $P(\text{input} | \text{output})$ is nonzero, the LLM's predictions will be influenced by the unconditional probability of the output, $P(\text{output})$: among the candidates that yield a nonzero value for $P(\text{input} | \text{output})$, LLMs will be biased toward selecting ones with a high $P(\text{output})$.

The argument made so far only covers non-deterministic situations, when there are multiple outputs that yield a non-zero $P(\text{input} | \text{output})$. Now consider deterministic settings, where there is only one output that could go with a given input; rot-13 is one such setting. 
In principle, $P(\text{output})$ should not matter in such cases because the likelihood, $P(\text{input} | \text{output})$, fully determines the output on its own. In practice, however, LLMs are unlikely to have perfectly estimated the distribution $P(\text{input} | \text{output})$, meaning that there will in fact be multiple candidate outputs that yield a nonzero likelihood. With multiple output candidates in consideration, $P(\text{output})$ will then affect the model's predictions in practice, even if in theory it should be irrelevant.

The results, in Figure~\ref{fig:initial_shift_results} (middle), support this prediction. We evaluated GPT-3.5 and GPT-4 on decoding text written in rot-13 across three conditions that varied in the probability of the output sentence (see Section~\ref{sec:sentenceprobs} for details on how output probability was modulated). For both models, accuracy increased as the log probability of the correct output was increased, supporting the hypothesis that these systems are influenced by the unconditional probability of potential outputs.

\subsection{Sensitivity to input probability}

The previous section described one way in which probability is relevant to LLMs---namely, the fact that LLMs use probability when performing inference. Probability also has a second, and less direct, effect on LLMs because of how it influences LLM training data. The probability of a piece of text influences how often that text will occur in the training set, so a trained LLM will have had less experience with low-probability strings than high-probability strings. 
Therefore, the model may capture information about low-probability strings less robustly than information about high-probability strings. 
As a concrete example, when the model has encountered shift ciphers during training, it is more likely to have seen the enciphered versions of common words such as \textit{time} than rare words such as \textit{emit}, and it may therefore be more successful at encoding or decoding \textit{time} than \textit{emit}. 

Information learned during pretraining is important both for processing the input and producing the output. We have already hypothesized that LLMs will be sensitive to output probability (Section~\ref{sec:outputprobmotivation}); this new argument adds the additional hypothesis that they will be sensitive to input probability. 
Figure~\ref{fig:initial_shift_results} (bottom) shows that, when encoding text into rot-13, GPT-4 achieves higher accuracy when the input is high-probability than when it is low-probability, illustrating that LLMs can indeed be sensitive to input probability. GPT-3.5 shows no significant effect of input probability. In the next subsection, we describe why LLMs may show less sensitivity to input probability than output probability.

\subsection{Asymmetry between input probability and output probability}\label{sec:asymmetrymotivation}

We have hypothesized that LLMs will be sensitive to the probability of both the input and the target output. 
Though these effects sound similar, we have predicted them for two fairly different reasons. Input probability was predicted to matter only in situations where the task depends on information about the input that models mainly learn through experience with that specific input; thus, for rarer inputs, models will perform the task less well due to a lack of experience with the relevant inputs. We expect that such situations only arise occasionally, such that a model's dependence on input probability may not manifest itself routinely.
In contrast, output probability was predicted to matter not only in the same situations where input probability matters but also in an additional context: whenever the model has any uncertainty about what the correct output is (in which case the model will use probability to help resolve its uncertainty). 
We expect that this situation is common since neural network outputs rarely include probabilities that are precisely equal to zero. We therefore expect that output probability will be more broadly influential than input probability.
An initial, intuitive piece of evidence for this hypothesis is the fact that, in Figure~\ref{fig:initial_shift_results}, output probability substantially affects model performance, whereas input probability shows only a minor effect and only in one model. We will expand on this idea more rigorously below.

\section{Overview of experiments}

In the next several sections, we test the hypotheses developed in the previous section by analyzing LLM performance on a wider range of tasks. Here we briefly describe the tasks, sentences, and models that we investigate. In the next sections, we will then discuss in more detail the particular tasks that are relevant to each hypothesis and how we used these tasks to test each hypothesis. 

\subsection{Tasks}\label{sec:tasks}

We evaluated models on the tasks described in Table~\ref{tab:tasks_all_examples}. See Appendix~\ref{app:samplesize} for the sample size used with each task.
We chose these tasks for two reasons. First, as described in the introduction, we selected tasks that push models into low-probability situations so that models will have a non-negligible error rate---a requirement for our goal of observing what causes the error rate to increase or decrease.
Second, most of the tasks that we chose are ones that can be solved with a simple, deterministic algorithm that is invariant to various changes to the task. For example, consider a shift cipher that moves each letter $n$ positions forward in the alphabet. The most straightforward way to decode this cipher is to shift each letter back $n$ positions---an algorithm that works equally well regardless of the value of $n$ and the identity of the particular words being processed. 
Thus, if a person demonstrated that they could decipher one message written in rot-13, we might assume that they knew this algorithm and could therefore perform equally well on any other shift cipher or on any other inputs. 
However, we have hypothesized that LLMs are not invariant to such properties: we expect their performance to vary based on which task variant is used and which inputs are used. The tasks described in the table allow us to test these predictions.

Many of our tasks involve character-level manipulations of words. At first blush, this factor might seem unfair to the GPT models because they operate over subword tokens rather than characters. 
To address this concern, we gave both models a spelling test where we asked them to spell 1,000 single-token words; details are in Appendix~\ref{app:spelling}. Both models scored close to 100\% (99.8\% for GPT-3.5 and 99.9\% for GPT-4). Thus, these models robustly encode the spelling of their tokens, making it reasonable to test them on tasks that involve character-level manipulations. For additional evidence that LLMs encode the character-level structure of tokens, see \citet{itzhak2022models} and \citet{kaushal2022tokens} (the latter of which also contains discussion of how LLMs might come to learn this information).

\begin{table}[th!]
    \centering
\renewcommand{\arraystretch}{1.7}
\begin{tabular}{p{3.3cm}p{6cm}p{6.3cm}} \toprule
    Task & Description & Example \\ \midrule
    \textbf{Article swapping}  & 
    Swap each article (\textit{a}, \textit{an}, or \textit{the}) with the preceding word. & {\small \texttt{In box the there was key a.}} \newline $\rightarrow$ {\small \texttt{In the box there was a key.}} 
    \\
    \textbf{Reversal}  & 
    Reverse a sequence of words. &
    {\small\texttt{everyone!\ morning Good,}} \newline $\rightarrow$ {\small \texttt{Good morning, everyone!}} 
    \\ 
    \textbf{Counting}   & 
    Count the words or letters in a list.  & {\small \texttt{lively news exhibit steep}} \newline $\rightarrow$ {\small \texttt{4}} 
    \\
    \textbf{Acronyms} &
    Join the first letters of the words in a list. & {\small \texttt{view inch show into tray east deal}} \newline $\rightarrow$ {\small \texttt{VISITED}}
    \\
    \textbf{Linear function} 
    & 
    Apply $f(x) = (9/5)x + 32$.
    & {\small \texttt{328}} \newline $\rightarrow$ {\small \texttt{622.4}} 
    \\
    \textbf{Multiplication}  & 
    Multiply two three-digit numbers. & {\small \texttt{351 times 373}} \newline $\rightarrow$ {\small \texttt{130923}}
    \\
    \textbf{Sorting}  & 
    Sort a list of words in alphabetical or reverse alphabetical order. & {\small \texttt{immense, tariff, golden, magic}} \newline $\rightarrow$ {\small \texttt{golden, immense, magic, tariff}} \\
    \textbf{Keyboard cipher}  & 
    Replace each letter with the one to the right of it on a keyboard. & {\small \texttt{Hello world!}} \newline $\rightarrow$ {\small \texttt{Jraap eptaf!}}
    \\
    \textbf{Shift cipher} 
     & 
    Decode by shifting each letter 13 positions backward in the alphabet.
     & 
     {\small \texttt{Fgnl urer!}} \newline $\rightarrow$ {\small \texttt{Stay here!}}
      \\
    \textbf{Pig Latin} & 
        Move the first consonant cluster of each word to the end and add \textit{-ay}. &  {\small \texttt{the frogs aren't noisy.}} \newline $\rightarrow$ {\small \texttt{ethay ogsfray aren'tay oisynay.}}  \\
    \textbf{Birthdays} & 
        Return the birth date of a provided public figure. &  {\small \texttt{Jeremy Lin}} \newline $\rightarrow$ {\small \texttt{August 23, 1988}}  \\
    \bottomrule
\end{tabular}

    \caption{The tasks that we used to evaluate LLMs.
    To keep this table manageably sized, some of the examples are not from the datasets that we used to evaluate LLMs but are instead shorter examples of the tasks that those datasets target.}
    \label{tab:tasks_all_examples}
\end{table}

\subsection{Sentences}\label{sec:sentenceprobs}

Two of our hypotheses relate to the probability of the specific examples a model is processing: we have predicted that models will perform worse on low-probability sentences than high-probability ones.
To facilitate testing of these hypotheses, we generated three sets of sentences---a high-probability set, a medium-probability set, and a low-probability set. The sentences in \ref{ex:probexamples} below show one example from each set. 
The high-probability set consisted of 100 sentences drawn from GlobalVoices, an online news service that we chose because its content is published under a permissive license that allows sharing and modification.\footnote{\url{https://globalvoices.org/about/global-voices-attribution-policy/}} The medium-probability set was created by taking the high-probability sentences and using RoBERTa \citep{liu2019roberta} to replace some of the words with others that had a low probability in that context (but were still grammatical).
Each replacement word was required to be the same length as the word it was replacing, whether length was measured in letters or subword tokens. 
The low-probability set was created by shuffling the words of the medium-probability examples, except that the first and last words were left in place. For more details on the creation of these sentence sets, see Appendix~\ref{app:rot13_sentences}.

\ex. \label{ex:probexamples}
\a. \textbf{High-probability:} Are they now building a bridge of their own?\label{ex:highprob}
\b. \textbf{Medium-probability:} Are dogs yet climbing a jungle of their own?\label{ex:mediumprob}
\c. \textbf{Low-probability:} Are their jungle of dogs a yet climbing own?\label{ex:lowprob}

\subsection{Models}\label{sec:models}

The models that we investigated were GPT-3.5 and GPT-4 \citep{openai2023gpt4}. We chose to include GPT-4 because it is the state-of-the-art in many areas and because it is the focus of the ``sparks of AGI'' paper \citep{bubeck2023sparks} that our work connects to. As a second model, we used GPT-3.5 for two reasons: first, it allows us to investigate the effects of model scale by comparing to GPT-4; second, it uses the same tokenizer as GPT-4, allowing us to evaluate it on the same examples as GPT-4 (our datasets control for tokenization, so if we were to evaluate a model that uses a different tokenizer we would need to generate new datasets). Both of these systems are large-scale Transformer \citep{vaswani2017attention} language models. We accessed them through the OpenAI API and had them generate responses with a temperature of 0.0. We used the most recent time-stamped versions that were available at the time we ran the experiments: \texttt{gpt-3.5-turbo-0613} and \texttt{gpt-4-0613}.

We emphasize that the results we found may not transfer identically to other language models, or even to other versions of the ones that we have tested, such as newer versions or versions accessed in a different way (i.e., via the chat interface rather than via the API). In fact, anecdotally, we noticed that that the chat interface for GPT-4 performs much better than the API on the task of reversal, showing that in at least some cases these different versions perform qualitatively differently. 
However, there are other tasks that our brief tests found were also difficult for the chat interface, such as shift ciphers; but we did not systematically investigate the chat interface because the size of our datasets would make this investigation prohibitive. In general, it does not cause problems for our hypotheses if there are some settings in which language models perform our tasks well, since we do not claim that our tasks are fundamentally impossible for language models but rather that, when LLMs are not at ceiling on these tasks, they will show certain predictable patterns regarding which queries they find easy or difficult. See Section~\ref{sec:notarguing} for further discussion.

\subsection{Notes about figures}

In many of our figures, we include examples of model predictions. These examples typically do not show the complete prompts that we used because these prompts were often long (e.g., because we included an example of the task to be performed). See Appendix~\ref{app:prompts} for the full prompts that we used.

\section{Sensitivity to task probability}

\begin{figure*}
    \centering
    \includegraphics[width=\textwidth]{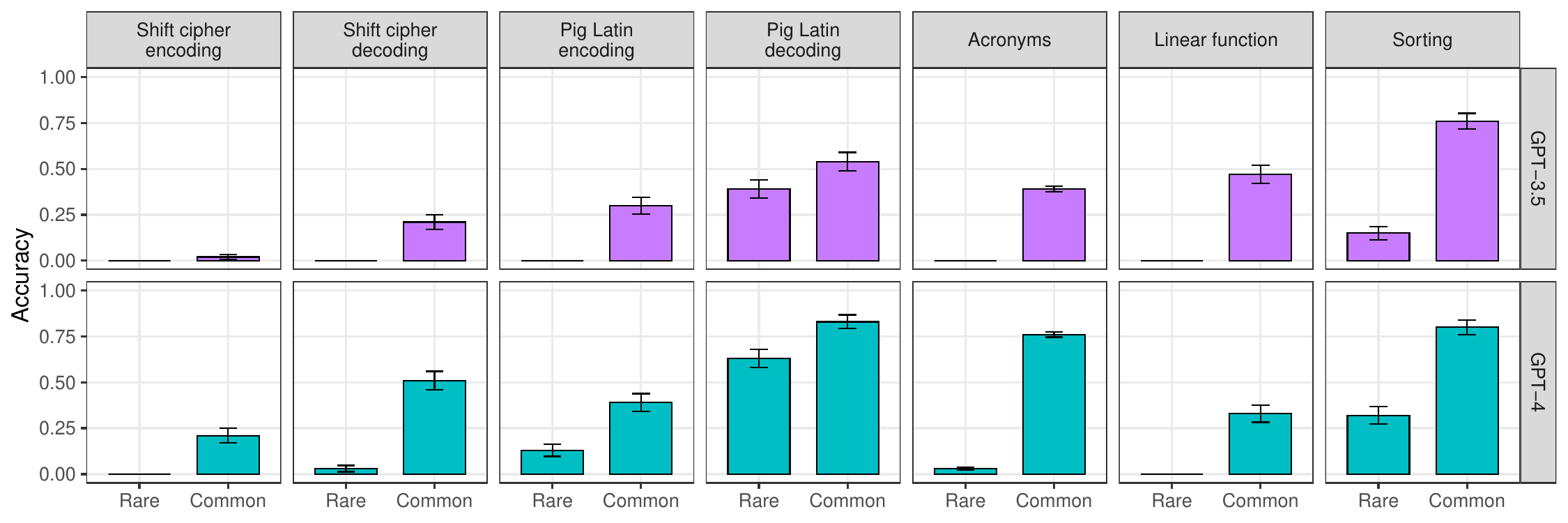}
    \caption{Comparing performance on common vs.\ rare versions of tasks. For shift ciphers, the common variant that we illustrate is rot-13, and the rare variant is rot-2. For Pig Latin, the common variant is Pig Latin, and the rare variant is Boar Etruscan (see Section~\ref{sec:taskprob_piglatin}). For acronyms, the common variant is acronyms formed from the first letters of words, while the rare variant is acronyms formed from the second letters of words. The common linear function is $f(x) = (9/5)x + 32$ (which is the Celsius-to-Fahrenheit conversion), and the rare linear function is $f(x) = (7/5)x + 31$. The common version of sorting is sorting in alphabetical order, while the rare variant uses reverse alphabetical order. Error bars show one standard error.}
    \label{fig:taskpairs}
\end{figure*}

We have hypothesized that LLMs will perform better on tasks that are frequently illustrated in Internet text than on tasks that occur more rarely---even when the rare task is no more complex than the common one. In this section we test this hypothesis by evaluating models on common variants of several tasks and rarer variants of those same tasks. Figure~\ref{fig:taskpairs} gives an overview of the results.

\subsection{Shift ciphers}

\paragraph{Summary:} When asked to decode shift ciphers with various shift levels, GPT-4 scores 50\% or above for the three most common shift levels (1, 3, and 13) yet scores below 3\% for all other shifts.

\begin{figure*}
        \centering

    {\hspace*{1.1cm}
    \begin{minipage}[l]{380pt}
    \begin{AIbox}{Shift ciphers}{375pt}
    \parbox[t]{\linewidth}{
    {
    {\small \textbf{Rot-13:} Decode by shifting each letter \uline{\textcolor{highlightpurple}{\textbf{13}}} positions backward in the alphabet.}
    }

    \bigskip
    
    \hspace*{-0.0cm}
    \begin{tabular}{@{}c@{~}l@{~~}p{13cm}}
        & {\small \textbf{Input:}} & \sethlcolor{white}{\footnotesize \texttt{\hl{Ohg guvf gvzr, gurer znl nyfb or nabgure ernfba.}}} \\
        & {\small \textbf{Correct:}} & \sethlcolor{white}{\footnotesize \texttt{\hl{But this time, there may also be another reason.}}}\\
        \textcolor{checkgreen}{\textbf{\checkmark}} & {\small \textbf{GPT-4:}} & \sethlcolor{aigreen}{\footnotesize \texttt{\hl{But this time, there may also be another reason.}}}
    \end{tabular}
    
    \tcbline

    {\small 
    \textbf{Rot-8:} Decode by shifting each letter \uline{\textcolor{highlightpurple}{\textbf{8}}} positions backward in the alphabet.}

    \bigskip

    \hspace*{-0.0cm}
    \begin{tabular}{@{}c@{~}l@{~~}p{9.8cm}}
        & {\small \textbf{Input:}} & \sethlcolor{white}{\footnotesize \texttt{\hl{Jcb bpqa bqum, bpmzm uig itaw jm ivwbpmz zmiawv.}}} \\
        & {\small \textbf{Correct:}} & \sethlcolor{white}{\footnotesize \texttt{\hl{But this time, there may also be another reason.}}}\\
        \redx{} & {\small \textbf{GPT-4:}} & \sethlcolor{pink}{\footnotesize \texttt{\hl{Say what you}\sethlcolor{aigreen}\hl{,}\sethlcolor{pink}\hl{ think and then }\sethlcolor{aigreen}\hl{be}\sethlcolor{pink}\hl{ silent}\sethlcolor{aigreen}\hl{.}}}
    \end{tabular}
    }
    \end{AIbox}
    \end{minipage}
    }

        \vspace{3mm}
        \includegraphics[width=0.8\textwidth]{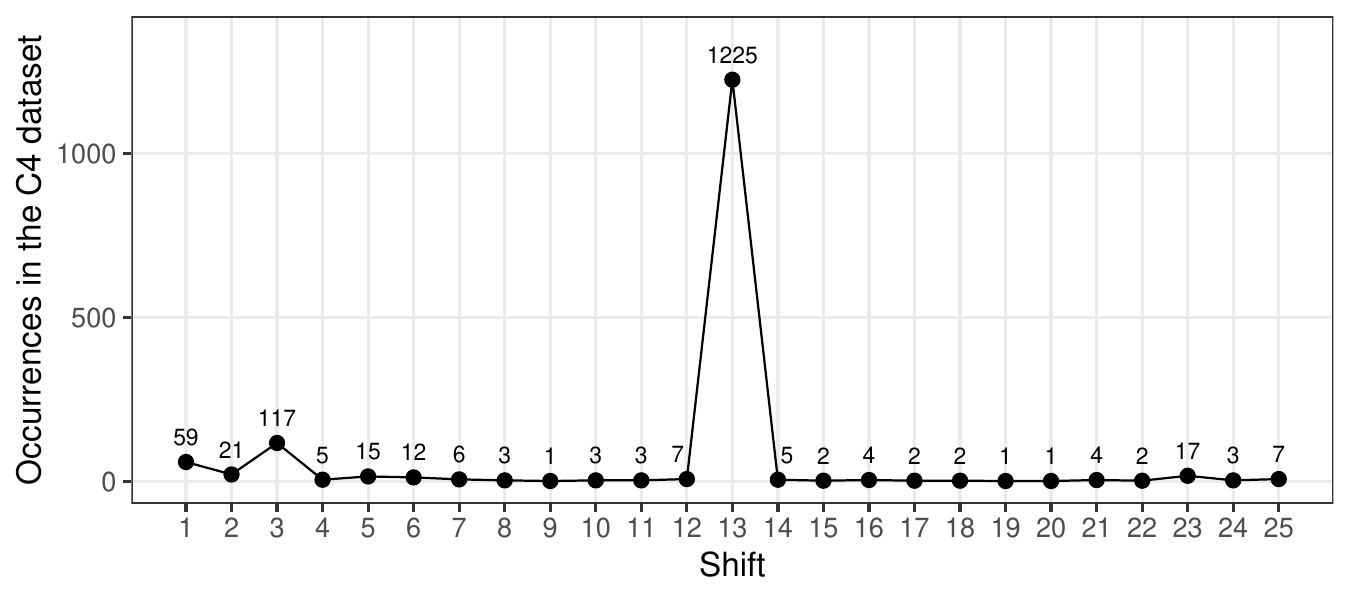}
        
        \includegraphics[width=0.8\textwidth]{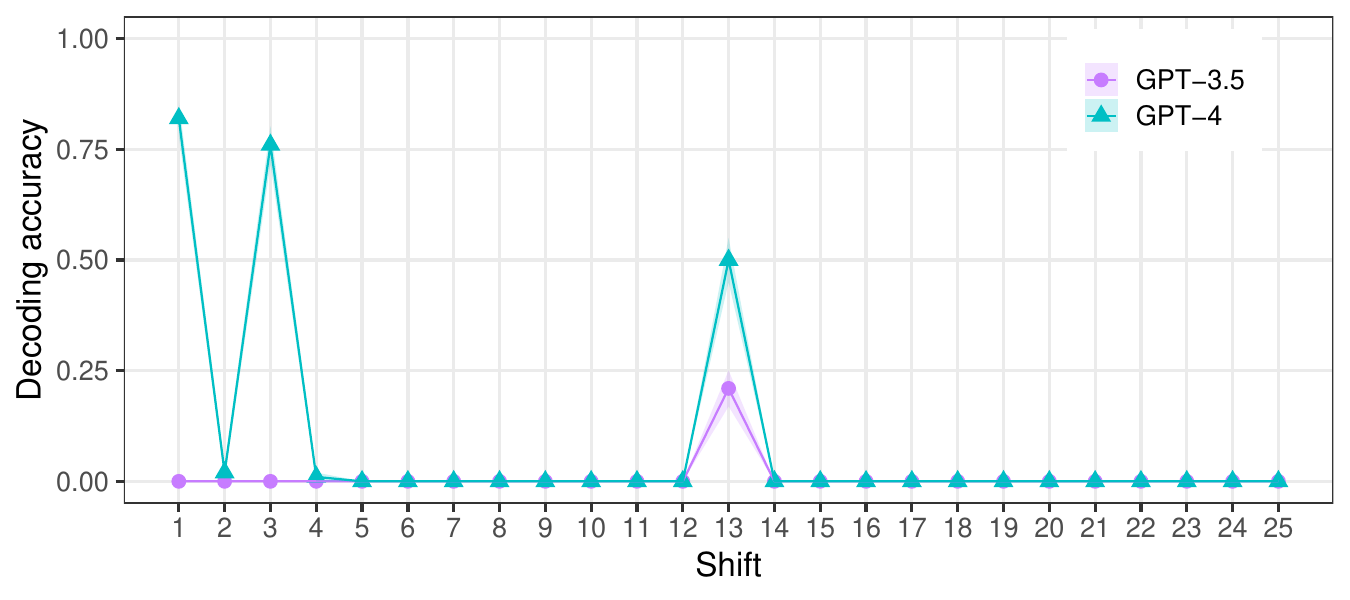}

    \caption{Analysis of shift ciphers with shift levels varying from 1 to 25. \textbf{Top:} Example of shift cipher stimuli for two shift levels. \textbf{Middle:} Number of occurrences of each shift level within the C4 corpus (see Appendix~\ref{app:shiftcounting} for details). \textbf{Bottom:} Accuracy on decoding shift ciphers with various shifts. The only shifts where models get non-negligible accuracy are 1, 3, and 13, which the middle plot shows to be the three most common shift levels.}
    \label{fig:shifts}
\end{figure*}

\paragraph{Details:}
The realm of shift ciphers provides an excellent setting to investigate the effects of task probability. There are 25 possible shift ciphers, corresponding to the 25 unique shifts that can be applied within the alphabet. These different shifts are not all used with the same frequency. Based on a manual analysis of the C4 corpus (\citealt{raffel2020exploring}), we found that the three most common shift levels are rot-1, rot-3, and rot-13 (Figure~\ref{fig:shifts}, middle; see Appendix~\ref{app:shiftcounting} for more details). 
Rot-1 is common because it is the smallest shift, so it is a natural first choice for illustrating the concept of a shift cipher (for instance, it is what we used for our initial example in Section~\ref{sec:describeshift}). Rot-3 is common because Julius Caesar famously used this cipher \citep{suetonius}.\footnote{Suetonius also notes that Augustus used a cipher with a shift of 1. Augustus does not get mentioned much in discussions of shift ciphers, so his use of rot-1 is probably not a major contributor to rot-1's frequency, but it nonetheless provides another illustration of the fact that rot-1 is a natural choice for a shift cipher.}
Because of this historical connection, shift ciphers are sometimes called Caesar ciphers, and many who describe or use shift ciphers follow Caesar's precedent by selecting a shift of 3 \citep[e.g.,][]{kahn1967codebreakers,singh1999code}. Finally, as mentioned above, rot-13 is common because there is a convention of using it in online message boards. The other twenty-two possible shifts are used occasionally but are not as common as these three.

If LLMs are indeed sensitive to task probability, we would therefore expect them to perform better on rot-1, rot-3, and rot-13 than on other shifts. To test this prediction, we evaluated our models on decoding text written with each shift level from 1 to 25.
Models were always told which shift cipher was being used and were shown an example of an English sentence and that sentence's encoding in the shift cipher at hand; see Appendix~\ref{sec:prompt_rot13_dec} for the exact prompt that was used.
The answer sentences were the 100 high-probability sentences that we drew from GlobalVoices (Section~\ref{sec:sentenceprobs}).

The results are striking (Figure~\ref{fig:shifts}, bottom): GPT-3.5's accuracy is 0.21 for rot-13 but 0.00 for all other shifts. GPT-4 scores 0.50 or above on the three highest-frequency shifts, but its accuracy is less than 0.03 for all other shifts. These results therefore follow the prediction that LLMs would perform better on commonly-used shifts than on rarely-used ones. 

The one aspect of the results that is unclear is why GPT-4 performs better on rot-1 and rot-3 than on rot-13, even though rot-13 is by far the most common shift cipher in C4. One possibility is that its training distribution may be meaningfully different from C4: rot-13 is most common on online forums while rot-1 and rot-3 show up most often in expository text about shift ciphers, so differences in the frequency of document types could influence the frequency of shift levels. 

Importantly, these results cannot be explained by the complexity of the different shift levels. From one perspective, all 25 shifts have the same level of complexity because they all require the same basic computational abilities (counting and shifting in the alphabet). If any difference in complexity could be said to exist, it would be that smaller shifts are simpler than large ones because they require less counting, but LLM performance is not driven by the size of the shift: for both models, rot-13 is substantially easier than many other, smaller shifts. 
One particularly striking comparison is that GPT-4 achieves an accuracy of 0.82 for rot-1 and 0.76 for rot-3 yet only 0.02 for rot-2, even though we might intuitively expect that a system that can perform rot-1 and rot-3 would also succeed at rot-2. 
Therefore, the results can be straightforwardly explained by each task's probability but not by its inherent difficulty.

To test the significance of the task effects observed here, we compared performance on rot-13 and rot-2 (as examples of a common and rare variant, respectively). Both models had a statistically significantly better performance on decoding rot-13 than rot-2 ($p < 0.01$ for both models; see Appendix~\ref{app:shiftsig2vs13}). We also evaluated both models on encoding for these two shift levels; in this case, GPT-4 still performed significantly better on rot-13 than rot-2 ($p < 0.01$), but for GPT-3.5 there was no significant difference, likely due to floor effects, since GPT-3.5's encoding accuracy was close to 0.0 for both task variants; see Figure~\ref{fig:taskpairs}.

\begin{figure}[bh!]
    \centering
    \begin{minipage}[l]{0.67\textwidth}

    \begin{AIbox}{Encoding with Pig Latin variants}{360pt}
    \parbox[t]{\linewidth}{

    \hspace{-0.2cm}
    \begin{tabular}{@{}p{12cm}}
         \textbf{Task:} Convert the following English sentence into Pig Latin (using a variant of Pig Latin that is specified in the prompt): ``\texttt{in times of crisis, art should be pragmatic.}''  \\
    \end{tabular}

    \bigskip

    \sethlcolor{yellow}
    \hspace*{-0.2cm}
    \begin{tabular}{@{}l@{~}p{13cm}}
        \multicolumn{2}{@{}l}{{\small \underline{\textbf{Pig Latin with \textit{-way}}}}} \\
        {\small \textbf{Correct:}} & {\footnotesize \texttt{inway imestay ofway isiscray, artway ouldshay ebay agmaticpray.}} \\
        {\small \textbf{GPT-4:}} & {\footnotesize \texttt{\sethlcolor{aigreen}\hl{inway} imestay \hl{ofway} isiscray, \hl{artway} \sethlcolor{pink}\hl{shouldway} ebay \hl{ragmaticpay}.}} \\ \\

        \multicolumn{2}{@{}l}{{\small \underline{\textbf{Pig Latin with \textit{-ay}}}}} \\
        {\small \textbf{Correct:}} & {\footnotesize \texttt{inay imestay ofay isiscray, artay ouldshay ebay agmaticpray.}} \\
        {\small \textbf{GPT-4:}} & {\footnotesize \texttt{\sethlcolor{aigreen}\hl{inay} imestay \hl{ofay} isiscray, \hl{artay} \sethlcolor{pink}\hl{shoulday} ebay \hl{ragmaticpay}.}} \\ \\

        \multicolumn{2}{@{}l}{{\small \underline{\textbf{Pig Latin with \textit{-yay}}}}} \\
        {\small \textbf{Correct:}} & {\footnotesize \texttt{inyay imestay ofyay isiscray, artyay ouldshay ebay agmaticpray.}} \\
        {\small \textbf{GPT-4:}} & {\footnotesize \texttt{\sethlcolor{pink}\hl{inay} imestay \hl{ofay} isiscray, \hl{artay} \hl{shouldyay} ebay \hl{ragmaticpay}.}} \\ \\

        \multicolumn{2}{@{}l}{{\small \underline{\textbf{Pig Latin with \textit{-hay}}}}} \\
        {\small \textbf{Correct:}} & {\footnotesize \texttt{inhay imestay ofhay isiscray, arthay ouldshay ebay agmaticpray.}} \\
        {\small \textbf{GPT-4:}} & {\footnotesize \texttt{\sethlcolor{pink}\hl{inay} imestay \sethlcolor{aigreen}\hl{ofhay} isiscray, \sethlcolor{pink}\hl{artay} \hl{houldsay} ebay \hl{ragmaticpay}.}} \\ \\

        \multicolumn{2}{@{}l}{{\small \underline{\textbf{Pig Latin with \textit{-say}}}}} \\
        {\small \textbf{Correct:}} & {\footnotesize \texttt{insay imestay ofsay isiscray, artsay ouldshay ebay agmaticpray.}} \\
        {\small \textbf{GPT-4:}} & {\footnotesize \texttt{\sethlcolor{pink}\hl{inay} imestay \sethlcolor{aigreen}\hl{ofsay} isiscray, \hl{artsay} \sethlcolor{pink}\hl{shouldsay} ebay \hl{ragmaticpray}.}} \\
        
    \end{tabular}

    }
    \end{AIbox}
    \end{minipage}
    \hfill
    \begin{minipage}[l]{0.3\textwidth}
    \includegraphics[scale=0.57]{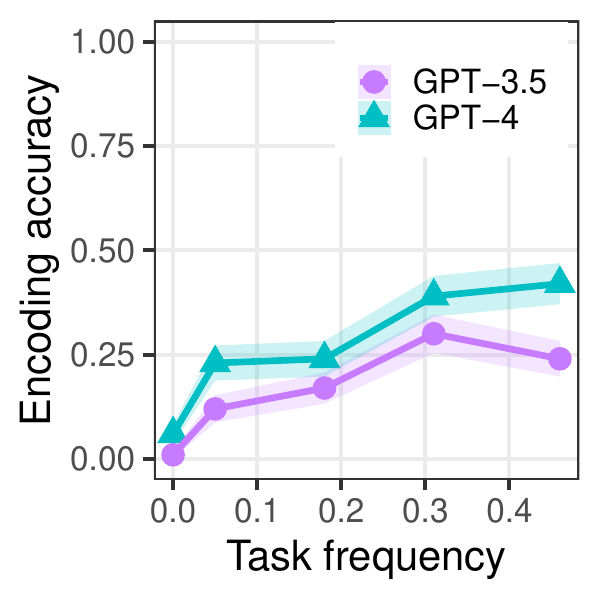}

    \vspace{7mm}

    \includegraphics[scale=0.57]{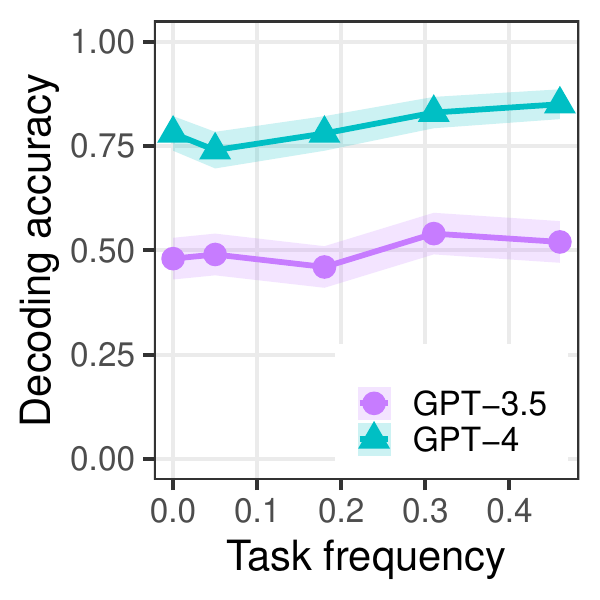}
        \end{minipage}
    \caption{\textbf{Left:} An English sentence and its equivalent in five Pig Latin variants, along with GPT-4's predictions for the Pig Latin sentences in these variants. The variants are listed from most to least frequent in the Pile corpus. Errors are highlighted in red; cases where GPT-4 produced the correct variant-specific form of a vowel-initial word are highlighted in green (note that the variants only differ in how they handle vowel-initial words). \textbf{Right:} Model accuracy on these variants when encoding (top) or decoding (bottom), as a function of the variant's frequency. The intervals around the lines show one standard error.}
    \label{fig:pigfrequency}
\end{figure}

\subsection{Pig Latin}\label{sec:taskprob_piglatin}

\paragraph{Summary:} GPT-4 achieves a much higher accuracy when converting a sentence into Pig Latin than when converting it into a different system that is no more complex but that is unattested in natural corpora (accuracy: 0.39 vs.\ 0.13). In addition, when comparing 4 minimally-differing variants that occur in natural corpora, GPT-4's performance increases with the corpus frequency of the variant, ranging from 0.23 to 0.42.

\paragraph{Details:} Pig Latin is a secret ``language'' based on English. To convert an English sentence into Pig Latin, the first consonant cluster of each word is moved to the end of the word and then \textit{-ay} is added to the end of the word. We compared Pig Latin to a system that we made up called Boar Etruscan that instead uses \textit{-uv} as the letter pair that is added at the end. Pig Latin and Boar Etruscan are illustrated below:

\ex. 
\a. \textbf{English:} there is a long and complicated history of how we ended up like this.
\b. \textbf{Pig Latin:} erethay isay aay onglay anday omplicatedcay istoryhay ofay owhay eway endeday upay ikelay isthay.
\c. \textbf{Boar Etruscan:} erethuv isuv auv ongluv anduv omplicatedcuv istoryhuv ofuv owhuv ewuv endeduv upuv ikeluv isthuv.

We tested GPT-3.5 and GPT-4 on both encoding into each of these fake languages and decoding from these fake languages. Model performance is shown in Figure~\ref{fig:taskpairs}, in which the ``common'' variant is Pig Latin and the ``rare'' variant is Boar Etruscan.
For encoding, both models performed much better on Pig Latin than Boar Etruscan, a difference that was statistically significant ($p < 0.01$ for GPT-3.5, $p < 0.001$ for GPT-4; see Appendix~\ref{app:pigsigtask}). For decoding, both models still scored better on Pig Latin than on Boar Etruscan, but the difference was now significant only for GPT-4 ($p < 0.05$) but not GPT-3.5 ($p = 0.08$). 

In addition to this binary comparison of attested vs.\ unattested, Pig Latin also gives us the opportunity for a finer-grained comparison because there are several variants of Pig Latin that have varying levels of commonness \citep{vaux2003underdetermination}. Specifically, all the major variants handle consonant-initial words the same (by moving the initial consonant cluster to the end and adding \textit{-ay}), but they vary in how they handle vowel-initial words: all the variants add some fixed syllable at the end of vowel-initial words, but the identity of that syllable varies. We identified all mentions of Pig Latin in the Pile dataset \citep{gao2020pile}, as a proxy for the training data of GPT models, and found 68 cases where the added syllable was \textit{-way}, 46 where it was \textit{-ay}, 26 where it was \textit{-yay}, and 8 where it was \textit{-hay}; an analysis of the C4 dataset \citep{raffel2020exploring} yielded similar proportions. We tested LLMs on all 4 of these variants, plus a fifth one where the added syllable was \textit{-say} (which has 0 occurrences in C4 and the Pile). See Figure~\ref{fig:pigfrequency} (left) for examples of the five Pig Latin variants. Models were always told which variant was being used and were given an example sentence in that variant; see Appendix~\ref{app:pigenc_prompt} for the specific prompt that was used.

We found that (particularly for encoding) model performance patterns with the frequency of the Pig Latin variant in the sample of internet text that we analyzed (Figure~\ref{fig:pigfrequency}). For both models, the effect of variant was significant for encoding ($p < 10^{-6}$ for GPT-3.5; $p < 10^{-8}$ for GPT-4) but not for decoding; see Appendix~\ref{app:pigsigtask}. Decoding accuracy may not vary much because the identity of an encoded vowel-initial word is easy to infer across all of these variants (since all that is required is removing a few letters at the end of the word), and models may find it easier to remove an unfamiliar ending (what must be done in decoding) than to produce an unfamiliar ending (what must be done in encoding).  This finding is striking because all five of these variants are very similar: they treat most words identically because they differ only for vowel-initial words; and even in these cases, they only differ from each other in a single letter. Nonetheless, these differences are enough to yield markedly differing performance along the lines we predicted.

\begin{figure}[t!]
    \centering
    \begin{minipage}[l]{0.64\textwidth}
    \centering
    \begin{AIbox}{Acronyms: Task probability}{345pt}
    \parbox[t]{\linewidth}{
    {
    {\small {\bf Common task: First-letter acronym.} 
    Combine the \underline{\textcolor{highlightpurple}{\textbf{first}}} letters of the words in the sequence.}
    }

    \bigskip
    
    \hspace*{-0.0cm}
    \begin{tabular}{@{}c@{~}l@{~~}p{9.8cm}}
        & {\small \textbf{Input:}} & \sethlcolor{white}{\footnotesize \texttt{\hl{penance aplenty rooster trample impasse envious subtext}}} \\
        & {\small \textbf{Correct:}} & \sethlcolor{white}{\footnotesize \texttt{\hl{PARTIES}}}\\
        \textcolor{checkgreen}{\textbf{\checkmark}} & {\small \textbf{GPT-4:}} & \sethlcolor{aigreen}{\footnotesize \texttt{\hl{PARTIES}}} \\ \\
        & {\small \textbf{Input:}} & \sethlcolor{white}{\footnotesize \texttt{\hl{unbound newness cranium likable emerald abalone reissue}}} \\
        & {\small \textbf{Correct:}} & \sethlcolor{white}{\footnotesize \texttt{\hl{UNCLEAR}}}\\
        \redx{} & {\small \textbf{GPT-4:}} & \sethlcolor{pink}{\footnotesize \texttt{\hl{UNCLERA}}} \\
    \end{tabular}
    
    \tcbline

    {\small {\bf Uncommon task: Second-letter acronym.} 
    Combine the \underline{\textcolor{highlightpurple}{\textbf{second}}} letters of the words in the sequence.}

    \bigskip

    \hspace*{-0.0cm}
    \begin{tabular}{@{}c@{~}l@{~~}p{9.8cm}}
        & {\small \textbf{Input:}} & \sethlcolor{white}{\footnotesize \texttt{\hl{aplenty maestro precept strayed figment megaton ascetic}}} \\
        & {\small \textbf{Correct:}} & \sethlcolor{white}{\footnotesize \texttt{\hl{PARTIES}}}\\
        \redx{} & {\small \textbf{GPT-4:}} & \sethlcolor{pink}{\footnotesize \texttt{\hl{PLEEGET}}} \\ \\
        & {\small \textbf{Input:}} & \sethlcolor{white}{\footnotesize \texttt{\hl{quattro ennoble scissor fluency regency hawkish pricked}}} \\
        & {\small \textbf{Correct:}} & \sethlcolor{white}{\footnotesize \texttt{\hl{UNCLEAR}}}\\
        \redx{} & {\small \textbf{GPT-4:}} & \sethlcolor{pink}{\footnotesize \texttt{\hl{UENLUEAW}}} \\
    \end{tabular}
    }
    \end{AIbox}
    \end{minipage}
    \hspace{2mm}
    \begin{minipage}[r]{0.3\textwidth}
    \centering
    \vspace{0.3cm}
    
    \includegraphics[scale=0.55]{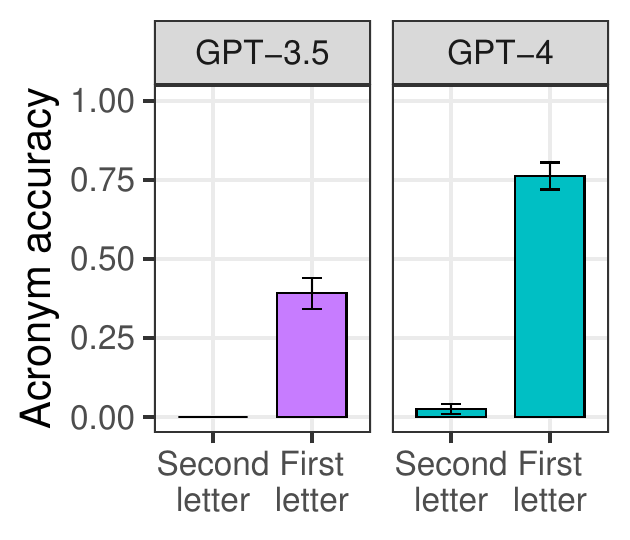}

    \end{minipage}
    \caption{\textbf{Left:} Examples of GPT-4 predictions on the acronym tasks. \textbf{Right:} Model accuracy at acronym formation (this plot repeats the acronym results from Figure~\ref{fig:taskpairs}). Error bars show one standard error.}
    \label{fig:acronym_pairs}
\end{figure}

\subsection{Acronyms} 

\paragraph{Summary:} GPT-4 performed much better at the common operation of joining the first letter of each word (accuracy: 0.76) than the rare operation of joining the second letter of each word (accuracy: 0.03).

\paragraph{Details:} 
People frequently join together the first letter of each word in a sequence of words, but it is rare to join the second letter of each word. We therefore use first-letter acronym formation (e.g., \ref{ex:firstletter}) as a common task and second-letter acronym formation (e.g., \ref{ex:secondletter}) as a rare task:

\ex. First-letter acronym\label{ex:firstletter}
\a. \textbf{Input:} revolve edifice scrappy panicky outlast negated drizzle
\b. \textbf{Output:} RESPOND 

\ex. Second-letter acronym\label{ex:secondletter}
\a. \textbf{Input:} prequel leaping ascetic splurge policed invader edifice
\b. \textbf{Output:} RESPOND 

The outputs were matched across the two tasks (e.g., the examples above both use the answer RESPOND), and the input words were drawn from the same vocabulary across tasks (specifically, the highest-probability bin of input words described in Section~\ref{sec:acronyms_output_prob}).
For both models, performance was much higher on the first-letter acronyms than second-letter ones, and the effect was statistically significant ($p < 10^{-4}$ for GPT-3.5, and $p < 10^{-15}$ for GPT-4; see Appendix~\ref{app:acronymsig1vs2}). See Figure~\ref{fig:acronym_pairs} for examples of model predictions on the acronym tasks. Note that \citep{wei2022chain} found that chain-of-thought prompting could substantially improve accuracy on a task similar to our acronym task, suggesting that our task might similarly show improved performance in that setting; however, if this prediction is borne out, it would not affect our claims, for the reasons detailed in Section~\ref{sec:notarguing}.

\begin{figure}[t!]
    \centering
    \begin{minipage}[l]{0.5\textwidth}
    \centering
    \begin{AIbox}{Linear functions: Task probability}{280pt}
    \parbox[t]{\linewidth}{
    {
    {\small {\bf Common function:} 
    $f(x) = (9/5)x + 32$}
    }

    \bigskip
    
    \hspace*{-0.0cm}
    \begin{tabular}{@{}c@{~}l@{~~}p{1cm}p{1.5cm}@{}c@{~}l@{~~}p{2cm}}
        & {\small \textbf{Input:}} & \sethlcolor{white}{\footnotesize \texttt{\hl{64}}} & & & {\small \textbf{Input:}} & \sethlcolor{white}{\footnotesize \texttt{\hl{577}}} \\
        & {\small \textbf{Correct:}} & \sethlcolor{white}{\footnotesize \texttt{\hl{147.2}}} & & & {\small \textbf{Correct:}} & \sethlcolor{white}{\footnotesize \texttt{\hl{1070.6}}}\\
        \textcolor{checkgreen}{\textbf{\checkmark}} & {\small \textbf{GPT-4:}} & \sethlcolor{aigreen}{\footnotesize \texttt{\hl{147.2}}} & & \redx{} & {\small \textbf{GPT-4:}} & \sethlcolor{pink}{\footnotesize \texttt{\hl{1069.6}}}\\ 
    \end{tabular}
    
    \tcbline

    {\small {\bf Rare function:} 
    $f(x) = (7/5)x + 31$}

    \bigskip

    \hspace*{-0.0cm}
    \begin{tabular}{@{}c@{~}l@{~~}p{1cm}p{1.5cm}@{}c@{~}l@{~~}p{2cm}}
        & {\small \textbf{Input:}} & \sethlcolor{white}{\footnotesize \texttt{\hl{64}}} & & & {\small \textbf{Input:}} & \sethlcolor{white}{\footnotesize \texttt{\hl{577}}} \\
        & {\small \textbf{Correct:}} & \sethlcolor{white}{\footnotesize \texttt{\hl{120.6}}} & & & {\small \textbf{Correct:}} & \sethlcolor{white}{\footnotesize \texttt{\hl{838.8}}} \\
        \redx{} & {\small \textbf{GPT-4:}} & \sethlcolor{pink}{\footnotesize \texttt{\hl{89.8}}} & & \redx{} & {\small \textbf{GPT-4:}} & \sethlcolor{pink}{\footnotesize \texttt{\hl{805.4}}}\\
    \end{tabular}
    }
    \end{AIbox}
    \end{minipage}
    \hspace{1cm}
    \begin{minipage}[r]{0.3\textwidth}
    \centering
    \vspace{0.3cm}
    
    \includegraphics[scale=0.55]{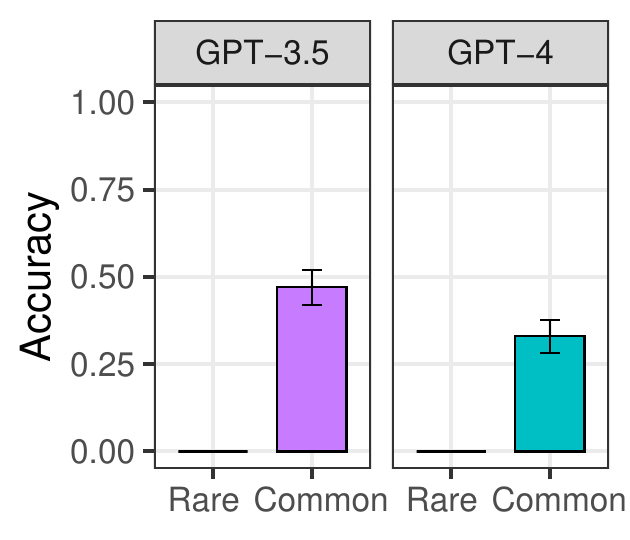}

    \end{minipage}
    \caption{\textbf{Left:} Examples of GPT-4 predictions on the linear functions. \textbf{Right:} Model accuracy at linear functions (this plot repeats the linear function results from Figure~\ref{fig:taskpairs}) Error bars show one standard error.}
    \label{fig:linear_pairs}
\end{figure}

\subsection{Linear functions}\label{sec:taskfreqlinear}

\paragraph{Summary:} GPT-4 performed much better at applying a linear function that is commonly illustrated in text (the Celsius-to-Fahrenheit conversion) than one that is similar but rare (accuracies: 0.33 vs.\ 0.00).

\paragraph{Details:} We compared two different tasks where models had to apply a linear function to a given number. As a common function, we used $f(x) = (9/5)x + 32$; this function occurs frequently in natural text because it is the function that converts temperatures from Celsius to Fahrenheit, so text written for multinational audiences often includes nearby pairs of numbers illustrating both $x$ and $f(x)$.
As a rare function, we used $f(x) = (7/5)x + 31$, which has no special significance and thus does not show up often in natural text. Both models scored reasonably well for the common function yet had an accuracy of 0\% for the rarer one (Figure~\ref{fig:linear_pairs}).

The results shown in Figure~\ref{fig:linear_pairs} use the same inputs across the rare and common math tasks, necessarily resulting in different outputs. To ensure that the different results were not simply driven by one function producing outputs that were higher-probability, we also ran a different version of the task where the outputs were the same across tasks (necessitating different inputs), and the results remained similar: on the common task, GPT-3.5 and GPT-4 scored 0.60 and 0.44, respectively, while their scores on the rare task were 0.01 and 0.00. For both models, in both the output-controlled setting and the input-controlled setting, the difference between tasks was statistically significant (with $p < 0.01$; see Appendix~\ref{app:linearsigtask})

\subsection{Sorting}

\paragraph{Summary:} GPT-4 performs better at sorting a list of words into alphabetical order (accuracy: 0.80) than reverse alphabetical order (accuracy: 0.32). Alphabetical order is mentioned much more commonly in the C4 dataset than reverse alphabetical order. In contrast, ascending order and descending order are mentioned with similar frequencies, and when sorting lists of numbers GPT-4 shows similar accuracy with ascending order (accuracy: 0.82) and descending order (accuracy: 0.80).

\begin{figure}[t!]
    \centering
    \begin{minipage}[l]{0.66\textwidth}
    \centering
    \begin{AIbox}{Sorting: Task probability}{365pt}
    \parbox[t]{\linewidth}{
    {
    {\small {\bf Common task:} 
    Sort in \underline{\textcolor{highlightpurple}{\textbf{alphabetical}}} order.}
    }

    \bigskip
    
    \hspace*{-0.0cm}
    \begin{tabular}{@{}c@{~}l@{~~}p{9.8cm}}
        & {\small \textbf{Input:}} & \sethlcolor{white}{\footnotesize \texttt{\hl{big, evergreen, flagrant, donkey, illegal, clean, exalted, normative, bountiful, archaic, insufferable, better, fair}}} \\
        & {\small \textbf{Correct:}} & \sethlcolor{white}{\footnotesize \texttt{\hl{archaic, better, big, bountiful, clean, donkey, evergreen, exalted, fair, flagrant, illegal, insufferable, normative}}}\\
        \textcolor{checkgreen}{\textbf{\checkmark}} & {\small \textbf{GPT-4:}} & \sethlcolor{aigreen}{\footnotesize \texttt{\hl{archaic, better, big, bountiful, clean, donkey, evergreen, exalted, fair, flagrant, illegal, insufferable, normative}}} \\
    \end{tabular}
    
    \tcbline

    {\small {\bf Uncommon task:} 
    Sort in \underline{\textcolor{highlightpurple}{\textbf{reverse alphabetical}}} order.}

    \bigskip

    \hspace*{-0.0cm}
    \begin{tabular}{@{}c@{~}l@{~~}p{9.8cm}}
        & {\small \textbf{Input:}} & \sethlcolor{white}{\footnotesize \texttt{\hl{big, evergreen, flagrant, donkey, illegal, clean, exalted, normative, bountiful, archaic, insufferable, better, fair}}} \\
        & {\small \textbf{Correct:}} & \sethlcolor{white}{\footnotesize \texttt{\hl{normative, insufferable, illegal, flagrant, fair, exalted, evergreen, donkey, clean, bountiful, big, better, archaic}}}\\
        \redx{} & {\small \textbf{GPT-4:}} & \sethlcolor{aigreen}{\footnotesize \texttt{\hl{normative, insufferable, illegal, }\sethlcolor{pink}\hl{evergreen, exalted, fla-grant, fair}\sethlcolor{aigreen}\hl{, donkey, clean, bountiful, big, better, archaic}}} \\
    \end{tabular}
    }
    \end{AIbox}
    \end{minipage}
    \hfill
    \begin{minipage}[r]{0.3\textwidth}
    \centering
    \vspace{0.3cm}
    
    \includegraphics[scale=0.55]{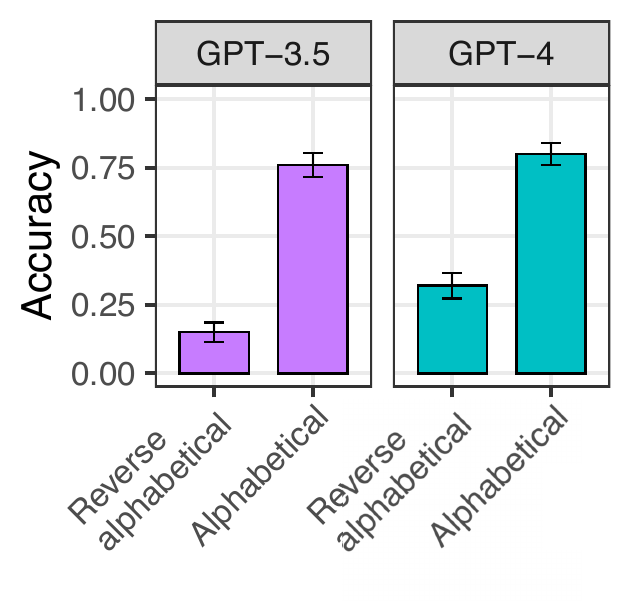}

    \end{minipage}
    \caption{\textbf{Left:} Examples of GPT-4 predictions on sorting tasks. \textbf{Right:} Model accuracy at sorting (this plot repeats the sorting results from Figure~\ref{fig:taskpairs}). Error bars show one standard error.}
    \label{fig:sorting_pairs}
\end{figure}

\paragraph{Details:} We gave models a list of words and told them to sort the list into either alphabetical order or reverse alphabetical order. Based on a corpus analysis using the C4 corpus, we estimate that alphabetical order is approximately 150 times more common than reverse alphabetical order in Internet text (see Appendix~\ref{app:sortingfreq}). 
Paralleling this difference in frequency, both models perform better on alphabetical order than reverse alphabetical order (Figure~\ref{fig:sorting_pairs}). This difference between orderings is statistically significant for both GPT-3.5 ($p < 10^{-8}$) and GPT-4 ($p < 10^{-6}$); see Appendix~\ref{app:sortingsig}.

Although these two ways of sorting \textit{words} show a marked difference in frequency, the two most prevalent ways of sorting \textit{numbers}---in ascending order or in descending order---do not show such a large difference. Specifically, we found that descending order is mentioned approximately 1.5 times as often as ascending order in the C4 corpus, a much smaller ratio than the ratio of 150 found for word orderings; further, we suspect that ascending order may be used without being explicitly named more often than descending order is, which would further reduce the ratio. As we would expect based on the similar corpus frequency levels, models' performance on sorting numbers is similar when they are asked to sort in ascending order or descending order (Table~\ref{tab:sortingstats}), and logistic regressions found no statistically significant difference between the two orderings for either model (Appendix~\ref{app:sortingsig}).

In sum, when we tested two orderings that had a large frequency difference (alphabetical order vs.\ reverse alphabetical order), we found a large difference in model performance across orderings. However, when we compared two orderings that had similar levels of frequency (ascending order vs.\ descending order), models showed little difference in performance across ordering. These results support the broader claim that model performance on a task is influenced by the task's frequency in pretraining data.

\begin{table}[]
    \centering
    \begin{tabular}{ccccc} \toprule
         & Count in C4 & GPT-3.5 accuracy & GPT-4 accuracy \\ \midrule
        Alphabetical order & 95,942 & 0.76 &  0.80 \\
        Reverse alphabetical order & 629 & 0.15 & 0.32 \\ \midrule
        Ascending order & 21,562 & 0.66 & 0.82 \\
        Descending order & 31,378 & 0.58 & 0.80 \\ \bottomrule
    \end{tabular}
    \caption{Corpus counts and model accuracies for various sorting tasks. When alphabetical and reverse alphabetical order were being tested, the units to be sorted were words. When ascending and descending order were being tested, the units to be sorted were numbers.}
    \label{tab:sortingstats}
\end{table}

\subsection{Summary of task probability effects}

We investigated seven pairs of tasks, with each pair containing a common task variant and a rare task variant. Across all seven pairs, models performed substantially better on the common task variant than the rare one.

Contemporaneous work by \citet{wu2023reasoning} tested a similar hypothesis and drew similar conclusions: they evaluated LLM performance on the default version of a task (e.g., executing Python code under the true assumption that Python uses 0-based indexing) and a counterfactual version of the task (e.g., adopting the false assumption that Python uses 1-based indexing). They found that LLMs reliably performed worse on the counterfactual tasks than the default tasks. 
We view our work as highlighting a general phenomenon of which \citeauthor{wu2023reasoning}'s conclusion is a special case: we show that LLMs are sensitive to task probability, a factor that encompasses the default-vs.-counterfactual situations that \citeauthor{wu2023reasoning} study but  that also includes other cases where there is no default. For instance, we showed that model accuracy varied along with task probability for linear functions even though there is no default among the class of linear functions; similarly, model accuracy on Pig Latin was influenced by the frequency of Pig Latin variants even though there is no agreed-upon default version of Pig Latin. 
\citeauthor{wu2023reasoning} also discuss continuous task frequency in their Section 5.1 as an explanation for why some counterfactual task variants are harder than others, but this discussion still operates within the default-vs.-counterfactual distinction.
As part of our more direct focus on probability, we used corpus analyses to measure task frequency for several of our experiments, a type of analysis that was not used in \citeauthor{wu2023reasoning}'s work.
A final difference is that we situate task frequency within a broader teleological perspective, while \citeauthor{wu2023reasoning}'s motivation for analyzing task variants is a distinct (but also very important) goal of evaluating how general LLM abilities are.
Overall, because our work and \citeauthor{wu2023reasoning}'s use non-overlapping sets of tasks and draw compatible conclusions,
we view these two papers as mutually reinforcing.

\section{Sensitivity to output probability}\label{sec:outputprob}

We have hypothesized that LLMs will perform better when the correct answer is a high-probability string than when it is a low-probability string, even in deterministic situations where the answer could be determined without considering probability.
In this section we test this hypothesis.

\begin{figure}
    \centering
    \begin{minipage}[l]{0.64\textwidth}
    \centering
    \begin{AIbox}{Shift cipher: Output probability}{350pt}
    \parbox[t]{\linewidth}{

    {\small \underline{\textbf{Rot-13: Example of high-probability output}}}
    
    \begin{tabular}{@{}c@{~}l@{~~}p{9.8cm}}
        & {\small \textbf{Input:}} & \sethlcolor{white}{\footnotesize \texttt{\hl{Gur erfbyhgvba bs pbasyvpgf unf gb or onfrq ba gur gehgu.}}} \\
        & {\small \textbf{Correct:}} & \sethlcolor{white}{\footnotesize \texttt{\hl{The resolution of conflicts has to be based on the truth.}}}\\
        \textcolor{checkgreen}{\textbf{\checkmark}} & {\small \textbf{GPT-4:}} & \sethlcolor{aigreen}{\footnotesize \texttt{\hl{The resolution of conflicts has to be based on the truth.}}}
    \end{tabular}

    \bigskip
    
    {\small \underline{\textbf{Rot-13: example of medium-probability output}}}

    \begin{tabular}{@{}c@{~}l@{~~}p{9.8cm}}
        & {\small \textbf{Input:}} & \sethlcolor{white}{\footnotesize \texttt{\hl{Nyy rfpnyngvba bs shapgvbaf unq gb or onfrq ba uvf sbphf.}}} \\
        & {\small \textbf{Correct:}} & \sethlcolor{white}{\footnotesize \texttt{\hl{All escalation of functions had to be based on his focus.}}}\\
        \redx{} & \textbf{GPT-4:} & \sethlcolor{aigreen}{\footnotesize \texttt{\hl{All escalation of     }\sethlcolor{pink}\hl{conflicts}\sethlcolor{aigreen}\hl{ had to be based on his }\sethlcolor{pink}\hl{proofs}\sethlcolor{aigreen}\hl{.}}}
    \end{tabular}
    }
    \end{AIbox}
    \end{minipage}
    \hfill
    \begin{minipage}[r]{0.3\textwidth}
    \centering
    \vspace{0.7cm}
    
    \includegraphics[scale=0.55]{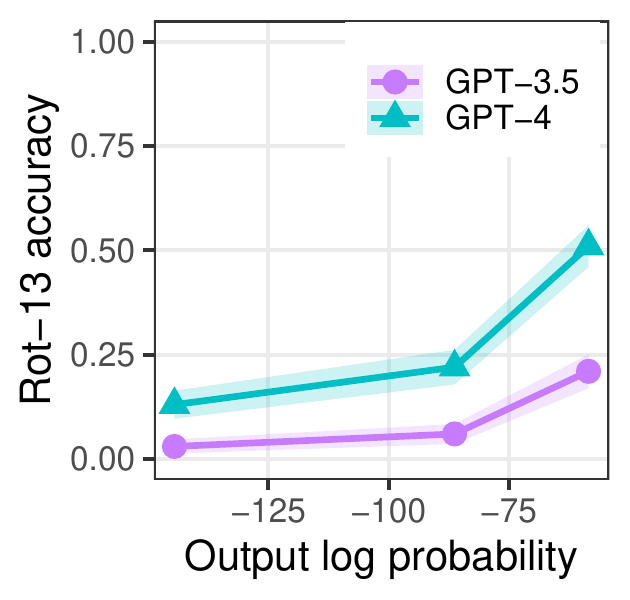}
    \end{minipage}

    \begin{minipage}[l]{0.64\textwidth}
    \centering
    \begin{AIbox}{Reversal: Output probability}{350pt}
    \parbox[t]{\linewidth}{

    {\small \underline{\textbf{Reversal: Example of medium-probability output}}}
    
    \begin{tabular}{@{}c@{~}l@{~~}p{9.8cm}}
        & {\small \textbf{Input:}} & \sethlcolor{white}{\footnotesize \texttt{\hl{paintings. the with pleased totally not was he True,}}} \\
        & {\small \textbf{Correct:}} & \sethlcolor{white}{\footnotesize \texttt{\hl{True, he was not totally pleased with the paintings.}}}\\
        \textcolor{checkgreen}{\textbf{\checkmark}} & {\small \textbf{GPT-4:}} & \sethlcolor{aigreen}{\footnotesize \texttt{\hl{True, he was not totally pleased with the paintings.}}}
    \end{tabular}

    \bigskip
    
    {\small \underline{\textbf{Reversal: example of low-probability output}}}

    \begin{tabular}{@{}c@{~}l@{~~}p{9.8cm}}
        & {\small \textbf{Input:}} & \sethlcolor{white}{\footnotesize \texttt{\hl{paintings. the with pleased he totally was not True,}}} \\
        & {\small \textbf{Correct:}} & \sethlcolor{white}{\footnotesize \texttt{\hl{True, not was totally he pleased with the paintings.}}}\\
        \redx{} & \textbf{GPT-4:} & \sethlcolor{aigreen}{\footnotesize \texttt{\hl{True, not was     }\sethlcolor{pink}\hl{he totally}\sethlcolor{aigreen}\hl{ pleased with the paintings.}}}
    \end{tabular}
    }
    \end{AIbox}
    \end{minipage}
    \hfill
    \begin{minipage}[r]{0.3\textwidth}
    \centering
    \vspace{0.7cm}
    
    \includegraphics[scale=0.55]{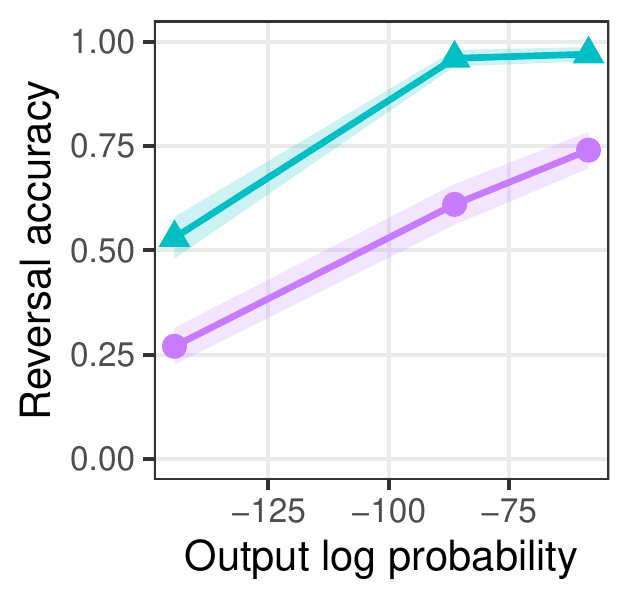}
    \end{minipage}

    \vspace{3mm}

    \begin{minipage}[l]{0.64\textwidth}
    \centering
    \begin{AIbox}{Pig Latin: Output probability}{350pt}
    \parbox[t]{\linewidth}{

    {\small \underline{\textbf{Pig Latin: Example of high-probability output}}}
    
    \begin{tabular}{@{}c@{~}l@{~~}p{8.5cm}}
        & {\small \textbf{Input:}} & \sethlcolor{white}{\footnotesize \texttt{\hl{utbay isthay imetay, erethay aymay alsoay ebay} \newline \hl{anotheray easonray.}}} \\
        & {\small \textbf{Correct:}} & \sethlcolor{white}{\footnotesize \texttt{\hl{but this time, there may also be another reason.}}}\\
        \textcolor{checkgreen}{\textbf{\checkmark}} & {\small \textbf{GPT-4:}} & \sethlcolor{aigreen}{\footnotesize \texttt{\hl{but this time, there may also be another reason.}}}
    \end{tabular}

    \bigskip
    
    {\small \underline{\textbf{Pig Latin: example of medium-probability output}}}

    \begin{tabular}{@{}c@{~}l@{~~}p{8.5cm}}
        & {\small \textbf{Input:}} & \sethlcolor{white}{\footnotesize \texttt{\hl{etyay ithway ustray, erethay aymay evenay ebay} \newline \hl{anotheray eformray.}}} \\
        & {\small \textbf{Correct:}} & \sethlcolor{white}{\footnotesize \texttt{\hl{yet with rust, there may even be another reform.}}}\\
        \redx{} & \textbf{GPT-4:} & \sethlcolor{aigreen}{\footnotesize \texttt{\hl{yet with     }\sethlcolor{pink}\hl{trust}\sethlcolor{aigreen}\hl{, there may even be another reform.}}}
    \end{tabular}
    }
    \end{AIbox}
    \end{minipage}
    \hfill
    \begin{minipage}[r]{0.3\textwidth}
    \centering
    \vspace{0.7cm}
    
    \includegraphics[scale=0.55]{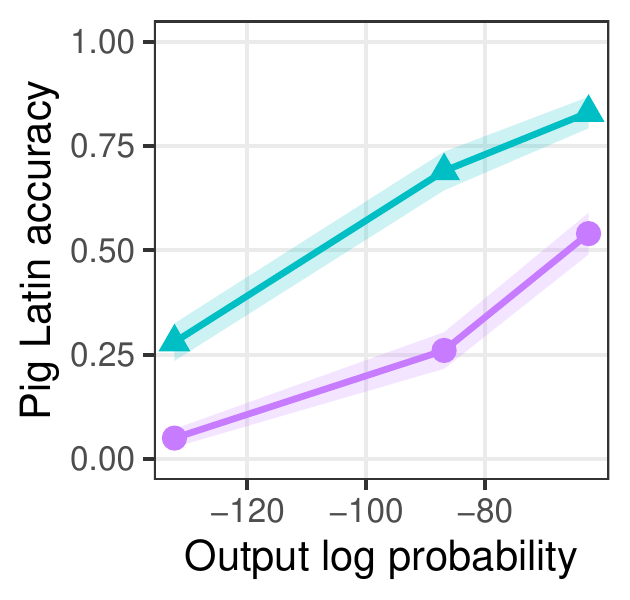}
    \end{minipage}

    \caption{The effect of output probability on model accuracy across three tasks. In all plots, the intervals around the lines show one standard error.}
    \label{fig:decoding_outputprob}
\end{figure}

\subsection{Shift ciphers}

\paragraph{Summary:} When deciphering rot-13 text, GPT-4's accuracy ranges from 0.51 when the output sentence is high-probability to 0.13 when the output sentence is low-probability. Through manual analyses of model predictions, we find that many model errors arise in situations where the correct answer is low-probability, in which case models often produce an output that is incorrect but has a higher probability. 

\paragraph{Details:} We asked models to decode sentences written in rot-13, with the target outputs belonging to our high-probability, medium-probability, or low-probability set of sentences (Section~\ref{sec:sentenceprobs}). We used GPT-2 \citep{radford2019language} to estimate the probability of the sentences in each condition and verified that the conditions differed in output probability but were similar in input probability (Table~\ref{tab:dataset_stats} on page \pageref{tab:dataset_stats}).

Model performance varied substantially with output probability (Figure~\ref{fig:decoding_outputprob}, top). For example, when the correct answer was high-probability, GPT-4 had an accuracy of 0.51, but when it was low-probability, the accuracy dropped to 0.13. Using a logistic regression, we found that output log probability indeed had a statistically significant effect for both models ($p < 10^{-4}$ for both; see Appendix~\ref{app:shiftsigdec}).

\paragraph{Targeted examples:} As a different way to investigate the effects of output probability, we tested models on examples that were more targeted, such that our hypotheses not only predict poor performance but also make predictions about \textit{which} incorrect answers models will produce. 
To create such examples, we started with our high-probability sentences and then manually changed one word in each to a new word that, in the context of the sentence, was still grammatical but was unlikely. The new word always had a Levenshtein edit distance \citep{levenshtein1966binary} of at most 2 from the original word (counting a single insertion, deletion, or substitution as 1 edit). For example, given \ref{ex:adversariala}, we produced \ref{ex:adversarialb}. We evaluated GPT-3.5 and GPT-4 on decrypting rot-13 text where the correct outputs were our modified sentences.

\ex. 
\a. Well, if they don't come, so be it.\label{ex:adversariala}
\b. Well, if they don't code, so be it.\label{ex:adversarialb}

\begin{figure}[t!]
    \centering
    \begin{minipage}[l]{0.67\textwidth}
    \centering
    \begin{AIbox}{Shift ciphers: Output regularization}{365pt}
    \parbox[t]{\linewidth}{

    {\small \underline{\textbf{Full sentence correct:}}}
    
    \begin{tabular}{@{}c@{~}l@{~~}p{10.5cm}}
        & {\small \textbf{Input:}} & \sethlcolor{white}{\footnotesize \texttt{\hl{Gung'f bhe jnl bs gunaxvat naq jung jr onfr bhe jbex ba.}}} \\
        & {\small \textbf{Correct:}} & \sethlcolor{white}{\footnotesize \texttt{\hl{That's our way of  {\underline{thanking}} and what we base our work on.}}}\\
        \textcolor{checkgreen}{\textbf{\checkmark}} & {\small \textbf{GPT-4:}} & \sethlcolor{white}{\footnotesize \texttt{\hl{That's our way of  }\sethlcolor{aigreen}\hl{{\underline{thanking}}}\sethlcolor{white}\hl{ and what we base our work on.}}}
    \end{tabular}

    \bigskip
    
    {\small \underline{\textbf{Full sentence regularized}}}

    \begin{tabular}{@{}c@{~}l@{~~}p{8.5cm}}
        & {\small \textbf{Input:}} & \sethlcolor{white}{\footnotesize \texttt{\hl{V jvyy abg tvir hc rira sbe n zbahzrag.}}} \\
        & {\small \textbf{Correct:}} & \sethlcolor{white}{\footnotesize \texttt{\hl{I will not give up even for a {\underline{monument}}.}}}\\
        \redx{} & \textbf{GPT-4:} & \sethlcolor{white}{\footnotesize \texttt{\hl{I will not give up even for a }\sethlcolor{pink}\hl{{\underline{moment}}}\sethlcolor{white}\hl{.}}}
    \end{tabular}

    \bigskip
    
    {\small \underline{\textbf{Critical word correct (but with an error elsewhere)}}}

    \begin{tabular}{@{}c@{~}l@{~~}p{10.5cm}}
        & {\small \textbf{Input:}} & \sethlcolor{white}{\footnotesize \texttt{\hl{Gur eryngvbafuvc orgjrra gurfr tebhcf vf bhe znvy punyyratr.}}} \\
        & {\small \textbf{Correct:}} & \sethlcolor{white}{\footnotesize \texttt{\hl{The relationship between these groups is our {\underline{mail}} challenge.}}}\\
        \redx{} & \textbf{GPT-4:} & \sethlcolor{white}{\footnotesize \texttt{\hl{The }\sethlcolor{pink}\hl{relationships}\sethlcolor{white}\hl{ between these groups is our }\sethlcolor{aigreen}\hl{{\underline{mail}}}\sethlcolor{white}\hl{ challenge.}}}
    \end{tabular}

    \bigskip

    {\small \underline{\textbf{Critical word regularized (with an additional error elsewhere)}}}

    \begin{tabular}{@{}c@{~}l@{~~}p{10.5cm}}
        & {\small \textbf{Input:}} & \sethlcolor{white}{\footnotesize \texttt{\hl{Gur fvghngvba ng gur havirefvgl yrirr vf dhvgr qrcybenoyr.}}} \\
        & {\small \textbf{Correct:}} & \sethlcolor{white}{\footnotesize \texttt{\hl{The situation at the university {\underline{levee}} is quite deplorable}}}\\
        \redx{} & \textbf{GPT-4:} & \sethlcolor{white}{\footnotesize \texttt{\hl{The situation at the university }\sethlcolor{pink}\hl{{\underline{level}}}\sethlcolor{white}\hl{ is quite }\sethlcolor{pink}\hl{dependable}\sethlcolor{white}\hl{.}}}
    \end{tabular}
    }
    \end{AIbox}
    \end{minipage}
    \hfill
    \begin{minipage}[l]{0.3\textwidth}
    \centering
    \vspace{12mm}
    \includegraphics[scale=0.6]{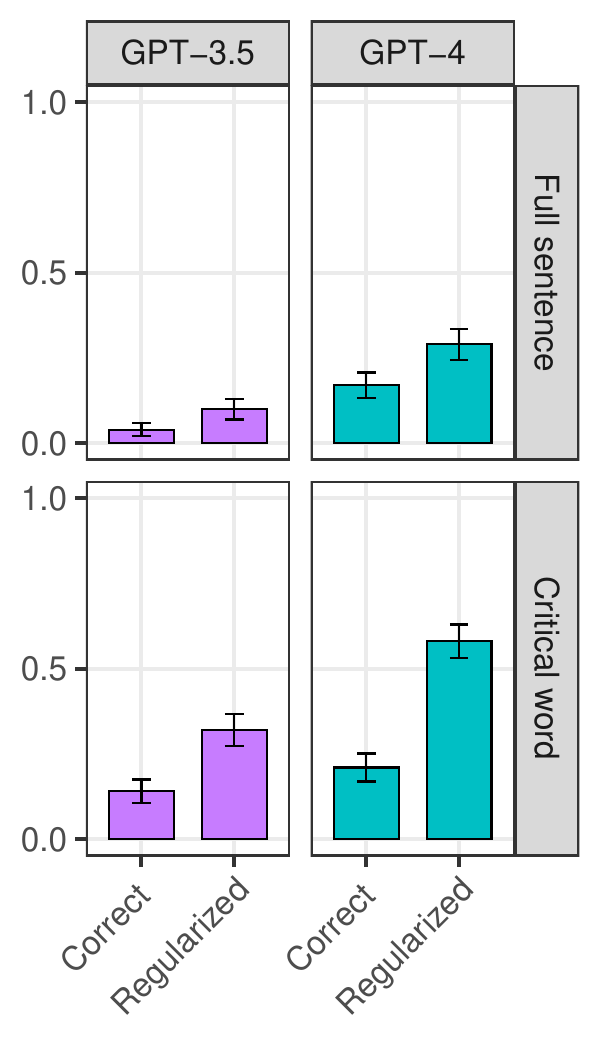}
    \end{minipage}

    \caption{Shift cipher performance on locally modified sentences. \textbf{Left:} Examples of the stimuli. In each case, the correct answer contained one word (underlined) that was implausible in context but that was similar to a more plausible word; e.g., in the first example, the correct word \textit{thanking} is similar to the incorrect but more plausible word \textit{thinking}. We refer to the incorrect but plausible options as \textit{regularized}. \textbf{Right:} The plots show how often model outputs matched the correct or regularized answer, either entirely (\textit{Full sentence}) or at the \textit{critical word} (e.g., \textit{thanking} vs.\ \textit{thinking}). Adjacent bars do not sum to 1.0 because models might produce an answer that is not the same as either option we consider. Error bars show one standard error.}
    \label{fig:adversarial}
\end{figure}

Under the hypothesis that models favor high-probability answers, we predict that they will ``regularize'' their outputs for these cases by producing the high-probability sentence that is similar to the correct answer (e.g., \ref{ex:adversariala}) rather than the correct answer (e.g., \ref{ex:adversarialb}).
We evaluated both whether the full output sentence matched the correct or regularized answer, and also whether just the critical word matched the correct word or the regularized version. 
Under both metrics, both models produced the regularized version more often than the correct version (Figure~\ref{fig:adversarial}). 
Here are some cases where GPT-4 produced an incorrect regularized answer:

\ex.
\a. \textbf{Correct output:} Because of this, their names were \underline{chanted} for security reasons.
\b. \textbf{GPT-4 output:} Because of this, their names were \underline{changed} for security reasons.

\ex.
\a. \textbf{Correct output:} Are they now building a bridge of their \underline{owl}?
\b. \textbf{GPT-4 output:} Are they now building a bridge of their \underline{own}?

\ex.
\a. \textbf{Correct output:} We are very careful about the voices we want to address, so we need to collaborate with \underline{otters} for sustainable impact.
\b. \textbf{GPT-4 output:} We are very careful about the voices we want to address, so we need to collaborate with \underline{others} for sustainable impact.

\ex.
\a. \textbf{Correct output:} This is a great achievement for us and a very \underline{beige} responsibility.
\b. \textbf{GPT-4 output:} This is a great achievement for us and a very \underline{big} responsibility.

A related type of regularization was shown in \citet{mckenzie2023inverse}, whose ``resisting correction'' results showed that models, when asked to repeat back a sentence containing a typo, often fixed the typo. Our findings show that such modifications can occur even when the sentence could be correct as written but is simply low-probability.

These results provide further evidence for the conclusion that LLMs are sensitive to the probability of the outputs that they produce. 
There are certainly contexts where such regularization is desirable, such as when it leads a model to fix typos. Therefore, we included this experiment not as a way to showcase model failure but rather as a way to highlight the types of strategies used by LLMs---strategies that may have desirable effects in some contexts but undesirable effects in others.

\paragraph{Further analysis of model errors:} For decoded rot-13 text, model predictions generally adhered fairly closely to the correct answer. Even when models made mistakes, they were not egregious, as in the following examples from rot-13 decoding:

\ex. 
\a. \textbf{Correct answer:} What are your hopes for the independent filmmaking scene in Indonesia?
\b. \textbf{GPT-4 output:} What are your hopes for the impending filmmaking scene in Indianapolis?

As the overall lower accuracy would suggest, model errors in rarer shifts deviate more from the correct answer. Consider rot-10: GPT-4's responses in this case have an average Levenshtein distance of 68.4, vs.\ 3.4 for rot-13. In such cases where the model output diverges dramatically from the correct answer, an interesting pattern emerges where the outputs are often recognizable sayings or quotations. For instance, among the rot-10 predictions are a popular motivational saying \ref{ex:repeatedly}\footnote{This saying originated from Will Durant but is often attributed to Aristotle.} and (a modified version of) a sentence famous for using all letters of the alphabet \ref{ex:quickbrownfox}:

\ex. \label{ex:repeatedly}
\a. \textbf{Correct answer:} We are very careful about the voices we want to address, so we need to collaborate with others for sustainable impact.
\b. \textbf{GPT-4 output:} We are what we repeatedly do. Excellence, then, is not an act, but a habit.

\ex. \label{ex:quickbrownfox}
\a. \textbf{Correct answer:} She never regretted her passion for the artistic craft, nor did she waver in her tireless dedication.
\b. \textbf{GPT-4 output:} The quick brown fox jumps over the lazy dog, but not the sheep in the background.

\noindent 
In another case, GPT-4 produced the start of a Shakespearean soliloquy \ref{ex:shakespeare}. Clearly something is rot-10 in the state of Denmark:

\ex. \label{ex:shakespeare}
\a. \textbf{Correct answer:} As a doctor of humanities, he was a university professor, founded a university and a newspaper, and won awards in journalism and literature.
\b. \textbf{GPT-4 output:} To be or not to be, that is the question, whether tis nobler in the mind to suffer the slings and arrows of outrageous fortune.

\noindent
This tendency to produce well-known sayings is consistent with the effect of output probability described in Section~\ref{sec:outputprobmotivation}. The less well the model has learned a task, the more uncertainty it will have about the relationship between output and input, which is what defines the likelihood. It will therefore be more influenced by the prior, resulting in predictions that are not very closely connected to the input but that have a high probability in absolute terms, such as well-known quotations.

\paragraph{Lack of separation between instructions and input:} One other type of model error is worth discussing as an illustration of how models rely on probability. Specifically, the models' outputs often include words relating to ciphers even when the correct answer does not. The following examples are all from GPT-4 decoding rot-2 text:

\ex. \label{ex:cipherwords1}
\a. \textbf{Correct answer:} This decision was influenced by the political climate of the time.
\b. \textbf{GPT-4 output:} This decryption was engineered by the programming language of the time.

\ex. 
\a. \textbf{Correct answer:} We could suggest things, but we didn't know how a city council worked.
\b. \textbf{GPT-4 output:} We could seriously damage, but we didn't know how a code breaker worked.

\ex. 
\a. \textbf{Correct answer:} The definition of a donor was also narrowed, making it difficult for foreign donors to fit this requirement.
\b. \textbf{GPT-4 output:} The decryption of a cipher was also designed, making it easier for other people to get their messages to go this direction.

The previous sections showed ways in which models are sensitive to the unconditional probability of their output. However, these new examples cannot be explained in that way; e.g., in \ref{ex:cipherwords1}, the model-produced word \textit{decryption} has a much lower unconditional probability than the correct word \textit{decision}. 
Instead, these examples show models relying on the input but in an incorrect way: the input includes directions that talk about a cipher (using words such as \textit{cipher} and \textit{decode}), and this makes models likely to produce related words in their output, even when the output should not be influenced by the instructions in this way. 

\subsection{Reversal}

\paragraph{Summary:} GPT-4 performed much better at reversing a list of words when the target output was a high-probability sentence (yielding an accuracy of 0.97) than when it was a low-probability sentence (yielding an accuracy of 0.53).

\paragraph{Details:} We gave models a sentence that had been reversed and asked them to reverse the order of the words (i.e., returning it to its original, grammatical order). We evaluated them on this task when the outputs were drawn from our high-probability, medium-probability, or low-probability set of sentences.
For both models, accuracy increased as output probability increased (Figure~\ref{fig:decoding_outputprob}, middle); logistic regressions (Appendix~\ref{app:revsigdec}) found that output probability had a highly significant effect for both GPT-3.5 ($p < 10^{-8}$) and GPT-4 ($p < 10^{-11}$).

\subsection{Pig Latin}

\paragraph{Summary:} When converting a Pig Latin sentence to English, GPT-4 had a much higher accuracy for examples with a high-probability output sentence (accuracy: 0.83) than examples with a low-probability output sentence (accuracy: 0.28).

\paragraph{Details:} We gave models a sentence written in Pig Latin and asked them to convert it to English. We evaluated models on examples where the output sentences were high-probability, medium-probability, or low-probability. For both models, accuracy increased as output probability increased (Figure~\ref{fig:decoding_outputprob}, bottom); logistic regressions (Appendix~\ref{app:pigsigdec}) found that output probability had a highly significant effect for both GPT-3.5 ($p < 10^{-11}$) and GPT-4 ($p < 10^{-14}$).

\subsection{Article swapping}\label{sec:article_swap_outputprob}

\begin{figure}
    \centering
    \begin{minipage}[l]{0.68\textwidth}
    \centering
    \begin{AIbox}{Article swapping: Output probability}{367pt}
    \parbox[t]{\linewidth}{

    {\small \underline{\textbf{Article swapping: Example of high-probability output}}}
    
    \hspace*{-0.1cm}
    \resizebox{360pt}{!}{
    \begin{tabular}{@{}c@{~}l@{~~}p{10.8cm}}
        & {\small \textbf{Input:}} & \sethlcolor{white}{\footnotesize \texttt{\hl{She distributed items the around room the and we tried them on.}}} \\
        & {\small \textbf{Correct:}} & \sethlcolor{white}{\footnotesize \texttt{\hl{She distributed the items around the room and we tried them on.}}}\\
        \textcolor{checkgreen}{\textbf{\checkmark}} & {\small \textbf{GPT-4:}} & \sethlcolor{aigreen}{\footnotesize \texttt{\hl{She distributed the items around the room and we tried them on.}}}
    \end{tabular}
    }

    \bigskip
    
    {\small \underline{\textbf{Article swapping: example of medium-probability output}}}

    \hspace*{-0.1cm}
    \resizebox{360pt}{!}{
    \begin{tabular}{@{}c@{~}l@{~~}p{10.8cm}}
        & {\small \textbf{Input:}} & \sethlcolor{white}{\footnotesize \texttt{\hl{Hal distributed props the across park the and he takes more in.}}} \\
        & {\small \textbf{Correct:}} & \sethlcolor{white}{\footnotesize \texttt{\hl{Hal distributed the props across the park and he takes more in.}}}\\
        \redx{} & \textbf{GPT-4:} & \sethlcolor{aigreen}{\footnotesize \texttt{\hl{Hal distributed the props across the park and he takes }\sethlcolor{pink}\hl{in more}\sethlcolor{aigreen}\hl{.}}}
    \end{tabular}
    }

        \bigskip
    
    {\small \underline{\textbf{Article swapping: example of low-probability output}}}

    \hspace*{-0.1cm}
    \resizebox{360pt}{!}{
    \begin{tabular}{@{}c@{~}l@{~~}p{10.8cm}}
        & {\small \textbf{Input:}} & \sethlcolor{white}{\footnotesize \texttt{\hl{Hal more he the props and the across distributed takes park in.}}} \\
        & {\small \textbf{Correct:}} & \sethlcolor{white}{\footnotesize \texttt{\hl{Hal more the he props the and across distributed takes park in.}}}\\
        \redx{} & \textbf{GPT-4:} & \sethlcolor{aigreen}{\footnotesize \texttt{\hl{Hal }\sethlcolor{pink}\hl{takes the more props and distributed them across the}\sethlcolor{aigreen}\hl{ park in.}}}
    \end{tabular}
    }
    }
    \end{AIbox}
    \end{minipage}
    \hspace{1mm}
    \begin{minipage}[r]{0.255\textwidth}
    \centering
    \vspace{0.7cm}
    
    \includegraphics[scale=0.5]{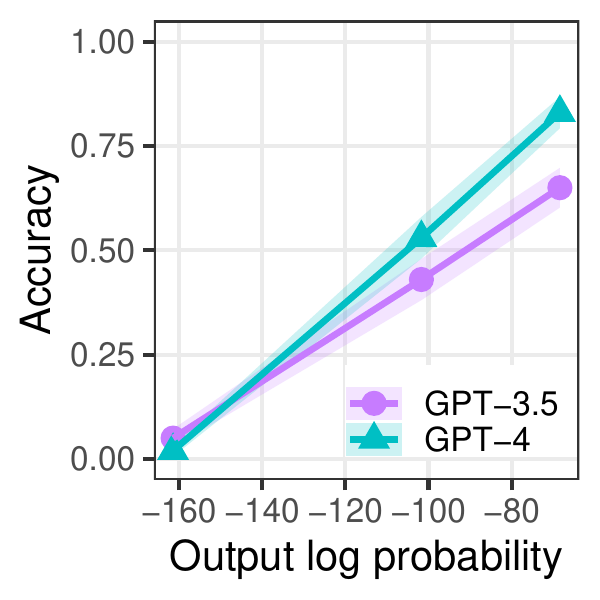}
    \end{minipage}

    \caption{\textbf{Left:} Examples of the article swapping task. \textbf{Right:} The effect of output probability on model accuracy on this task. The intervals around the lines show one standard error. Due to collinearity, we cannot tell with certainty whether the accuracy differences are driven by output probability or input probability, but based on the results of an additional set of experiments we have concluded that the results are most likely driven by output probability; see Appendix~\ref{app:article_swap_expanded_stats}.}
    \label{fig:swap_outputprob}
\end{figure}

\paragraph{Summary:} When given a sentence and asked to swap each article (\textit{a}, \textit{an}, or \textit{the}) with the preceding word, GPT-4 performed significantly better when the output sentence was a high-probability sentence (accuracy: 0.83) than when it was a random sequence of words (accuracy: 0.02).

\paragraph{Details:} We gave models a sequence that had been created by starting with a sentence and swapping each article with the following word, where articles are the words \textit{a}, \textit{an}, and \textit{the}; see Figure~\ref{fig:swap_outputprob} for examples. The model was then asked to swap each article with the \textit{preceding} word, returning the sentence to its original form. We evaluated models on cases where the output sentence was high-probability, medium-probability, or low-probability. For both models, accuracy increased substantially as output probability increased (Figure~\ref{fig:swap_outputprob}). For this task, input probability and output probability are highly correlated, likely because the input and output sequences have extensive overlap; thus, due to collinearity, we were unable to confidently interpret a logistic regression that factored in both input probability and output probability. When we ran a test that included only output probability (intending it to stand for general example probability, since the collinearity makes output probability a reasonable proxy for input probability as well), probability had a highly significant effect for both GPT-3.5 ($p < 10^{-11}$) and GPT-4 ($p < 10^{-15}$); see Appendix~\ref{app:articleswap_basic_stats}. Thus, it is clear that accuracy is connected to some aspect of example probability. 

Although we cannot confidently determine whether this effect is attributable mainly to input probability or output probability (or both), we ran an expanded set of experiments which provide evidence that output probability is the most important factor; see Appendix~\ref{app:article_swap_expanded_stats} for details of this analysis. For this reason, we group our results for this task under the heading of output probability rather than input probability.

\subsection{Acronyms}\label{sec:acronyms_output_prob}

\paragraph{Summary:} When asked to form an acronym from a sequence of words, GPT-4 performs better when the acronym that is produced is a common word (accuracy: 0.76) than when it is a rare word (accuracy: 0.67). 

\paragraph{Details:} Models were given a sequence of words and asked to concatenate the first letters of all the words; e.g., given \textit{lamprey insular turnkey esquire realist acetone luckier}, produce \textit{LITERAL}. 
In all conditions, all input words and output words were seven letters long and two subword tokens long (as measured by the tokenizer used by both GPT-3.5 and GPT-4). The input words were in all lowercase letters, and models were instructed to produce their output in capital letters.

The set of possible words was created by extracting all 2-token 7-letter words from the CMU Pronouncing Dictionary\footnote{\url{http://www.speech.cs.cmu.edu/cgi-bin/cmudict}} (used in this case as a wordlist, not for any phonetic purpose). The words were then ranked by the log probability assigned to them by GPT-2 \citep{radford2019language}. This process was repeated separately for lowercase words and words written in all capital letters, creating a list of 8920 lowercase words and 2303 all-capitalized words (the numbers differ due to differing tokenization). Each set was then split into five probability bins. We had five conditions, corresponding to the five probability bins for the output; across all conditions, we used only input words from the highest-probability bin. Thus, the input probability was held roughly constant, while the output probability varied.

\begin{figure}
    \centering
    \begin{minipage}[l]{0.42\textwidth}
    \centering
    \begin{AIbox}{Acronyms: Output probability}{230pt}
    \parbox[t]{\linewidth}{

    {\small \underline{\textbf{Example of high-probability output}}}
    
    \begin{tabular}{@{}c@{~}l@{~~}p{6cm}}
        & {\small \textbf{Input:}} & \sethlcolor{white}{\footnotesize \texttt{\hl{salient tribune outlast parades panache emerald dweller}}} \\
        & {\small \textbf{Correct:}} & \sethlcolor{white}{\footnotesize \texttt{\hl{STOPPED}}}\\
        \textcolor{checkgreen}{\textbf{\checkmark}} & {\small \textbf{GPT-4:}} & \sethlcolor{aigreen}{\footnotesize \texttt{\hl{STOPPED}}}
    \end{tabular}

    \bigskip

    {\small \underline{\textbf{Example of low-probability output}}}

    \begin{tabular}{@{}c@{~}l@{~~}p{6cm}}
        & {\small \textbf{Input:}} & \sethlcolor{white}{\footnotesize \texttt{\hl{gremlin affixed manatee empress lasagna impasse naively}}} \\
        & {\small \textbf{Correct:}} & \sethlcolor{white}{\footnotesize \texttt{\hl{GAMELIN}}}\\
        \redx{} & \textbf{GPT-4:} & \sethlcolor{pink}{\footnotesize \texttt{\hl{GAMENIL}}}
    \end{tabular}
    }
    \end{AIbox}
    \end{minipage}
    \hspace{0.5cm}
    \begin{minipage}[l]{0.33\textwidth}
    \includegraphics[scale=0.6]{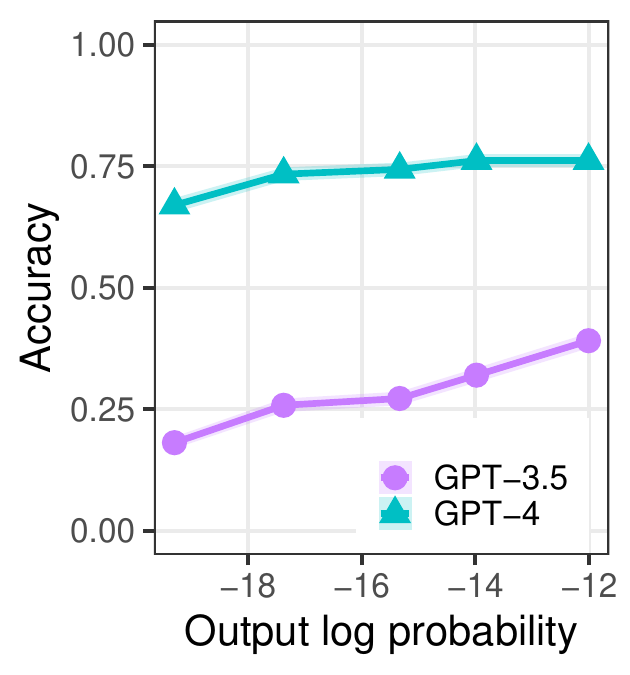}
    \end{minipage}
    \caption{Effect of output probability on acronym performance. The target output was always a real word that appeared in the CMU Pronouncing Dictionary, but the commonness of that word varied; the horizontal axis shows different output probability levels (e.g., \textit{STOPPED} is a common word, and \textit{GAMELIN} is a rare word).}
    \label{fig:acronyms_varyoutp}
\end{figure}

Both models performed better on conditions with high output probability than ones with low output probability (Figure~\ref{fig:acronyms_varyoutp}): GPT-3.5 had an accuracy of 0.39 in the highest-probability case, compared to 0.18 in the lowest-probability case, and for GPT-4 the corresponding numbers were 0.76 and 0.67. The effect of output probability is highly statistically significant in both cases, even though the effect size is modest for GPT-4 ($p < 10^{-15}$ for GPT-3.5, $p < 10^{-9}$ for GPT-4; see Appendix~\ref{app:acronym_significance}).

\begin{figure}
    \centering
    {\hspace*{-0.9cm}
    \begin{minipage}[l]{0.35\textwidth}
    \begin{AIbox}{Counting}{200pt}
    \parbox[t]{\linewidth}{
    {
    {\small \textbf{Counting characters:} 

    \noindent
    How many letters are in the list?}
    }

    \bigskip
    
    \hspace*{-0cm}
    \begin{tabular}{@{}c@{~}l@{~~}p{3.8cm}}
        & {\small \textbf{Input:}} & \sethlcolor{white}{\footnotesize \texttt{\hl{yyyyyyyyyyyyyyyyyyyy}}} \\
        & {\small \textbf{Correct:}} & \sethlcolor{white}{\footnotesize \texttt{\hl{20}}} \\
        \textcolor{checkgreen}{\textbf{\checkmark}} & {\small \textbf{GPT-4:}} & \sethlcolor{aigreen}{\footnotesize \texttt{\hl{20}}} \\ \\
        
        & {\small \textbf{Input:}} & \sethlcolor{white}{\footnotesize \texttt{\hl{yyyyyyyyyyyyyyyyyy}}} \\
        & {\small \textbf{Correct:}} & \sethlcolor{white}{\footnotesize \texttt{\hl{18}}}\\
        \redx{} & {\small \textbf{GPT-4:}} & \sethlcolor{pink}{\footnotesize \texttt{\hl{20}}}
    \end{tabular}

    }
    \end{AIbox}
    \end{minipage}
    }
    \hspace{7mm}
    \begin{minipage}{0.3\textwidth}

    \vspace{3mm}
    
    \includegraphics[scale=0.55]{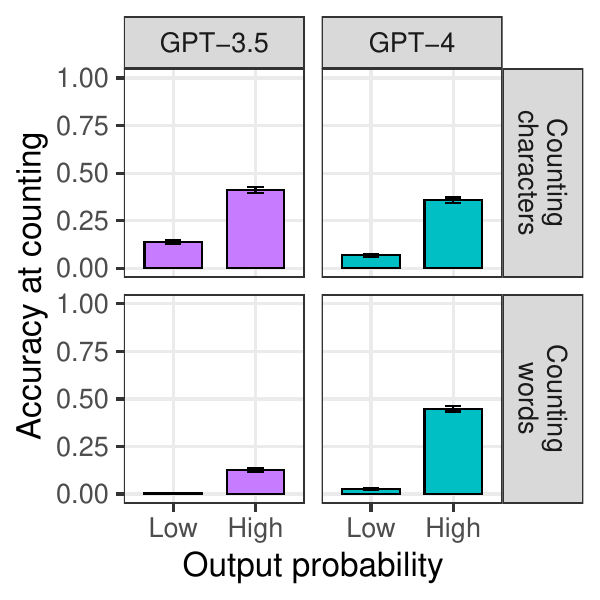}
    \end{minipage}

    \vspace{3mm}
    
    \includegraphics[scale=0.55]{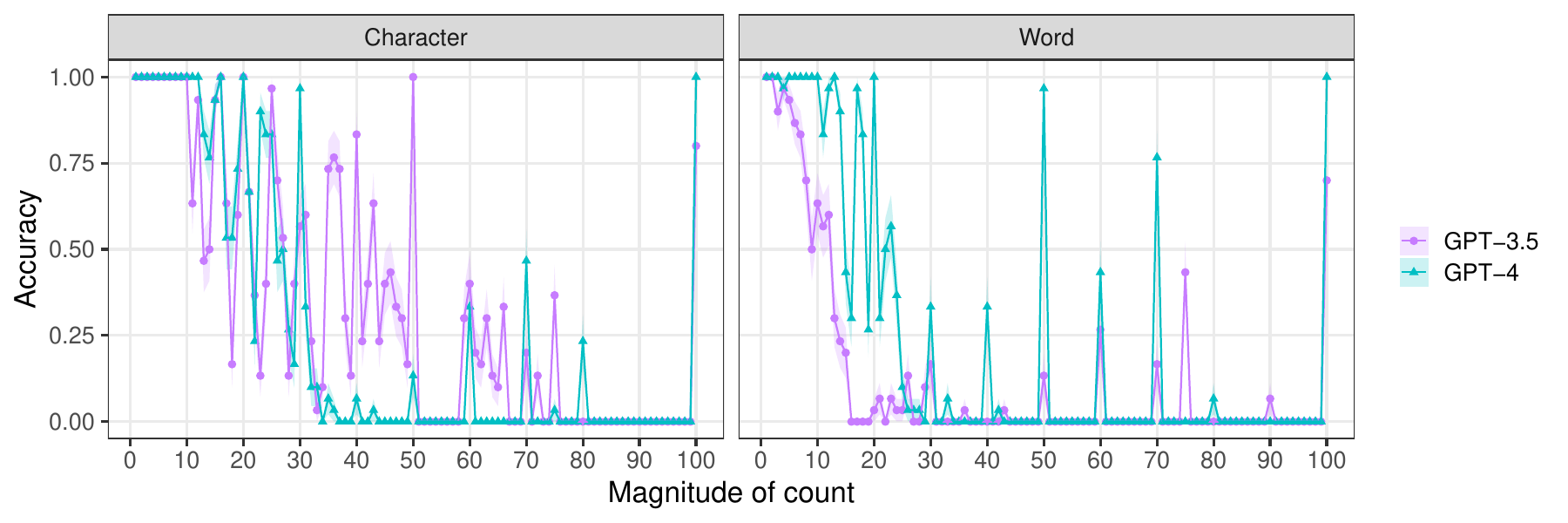}
    \includegraphics[scale=0.55]{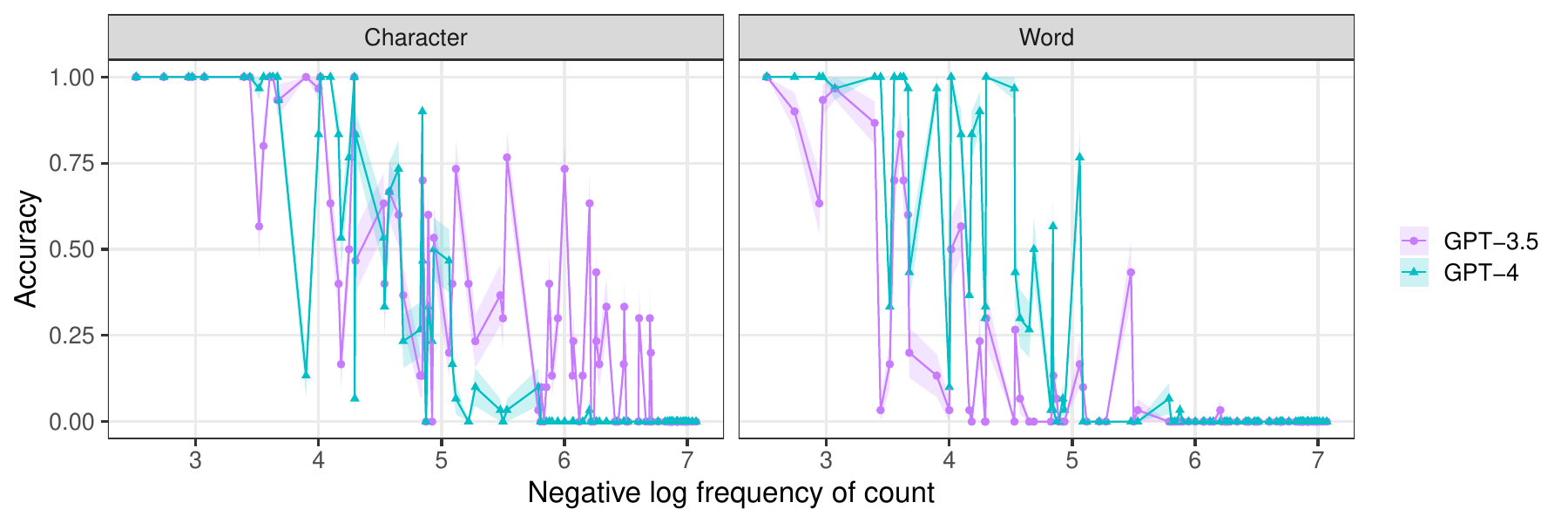}
    \caption{\textbf{Top left:} Examples of stimuli for counting characters; the stimuli for counting words are similar except that the characters being counted are all the same within an example, but the words being counted are all different. \textbf{Top right:} Counting accuracy in settings with low-probability and high-probability outputs. In this context, the output is a number (the number of elements in the list), and \textit{output probability} refers to the frequency of that number in an Internet corpus. Error bars show one standard error. \textbf{Middle and bottom:} Counting performance by the magnitude (middle) or frequency (bottom) of the number being counted to. We use the C4 corpus \citep{raffel2020exploring} to estimate each number's frequency. The intervals around the line show one standard error.}
    \label{fig:counting_freq_magnitude}
\end{figure}

\subsection{Counting}\label{sec:counting_outputprob}

\paragraph{Summary:} When asked to count how many characters or words are in a list, GPT-4 achieves a much higher accuracy when the correct count is a commonly-used number (e.g., 100) than when the correct count is a more rarely used number (e.g., 83).

\paragraph{Details:} We presented models with a list of either characters or words. The list length varied from 1 to 100; we showed 30 examples of each length and then computed an accuracy for each number based on those 30 examples. Figure~\ref{fig:counting_freq_magnitude} plots model accuracy as a function of the magnitude of the number being counted to (middle) or the log frequency of the number being counted to (bottom), where the frequency was estimated based on how often each number appears in the first 1 billion words in the C4 corpus \citep{raffel2020exploring}. 
The curve is smoother when based on frequency than when based on magnitude, a point made more formally by seeing that accuracy has a higher Spearman rank correlation with frequency than with magnitude (Table~\ref{tab:counting_correlations}). Finally, in logistic regressions, output log probability had a highly significant effect across conditions ($p < 10^{-15}$ in all cases; see Appendix~\ref{app:counting_significance}); output magnitude also had a significant effect in the character-counting cases ($p < 10^{-15}$ for GPT-3.5; $p < 10^{-14}$ for GPT-4) but not in the word-counting cases ($p = 0.56$ for GPT-3.5; $p = 0.07$ for GPT-4).

As another way to disentangle magnitude from frequency, we did a test that controlled for magnitude but varied frequency. We identified all numbers $n$ from 1 to 100 that were at least twice as frequent as $n-2$, $n-1$, $n+1$, or $n+2$. Such numbers were labeled high-probability, while their neighbors ($n-2$, $n-1$, $n+1$, and $n+2$) were labeled low-probability. We then evaluated models on 100 examples for each high-probability number and 25 examples for each low-probability number; average magnitude was thus matched because the mean of $n-2$, $n-1$, $n+1$, and $n+2$ is $n$. Accuracy was much higher for the high-probability outputs than the low-probability ones (Figure~\ref{fig:counting_freq_magnitude}, top right), and these differences were statistically significant (see Appendix~\ref{app:counting_significance}).

\citet{davis2023mathematics} and \citet{arkoudas2023gpt} also found that GPT-4 struggled with counting. Our experiment takes this result further by connecting it to probability, which is in turn connected to our broader teleological analysis of why models behave as they do.

\begin{table}[]
    \centering
    \begin{tabular}{ccc} \toprule
         & |cor(Accuracy, Magnitude)| & |cor(Accuracy, Frequency)| \\ \midrule
      GPT-3.5: Counting words & 0.57 & 0.74 \\
      GPT-4: Counting words & 0.70 & 0.84 \\ \midrule
      GPT-3.5: Counting characters & 0.81 & 0.84 \\
      GPT-4: Counting characters & 0.76 & 0.88 \\ \bottomrule
    \end{tabular}
    \caption{Investigating which features correlate with LLM accuracy at counting. The two features considered are the magnitude of the number being counted to and the frequency in a corpus of the number being counted to. The table shows the absolute value of the Spearman rank correlation coefficients between accuracy and each of these features. In all cases, accuracy correlates more strongly with frequency than with magnitude.}
    \label{tab:counting_correlations}
\end{table}

\subsection{Summary of output probability effects}

We have shown that the performance of GPT models is heavily influenced by the probability of the target output, even though the tasks being investigated were not inherently probabilistic. These results support our hypothesis that LLMs are sensitive to the probability of the sequences they must produce.

\section{Sensitivity to input probability}\label{sec:inputprob}

We have hypothesized that LLMs will sometimes perform worse on examples with low-probability inputs than ones with high-probability ones. However, we have also hypothesized that the influence of input probability will be less pervasive than the influence of output probability: we predicted that LLMs would use output probability in any scenario where they have some uncertainty about the output---a condition that we expect holds frequently---whereas we predicted that input probability would only matter in situations where an LLM's ability to process an input is highly dependent on prior experience with that specific input---a condition that we expect to hold less often under the view that, for many tasks, neural networks are not restricted to handling only the inputs they have seen but can also generalize across inputs. 
In this section, we investigate how model performance varies with input probability and compare these effects to the output probability effects observed in the previous section.

\begin{figure}[t]
    \centering
    \includegraphics[scale=0.5]{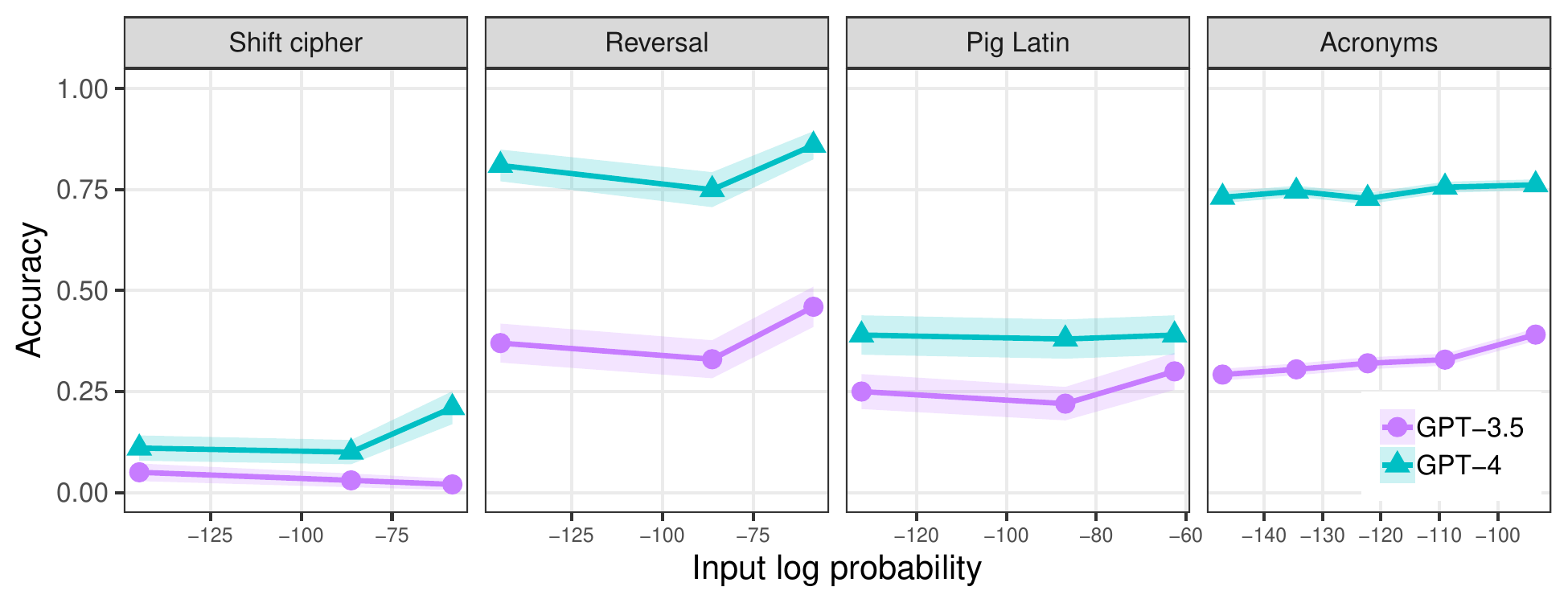}
    \caption{The effect of input probability on model accuracy across four tasks.}
    \label{fig:encoding_inputprob}
\end{figure}

\begin{table}[t]
    \centering
    \begin{tabular}{ccc} \toprule
    & Accuracy & Mean edit distance \\ \midrule
    GPT-3.5: High-probability & 0.02 & 24.2 \\
    GPT-3.5: Medium-probability & 0.03 & 24.0 \\
    GPT-3.5: Low-probability & 0.05 & 38.5\\ \midrule
    GPT-4: High-probability & 0.21 & 7.9 \\
    GPT-4: Medium-probability & 0.10 & 9.8 \\
    GPT-4: Low-probability & 0.11 & 10.4 \\ \bottomrule
    \end{tabular}
    \caption{Accuracy on encoding sentences using a simple cipher (rot-13). There were three groups of sentences of varying probability. The two metrics of success are accuracy (higher is better) and Levenshtein edit distance (lower is better).}
    \label{tab:probeffectsizes}
\end{table}

\subsection{Shift ciphers, reversal, Pig Latin, and acronyms}

\paragraph{Summary:} When encoding text with rot-13, GPT-4 achieves an accuracy of 21\% when the input sentence is high-probability but only 11\% when it is low-probability. For the other three tasks considered in this subsection (reversal, Pig Latin, and acronyms), input probability does not have a statistically significant effect for GPT-4.

\paragraph{Details:} To investigate input probability for shift ciphers, reversal, and Pig Latin, we reversed the setup used in Section~\ref{sec:outputprob}: instead of asking LLMs to decode sentences that have been manipulated through these three methods, we asked them to start with English text and encode it using rot-13, reversal, or Pig Latin. Thus, our three different sets of sentences now gave us three levels of input probability rather than three levels of output probability. 

For acronyms, we did not need to change the task framing from before, but now instead of manipulating the probability of the word that is produced as an acronym, we manipulated the probability of the words from which the acronym is formed by selecting inputs from each of the five probability bins described in Section~\ref{sec:acronyms_output_prob}; for the target outputs, we used only words from the highest-probability bin. Below are examples of a high-probability input and a low-probability input, both of which have \textit{TOASTER} as the target output:

\ex. \label{ex:acronym_inputprob}
\a. \textbf{High-probability input:} trundle octopus archaic stooped tabling esquire retrace
\b. \textbf{Low-probability input:} trapelo ottaway amiably scholle troiani esqueda royster

\noindent
The results are shown in Figure~\ref{fig:encoding_inputprob}. There were two cases where our logistic regressions found that input probability had a statistically significant effect on model performance: GPT-4 processing the shift cipher ($p < 0.05$) and GPT-3.5 forming acronyms ($p < 10^{-5}$). For the shift cipher, we noted that the lack of a significant effect for GPT-3.5 might be due to floor effects since its accuracy was close to zero across input probability levels (Figure~\ref{fig:encoding_inputprob}, left), so for the shift cipher we also computed results with the more lenient measure of edit distance between the model output and the correct answer, which is shown in Table~\ref{tab:probeffectsizes}. A linear regression predicting this more sensitive metric found that input probability had a significant effect for both GPT-3.5 and GPT-4, with $p < 0.01$ for both models. See Appendices~\ref{app:shift_input_prob_stats}, \ref{app:revsigenc}, \ref{app:pigsigenc}, and \ref{app:acronyminpsig} for details of the regressions. 
These results provide an existence proof that the probability of the input can indeed influence LLM performance; the fact that it only had a significant effect in a few cases is consistent with our prediction that input probability will not have a reliable influence.

\subsection{Counting}

\paragraph{Summary:} Input probability shows inconsistent effects when the task is counting: it has a significant effect in the expected direction (better performance on high-probability inputs) for GPT-3.5 counting characters and GPT-4 counting words, but it has a significant effect in the opposite direction for GPT-3.5 counting words and GPT-4 counting characters; note, however, that the effect size was very small in all cases except for GPT-3.5 counting characters (which was one of the cases that was significant in the expected direction). 

\paragraph{Details:} In order to vary the input probability when having models count characters, we used letters as high-probability characters and emojis as low-probability characters. GPT-3.5 performed better on counting high-probability characters (letters) than low-probability ones (emojis), while GPT-4 had a small but statistically significant effect in the opposite direction (Figure~\ref{fig:counting_inp}, top); see Appendix~\ref{app:counting_significance} for details of the significance tests. 
To modulate input probability when counting words, we estimated word probability using GPT-2 (as we did for acronyms) and then sorted the vocabulary by probability. We selected 450 words from the higher-probability half and 450 words from the lower-probability half to serve as our two types of words. In this case, GPT-4 performed better on counting high-probability words than low-probability ones, whereas GPT-3.5 unexpectedly performed better on counting low-probability words  than high-probability ones (Figure~\ref{fig:counting_inp}, bottom), though in both cases the effect size was small. 
It is unclear why two of the cases show effects in the direction that is opposite from what we expected. 
Note however that both of these cases had small effect sizes (Appendix~\ref{app:counting_significance}). 

\begin{figure}
    \centering
    \includegraphics[scale=0.6]{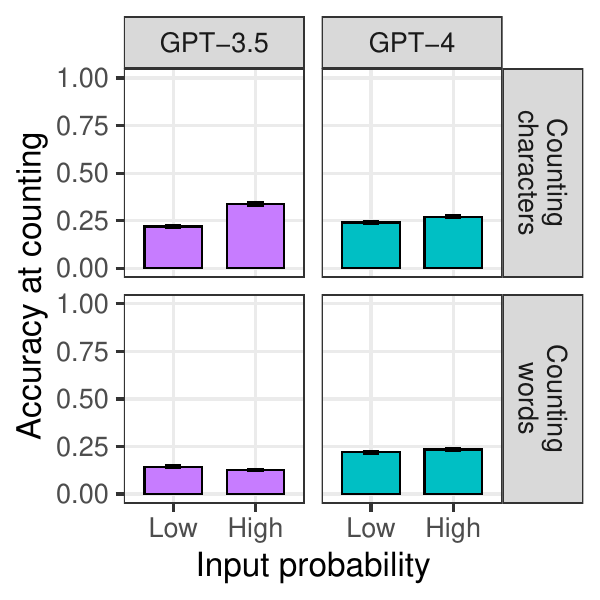}
    \caption{Accuracy on counting when input probability is modulated}
    \label{fig:counting_inp}
\end{figure}

\subsection{Asymmetry between input probability and output probability}

\paragraph{Summary:} A comparison of our results in Sections~\ref{sec:outputprob} and \ref{sec:inputprob} supports the hypothesis that LLMs are more pervasively influenced by output probability than input probability.

\paragraph{Details:} Now that we have run experiments modulating output probability and input probability, we have the opportunity to compare these two factors to see if input probability is indeed less influential, as we have hypothesized. One immediate piece of evidence supporting this conclusion is that all of our output probability experiments yielded significant effects---often with large effect sizes---whereas only a few input probability conditions yielded significant effects. 

In addition to noting which experiments produced significant effects, we can also look for an input/output asymmetry by considering model performance in the high-probability rot-13 conditions. GPT-4 scores 0.51 for decoding rot-13 in the high-probability setting but only 0.21 for encoding rot-13 in the high-probability setting; GPT-3.5 scores 0.21 for decoding and 0.02 for encoding. This asymmetry between decoding performance and encoding performance supports the conclusion that output probability is more influential than input probability. The argument for why these results support this conclusion is as follows:

\begin{itemize}
    \item In a shift cipher, encoded text is usually a very low-probability string of letters.\footnote{There are rare exceptions where shifting text produces other recognizable text. For instance, the rot-13 encoding of \textit{vex} is its synonym \textit{irk}, the rot-13 encoding of \textit{abjurer} is \textit{nowhere}, and fans of \textit{2001: A Space Odyssey} have pointed out that the name of the computer HAL becomes IBM when given a shift of 1.} Therefore, encoding text involves turning a high-probability sequence into a low-probability one, while decoding text involves the opposite. (The fact that we want to ensure that the English text is high-probability is why, in the previous paragraph, we considered only the high-probability conditions within the encoding and decoding tasks).
    \item We have ample evidence to believe that low probability is a source of difficulty for LLMs. Therefore, we have one task (encoding) where the input is easy to process yet the output is hard to process, and one task (decoding) where the opposite is true.
    \item Whichever ``side'' (input vs.\ output) the LLM is more sensitive to is the one that the LLM's overall performance would pattern with, since the variations in difficulty in the less-influential side would be overwhelmed by the variations in difficulty in the more-influential side.
    \item Since LLM performance follows the difficulty of the output rather than the difficulty of the input (i.e., higher accuracy for decoding than encoding), we have evidence that LLMs are more sensitive to output probability than input probability. 
\end{itemize}
One potential concern with the previous argument is that it assumes that encoding and decoding have no inherent difference in complexity as tasks, such that any differences in performance can be attributed to the specific text rather than the task being performed. In the general case, it is true that a function and its inverse may not always be equally easy to compute; e.g., it is much harder to find the prime factors of a large number than it is to multiply together prime factors. However, rot-13 is a special case because it is its own inverse---that is, rot-13 encoding and rot-13 decoding are in fact the same function. 
It still remains possible that the standard ways to perform each task differ in difficulty: encoding for rot-$n$ is traditionally performed by moving forward $n$ positions in the alphabet, while decoding is traditionally performed by moving backward $n$ positions. However, if anything, we would intuitively expect that moving forward would be easier than moving backward because the alphabet is much more often said in forward order than reverse order. Therefore, if this factor does play a role it would be in the direction of making encoding easier than decoding.
In sum, we have no reason to believe that encoding is inherently a more complex task than decoding, justifying the conclusion that the difference in accuracy is driven by the properties of the input and output strings, not by differences in the mapping that must be performed. 

In addition to rot-13, we have one other task that is its own inverse: reversal. Here too models performed better on decoding than encoding: GPT-4 scored 0.97 for high-probability decoding but only 0.86 for high-probability encoding, and GPT-3.5 scored 0.74 for high-probability decoding but 0.46 for high-probability encoding. Therefore, the reversal results, like the rot-13 results, support our hypothesis that LLM accuracy is affected more strongly by output probability than input probability.

\subsection{Birthdays}

\paragraph{Summary:} When asked the birthday of a public figure, GPT-4 performed much better when the person had a high probability of being mentioned  online (accuracy: 0.99) than when they had a low probability of being mentioned (accuracy: 0.23). This task provides an example where input probability has a large effect (something that did not occur in any of our previous experiments that modulated input probability).

\paragraph{Details:} Our experiments with input probability have revealed some cases where it has a statistically significant effect, but in these situations the effect size was small. This observation naturally raises the question of whether input probability ever has a large effect on LLM performance.
To attempt to identify such a case, we return to our prediction about when input probability should matter: when there is some piece of information about the input that is necessary for solving the task but that models would not learn if the input is low probability. 
Building on this analysis, we now consider a task that, under this analysis, is expected to yield very large effects of input probability: being provided a public figure's name and returning that person's birth date. Here, the piece of information about the input that is relevant is the person's birth date. 
This information cannot be retrieved from the input alone, so the only way it can be produced is if the model has encountered it during training---which is much more likely to happen if the person is frequently discussed.
Therefore, models should show large effects of input probability on this task.

\begin{figure}
    \centering
    \begin{minipage}[l]{0.42\textwidth}
    \centering
    \begin{AIbox}{Birthdays: Input probability}{230pt}
    \parbox[t]{\linewidth}{

    {\small \underline{\textbf{Example with a high-frequency input}}}
    
    \begin{tabular}{@{}c@{~}l@{~~}p{6cm}}
        & {\small \textbf{Input:}} & \sethlcolor{white}{\footnotesize \texttt{\hl{Carrie Underwood}}} \\
        & {\small \textbf{Correct:}} & \sethlcolor{white}{\footnotesize \texttt{\hl{March 10, 1983}}}\\
        \textcolor{checkgreen}{\textbf{\checkmark}} & {\small \textbf{GPT-4:}} & \sethlcolor{aigreen}{\footnotesize \texttt{\hl{March 10, 1983}}}
    \end{tabular}

    \bigskip

    {\small \underline{\textbf{Example with a low-frequency input}}}

        \begin{tabular}{@{}c@{~}l@{~~}p{6cm}}
        & {\small \textbf{Input:}} & \sethlcolor{white}{\footnotesize \texttt{\hl{Jacques Hanegraaf}}} \\
        & {\small \textbf{Correct:}} & \sethlcolor{white}{\footnotesize \texttt{\hl{December 14, 1960}}}\\
        \redx{} & \textbf{GPT-4:} & \sethlcolor{pink}{\footnotesize \texttt{\hl{November 16, 1960}}}
    \end{tabular}
    }
    \end{AIbox}
    \end{minipage}
    \hspace{0.5cm}
    \begin{minipage}[l]{0.33\textwidth}
    \includegraphics[scale=0.55]{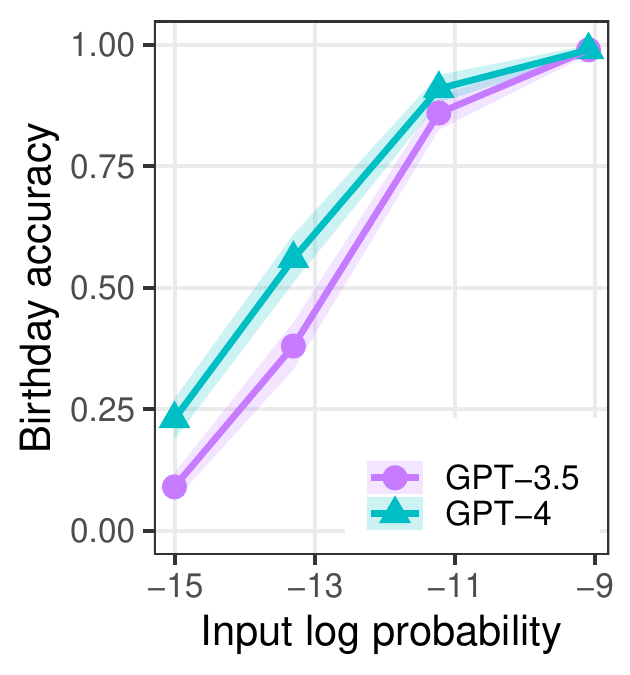}
    \end{minipage}
    \caption{\textbf{Left:} Example stimuli for asking models to provide the birthdays of public figures. \textbf{Right:} Model accuracy at providing a person's birthday, as a function of the probability of that person being mentioned in online text. The intervals around the lines show one standard error.}
    \label{fig:bdays}
\end{figure}

To test this prediction, we used the WikiBio dataset \citep{lebret2016wikibio} to create four sets of public figures, stratified by the frequency with which each person's name is mentioned in the C4 corpus; details are in Appendix~\ref{app:birthdayexampledetails}. For both models, accuracy varied substantially with input probability: GPT-4's accuracy ranged from 23\% in the lowest-probability condition to 99\% in the highest-probability condition, while GPT-3.5's accuracy ranged from 9\% to 99\% (Figure~\ref{fig:bdays}). The effect of input probability was highly significant for both models, with $p < 10^{-15}$ (Appendix~\ref{app:birthdaysig}).

These birthday results are an example of a finding that is obvious if one considers how LLMs work but potentially surprising otherwise. 
That is, when one interacts with an LLM, it is easy to imagine that it can search the Internet to find factual information. If LLMs indeed did that, then they should be able to get 100\% accuracy across the board on this birthday task, because all people in our dataset have their birthdays listed on Wikipedia.
In fact, however, the standard versions of GPT-3.5 and GPT-4 do not query external databases. Any factual information that they store is the result of them encountering that information during their training phase. Therefore, they can be expected to perform more robustly on pieces of information that have a high probability of occurring, and we have indeed found that they do so. 

Mapping names to birthdays is a task for which there is no systematic function that extracts the output from the input. This fact makes it unsurprising that LLM performance would be tied to what they have seen. Though this scenario is straightforward, we believe that it lends a window into a less obvious conclusion: even when there \textit{is} a systematic function that maps the input to the output, LLMs may not necessarily use it---they might still approach the task in a way akin to providing birthdays, where they memorize input-output mappings on a case-by-case basis rather than applying the function systematically.
We conjecture that, when LLMs are sensitive to input probability, it is because they (at least partially) use such a strategy in that situation. For example, the fact that we have found GPT-4 to be sensitive to input probability when encoding text into rot-13 suggests that it may not have learned the rot-13 function systematically but may have instead memorized the rot-13 versions of the specific words that it has seen written in this cipher.

\subsection{Summary of input probability effects}

Consistent with our hypotheses, we have found that input probability had a smaller impact on LLM performance than output probability for the tasks we have investigated. Though input probability often had little or no effect, we identified one case (namely, the birthday task) where it substantially influences accuracy, as well as a few cases (such as rot-13 encoding) where it has a moderate but statistically significant effect. 
Some additional examples of situations where input probability has a large effect can be found in the work of \citet{razeghi2022impact}, who showed that LLMs performed much better on numerical reasoning tasks (e.g., arithmetic questions) when the numbers in the question are common in pretraining corpora than when they are rare---a finding that is complementary to our results because, in the arithmetic tasks that we consider (linear functions and multiplication) we only assess the effects of task frequency (Section~\ref{sec:taskfreqlinear}), output probability (Section~\ref{sec:counting_outputprob}),  and question formatting (Section~\ref{sec:wording}).
Overall, then, we conclude that, under the right conditions, LLM performance can indeed be meaningfully affected by input probability.

\section{Other phenomena}

Beyond the probability-sensitivity that we have documented above, there are many other ways in which the problem that LLMs solve influences their performance. In this section we describe two more of these properties and illustrate them with experiments. In the following section, we discuss several additional properties that have been identified in prior work.

\subsection{Lack of embodiment}

There are some tasks that humans easily solve by relying on our presence in the physical world---e.g., by using information from sensory input, or by offloading work onto external tools and resources.
However, LLMs do not have bodies and do not navigate the physical world. All of their experience is in the form of text tokens. Therefore, LLMs can be expected to struggle on the tasks that are most straightforwardly solved by leveraging one's embodiment.

\subsubsection{Keyboard cipher}

\paragraph{Summary:} When asked to replace every letter in a sentence with the letter to the right of it on a QWERTY keyboard, GPT-4 scores 0\%.

\begin{table}[]
    \centering
    \begin{tabular}{ccccc} \toprule
         & \multicolumn{2}{c}{GPT-3.5} & \multicolumn{2}{c}{GPT-4} \\
         & Accuracy & Edit distance & Accuracy & Edit distance \\ \midrule
        Basic prompt & 0.00 & 21.5 & 0.00 & 18.0 \\
        Keyboard provided & 0.00 & 17.5 & 0.01 & 8.0 \\
        Correspondences provided & 0.00 & 51.5 & 0.12 & 4.0 \\ \bottomrule
    \end{tabular}
    \caption{Performance on the keyboard cipher. For accuracy, higher is better; for edit distance, lower is better. The edit distances are medians to address the presence of outliers in the GPT-3.5 data.}
    \label{tab:keyboard_results}
\end{table}

\begin{figure}[t!h!]
    \centering
    \begin{AIbox}{Keyboard cipher}{0.9\textwidth}
    \parbox[t]{\linewidth}{

    {\small \underline{\textbf{Prompt:}}}

    {\small
The keyboard cipher is a cipher where each letter is replaced with the letter to the right of it on a QWERTY keyboard. If the letter does not have a letter to its right, then you should wrap around to the start of the row; e.g., ``p'' maps to ``q''. For example, here is a message to be encoded:

\noindent
Original message: ``Hello world!''

\bigskip
\noindent
To encode this message, we replace each letter with the one to the right of it on a QWERTY keyboard:

\noindent
1. The first word is ``Hello''

\noindent
\hspace{0.1cm} * H -> J
 
\noindent
\hspace{0.1cm}  * e -> r
 
\noindent
\hspace{0.1cm}  * l -> a
 
\noindent
\hspace{0.1cm}  * l -> a
 
\noindent
\hspace{0.1cm}  * o -> p
 
\noindent
\hspace{0.1cm}  So ``Hello'' becomes ``Jraap''
 
\noindent
2. The second word is ``world!''

\noindent
\hspace{0.1cm}  * w -> e
 
\noindent
\hspace{0.1cm}  * o -> p
 
\noindent
\hspace{0.1cm}  * r -> t
 
\noindent
\hspace{0.1cm}  * l -> a
 
\noindent
\hspace{0.1cm}  * d -> f
 
\noindent
\hspace{0.1cm}  * ! -> !
 
\noindent
\hspace{0.1cm}  So ``world!'' becomes ``eptaf!''
 
\bigskip 
\noindent
Therefore, the message in the keyboard cipher is: ``Jraap eptaf!''

\bigskip 
\noindent
Here is another message. Encode the message in the keyboard cipher, word by word and letter by letter. On the last line, write the words ``Keyboard cipher:'' followed by the encoded message:

\noindent
Original message: ``But this time, there may also be another reason.''}

\bigskip
\noindent
{\small \underline{\textbf{Correct answer:}}}

\noindent
{\small Niy yjod yozr, yjrtr zsu sadp nr smpyjrt trsdpm.}

\bigskip
\noindent
{\small \underline{\textbf{GPT-4 output:}}}

\noindent
{\small \sethlcolor{aigreen}\hl{Niy }\sethlcolor{pink}\hl{yjoa yo,r yjtrr ,su saap}\sethlcolor{aigreen}\hl{ nr }\sethlcolor{pink}\hl{sbpyjrt trsapb}\sethlcolor{aigreen}\hl{.}}

    }
    \end{AIbox}
    \caption{The chain-of-thought prompt that we use to ask models to apply the keyboard cipher.}
    \label{fig:keyboard}
\end{figure}

\paragraph{Details:} The keyboard cipher is a cipher in which each letter is replaced with the letter to the right of it on a QWERTY keyboard (wrapping around if necessary); e.g., \textit{Good morning} becomes \textit{Hppf zptmomh}. A human with access to a keyboard can easily apply this cipher by referring to the keyboard. However, LLMs do not inhabit the physical world and therefore do not use a keyboard to produce their answers, so we therefore expect for them to struggle at this task. To test this prediction, we asked GPT-3.5 and GPT-4 to encode our 100 high-probability sentences (Section~\ref{sec:sentenceprobs}) into the keyboard cipher. For this experiment, we used a chain-of-thought prompt (shown in Figure~\ref{fig:keyboard}) because our prediction is an absolute one rather than a relative one and therefore merits testing models in the setting most favorable to them. As predicted, GPT-3.5 and GPT-4 both scored 0\% on this task (Table~\ref{tab:keyboard_results}, ``basic prompt'').

To some extent, this result may seem obvious---it may not be surprising that GPT models cannot perform this task without a keyboard in front of them. However, if we modify their prompt to include a textual representation of the keyboard layout (shown in \ref{ex:qwerty} below; see Appendix~\ref{app:promptkeyboardprovided} for the full prompt), their performance improves but is still flawed (Table~\ref{tab:keyboard_results}, ``keyboard provided''). For example, GPT-4's edit distance is cut approximately in half, meaning that it makes only half as many errors, but the edit distance remains relatively high (8.0, meaning that it typically makes 8 mistakes per sentence). This result shows that the models' failure is not merely due to a lack of information; even with this information provided, they are still not able to reliably use it properly---a difficulty that we conjecture can be tied back to the lack of embodiment: because they do not inhabit the physical world, they lack experience in navigating spatial arrangements such as the provided keyboard layout. 

\ex. \label{ex:qwerty} q w e r t y u i o p \\ a s d f g h j k l \\ z x c v b n m

When we instead modify the prompt to specify exactly what each letter maps to (\textit{a -> s, b -> n, c -> v, ...}), we see another jump in performance for GPT-4 (Table~\ref{tab:keyboard_results}, ``correspondences provided''), supporting the argument that one reason for GPT-4's difficulty is an inability to extract the necessary information from an array of letters. Thus, when we provide this information directly instead of requiring GPT-4 to extract it from an array of letters, its performance increases. Even in this final case, GPT-4's accuracy is far from 100\%; the remaining errors may be explained by difficulty with systematically manipulating symbols (see Section~\ref{sec:systematicity}). In contrast to GPT-4, GPT-3.5 performs worse in the ``Correspondences provided'' condition than in the other conditions. It is not clear why this drop in performance occurs. One possible explanation is that the prompt is much longer in this condition than in the others, so GPT-3.5 might have trouble sorting through all of the provided information.

Note that embodiment is not strictly necessary to impart the information required for this task: it can in principle be learned from textual input. Indeed, even in the ``basic prompt'' condition where models were not provided with a keyboard, they get many letters right: if we assumed that they only got correct the letters whose correspondences were illustrated in the prompt, and guessed randomly on all others, their edit distances would be 37.0, but they in fact display much lower edit distances (Table~\ref{tab:keyboard_results}). Thus, models appear to have learned some information about keyboard structure from their training data, just not enough to perform the task well. More generally, then, we are not claiming that text-only models could never acquire the information for which humans typically rely on embodiment. Rather, what we are claiming is that such information is difficult but not impossible for non-embodied models to acquire.

\subsection{Sensitivity to wording}\label{sec:wording}

Because LLMs are explicitly trained on language but not on ideas, we expect them to be sensitive to the linguistic form of the ideas they encounter: they may yield varying results when given the same idea expressed in different ways. 
Prior work has identified several ways in which LLMs are sensitive to task-irrelevant changes in wording: they may change their response based on the addition of irrelevant information \citep{kassner2020negated,shi2023large}, based on textual perturbations such as lowercasing text or introducing typos \citep{liang2022holistic}, based on the ordering of information in the prompt \citep{liu2023lost,zhao2021calibrate}, or based on the way in which the prompt is phrased \citep{zhao2021calibrate}. 
In the opposite direction, LLMs are sometimes \textit{less} sensitive to the prompt than they should be, displaying successful performance even when prompts are irrelevant or misleading \citep{webson2022prompt,min2022rethinking}; this result further illustrates how models do not understand their instructions in the same ways that humans likely do but instead rely on some unintuitive statistical cues relating to the way the prompt is worded.
Below, we provide two further experiments showing sensitivity to prompt wording (Section~\ref{sec:promptwording}) and formatting (Section~\ref{sec:multiplication}).

\subsubsection{Linear function}\label{sec:promptwording}

\paragraph{Summary:} GPT-4 performs significantly better at applying the function $f(x) = (9/5)x + 32$ when the prompt mentions that this function is the Celsius-Fahrenheit conversion than when it does not.

\begin{figure}[t]
    \centering

    \begin{minipage}[l]{0.57\textwidth}
    \centering
    \begin{AIbox}{Linear function: Prompt variants}{315pt}
    \parbox[t]{\linewidth}{

    {\small \underline{\textbf{Basic prompt}}}

    \noindent
    {\small Below is a number. Multiply it by 9/5 and then add 32. Your answer should be a single number:}

    \bigskip
    
    \begin{tabular}{@{~}l@{~~}c@{~}p{1cm}@{~~~}c@{~}p{1cm}@{~~~}c@{~}p{1cm}@{~~~}c@{~}p{1cm}}
        {\small \textbf{Input:}} & & \sethlcolor{white}{\footnotesize \texttt{\hl{558}}} & & \sethlcolor{white}{\footnotesize \texttt{\hl{731}}} & & \sethlcolor{white}{\footnotesize \texttt{\hl{842}}} & & \sethlcolor{white}{\footnotesize \texttt{\hl{977}}} \\
        
        {\small \textbf{Correct:}} & & \sethlcolor{white}{\footnotesize \texttt{\hl{1036.4}}} & & \sethlcolor{white}{\footnotesize \texttt{\hl{1347.8}}} & & \sethlcolor{white}{\footnotesize \texttt{\hl{1547.6}}} & & \sethlcolor{white}{\footnotesize \texttt{\hl{1790.6}}} \\
        
        {\small \textbf{GPT-4:}} & {\footnotesize \redx{}} & \sethlcolor{pink}{\footnotesize \texttt{\hl{1046.4}}} & \textcolor{checkgreen}{\footnotesize \textbf{\checkmark}} & \sethlcolor{aigreen}{\footnotesize \texttt{\hl{1347.8}}} & {\footnotesize \redx{}} & \sethlcolor{pink}{\footnotesize \texttt{\hl{1517.6}}} & {\footnotesize \redx{}} & \sethlcolor{pink}{\footnotesize \texttt{\hl{1750.6}}} \\
    \end{tabular}

    \bigskip

    \bigskip

    {\small \underline{\textbf{Priming control prompt}}}

    \noindent
    {\small In this task, you must return an answer based on an input.\\ \\ Below is a number. Multiply it by 9/5 and then add 32. Your answer should be a single number:}

    \bigskip
    
    \begin{tabular}{@{~}l@{~~}c@{~}p{1cm}@{~~~}c@{~}p{1cm}@{~~~}c@{~}p{1cm}@{~~~}c@{~}p{1cm}}
        {\small \textbf{Input:}} & & \sethlcolor{white}{\footnotesize \texttt{\hl{558}}} & & \sethlcolor{white}{\footnotesize \texttt{\hl{731}}} & & \sethlcolor{white}{\footnotesize \texttt{\hl{842}}} & & \sethlcolor{white}{\footnotesize \texttt{\hl{977}}} \\
        
        {\small \textbf{Correct:}} & & \sethlcolor{white}{\footnotesize \texttt{\hl{1036.4}}} & & \sethlcolor{white}{\footnotesize \texttt{\hl{1347.8}}} & & \sethlcolor{white}{\footnotesize \texttt{\hl{1547.6}}} & & \sethlcolor{white}{\footnotesize \texttt{\hl{1790.6}}} \\
        
        {\small \textbf{GPT-4:}} & \textcolor{checkgreen}{\footnotesize \textbf{\checkmark}} & \sethlcolor{aigreen}{\footnotesize \texttt{\hl{1036.4}}} & \textcolor{checkgreen}{\footnotesize \textbf{\checkmark}} & \sethlcolor{aigreen}{\footnotesize \texttt{\hl{1347.8}}} & {\footnotesize \redx{}} & \sethlcolor{pink}{\footnotesize \texttt{\hl{1537.6}}} & {\footnotesize \redx{}} & \sethlcolor{pink}{\footnotesize \texttt{\hl{1760.6}}} \\
    \end{tabular}

    \bigskip

    \bigskip

    {\small \underline{\textbf{Primed prompt}}}

    \noindent
    {\small In this task, you must convert a number from Celsius to Fahrenheit.\\ \\ Below is a number. Multiply it by 9/5 and then add 32. Your answer should be a single number:}

    \bigskip
    
    \begin{tabular}{@{~}l@{~~}c@{~}p{1cm}@{~~~}c@{~}p{1cm}@{~~~}c@{~}p{1cm}@{~~~}c@{~}p{1cm}}
        {\small \textbf{Input:}} & & \sethlcolor{white}{\footnotesize \texttt{\hl{558}}} & & \sethlcolor{white}{\footnotesize \texttt{\hl{731}}} & & \sethlcolor{white}{\footnotesize \texttt{\hl{842}}} & & \sethlcolor{white}{\footnotesize \texttt{\hl{977}}} \\
        
        {\small \textbf{Correct:}} & & \sethlcolor{white}{\footnotesize \texttt{\hl{1036.4}}} & & \sethlcolor{white}{\footnotesize \texttt{\hl{1347.8}}} & & \sethlcolor{white}{\footnotesize \texttt{\hl{1547.6}}} & & \sethlcolor{white}{\footnotesize \texttt{\hl{1790.6}}} \\
        
        {\small \textbf{GPT-4:}} & \textcolor{checkgreen}{\footnotesize \textbf{\checkmark}} & \sethlcolor{aigreen}{\footnotesize \texttt{\hl{1036.4}}} & \textcolor{checkgreen}{\footnotesize \textbf{\checkmark}} & \sethlcolor{aigreen}{\footnotesize \texttt{\hl{1347.8}}} & \textcolor{checkgreen}{\footnotesize \textbf{\checkmark}} & \sethlcolor{aigreen}{\footnotesize \texttt{\hl{1547.6}}} & {\footnotesize \redx{}} & \sethlcolor{pink}{\footnotesize \texttt{\hl{1780.6}}} \\
    \end{tabular}
    }
    \end{AIbox}
    \end{minipage}
    \hspace{20pt}
    \begin{minipage}[l]{0.25\textwidth}
    \includegraphics[scale=0.65]{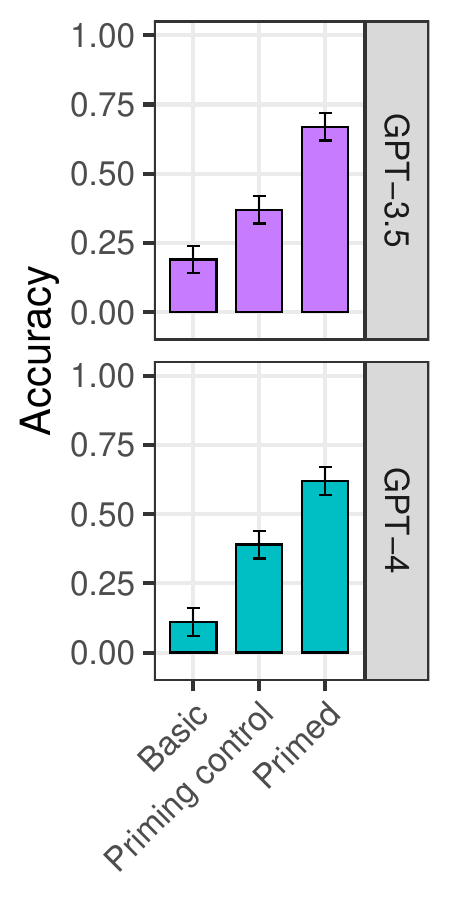}
    \end{minipage}

    \caption{\textbf{Left:} The prompts used to ask models to apply the function $f(x) = (9/5)x + 32$, with example predictions from GPT-4. \textbf{Right:} Model performance on the different prompts. Error bars show one standard error}
    \label{fig:celsiusprompts}
\end{figure}

\paragraph{Details:} We hypothesized that models would perform better at applying the function $f(x) = (9/5)x + 32$ when it was pointed out that this function was the Celsius-to-Fahrenheit conversion than when the function was simply presented as an abstract equation. This hypothesis is motivated by the fact that, when LLMs have previously encountered input-output pairs illustrating this function, it would usually have been in the context of Celsius-to-Fahrenheit conversion, so mentioning this context may help to activate the components of the LLM that are relevant to this function.
To test this hypothesis, we used three conditions illustrated in Figure~\ref{fig:celsiusprompts}. In the basic condition, models were simply told to apply this function. In the primed condition, they were given the same prompt as in the basic condition, except that it was prefixed with the sentence \textit{In this task, you must convert a number from Celsius to Fahrenheit}. Finally, the priming control condition was similar to the primed condition, but the new sentence that was added at the start was instead the generic statement \textit{In this task, you must return an answer based on an input}; we used this condition to check whether any differences between the primed condition and the basic condition were solely due to the presence of an additional sentence at the start of the prompt. 

We were most interested in models' abilities to compute the linear function's output rather than their ability to retrieve a memorized value. Accordingly, we restricted the input values to be greater than 500 and to not be divisible by 10; in a sample of Internet text from C4 \citep{raffel2020exploring}, we found that these constraints eliminated 99.8\% of examples mentioning a Celsius-Fahrenheit equivalence because people rarely discuss temperatures greater than 500°C, and when they do they often use round numbers that are divisible by 10 (see Appendix~\ref{app:lineardescription}). Therefore, applying these constraints restricts the evaluation to values that models are less likely to have memorized, though we cannot be rule out the possibility that models may have memorized these values because we do not have access to the training corpora for GPT-3.5 and GPT-4.

Both models performed much better when the task was framed as Celsius-Fahrenheit conversion (in the primed condition) than otherwise (in the basic and priming control conditions), as shown in the top of Figure~\ref{fig:celsiusprompts}. Using mixed-effects logistic regressions, we found that these differences were statistically significant for both models ($p < 0.01$ for GPT-3.5 and GPT-4; see Appendix~\ref{app:linearpromptsig}). This difference further shows that models' ability to apply this function is tied to the function's use in a particular scenario, as invoking that scenario improves performance.

At a more detailed level, there are two potential explanations for this result. First, it might be that models have learned the linear function in a systematic way, but their ability to access this systematic ability is increased by an explicit mention of Celsius and Fahrenheit. Second, it might be that models do not have a systematic ability to compute this function but have instead memorized many Celsius-Fahrenheit pairs. Under this account, mentioning the words \textit{Celsius} and \textit{Fahrenheit} is not only helpful but in fact necessary---models have no way to connect the two numbers except in the context of a memorized temperature-related pair. By using rare types of input values, we have attempted to reduce the chance of finding results that only illustrate a memorization-based explanation, pointing instead toward the explanation that posits systematic abilities that can only be accessed under certain settings, but we cannot rule out the memorization-based explanation: the GPT models have been trained on so much data that they likely have had opportunities to encounter rare temperature values.

Perhaps more surprisingly, models also perform much better in the priming control condition than in the basic condition. These differences were statistically significant for both models ($p < 0.01$ for GPT-3.5 and GPT-4; see Appendix~\ref{app:linearpromptsig}). We do not see any way in which the priming control prompt provides more information than the basic prompt, since all of the information in the initial sentence of the priming control condition is already apparent in the basic prompt by virtue of its structure. This finding shows how models can be highly sensitive to wording choices, reinforcing the point that they are primarily models of word sequences rather than idea-manipulating models.

\subsubsection{Multiplication}\label{sec:multiplication}

\paragraph{Summary:} When GPT-4 is asked to multiply two numbers, its performance varies widely depending on how the numbers are written. Its accuracy is 0.46 when the numbers are written in digits (294), 0.39 when the numbers are written in lowercase words (two hundred ninety-three), 0.35 when the numbers are written in all caps (TWO HUNDRED NINETY-THREE), and 0.17 when the numbers are written in alternating capitals (tWo HuNdReD nInEtY-tHrEe).

\begin{figure}[t]
    \centering
    \begin{minipage}[l]{0.42\textwidth}
    \centering
    \begin{AIbox}{Three-digit multiplication: Input format}{230pt}
    \parbox[t]{\linewidth}{

    {\small \underline{\textbf{Digits}}}
    
    \begin{tabular}{@{}c@{~}l@{~~}p{6cm}}
        & {\small \textbf{Input:}} & \sethlcolor{white}{\footnotesize \texttt{\hl{285 times 796}}} \\
        & {\small \textbf{Correct:}} & \sethlcolor{white}{\footnotesize \texttt{\hl{226860}}}\\
        \textcolor{checkgreen}{\textbf{\checkmark}} & {\small \textbf{GPT-4:}} & \sethlcolor{aigreen}{\footnotesize \texttt{\hl{226860}}}
    \end{tabular}

    \bigskip

    {\small \underline{\textbf{Lowercase words}}}
    
    \begin{tabular}{@{}c@{~}l@{~~}p{5.5cm}}
        & {\small \textbf{Input:}} & \sethlcolor{white}{\footnotesize \texttt{\hl{two hundred eighty-five times seven hundred ninety-six}}} \\
        & {\small \textbf{Correct:}} & \sethlcolor{white}{\footnotesize \texttt{\hl{226860}}}\\
        \textcolor{checkgreen}{\textbf{\checkmark}} & {\small \textbf{GPT-4:}} & \sethlcolor{aigreen}{\footnotesize \texttt{\hl{226860}}}
    \end{tabular}

    \bigskip
    
    {\small \underline{\textbf{All-caps words}}}

        \begin{tabular}{@{}c@{~}l@{~~}p{5.5cm}}
        & {\small \textbf{Input:}} & \sethlcolor{white}{\footnotesize \texttt{\hl{TWO HUNDRED EIGHTY-FIVE times SEVEN HUNDRED NINETY-SIX}}} \\
        & {\small \textbf{Correct:}} & \sethlcolor{white}{\footnotesize \texttt{\hl{226860}}}\\
        \redx{} & \textbf{GPT-4:} & \sethlcolor{pink}{\footnotesize \texttt{\hl{226620}}}
    \end{tabular}

    \bigskip
    
    {\small \underline{\textbf{Alternating capitals}}}

        \begin{tabular}{@{}c@{~}l@{~~}p{5.5cm}}
        & {\small \textbf{Input:}} & \sethlcolor{white}{\footnotesize \texttt{\hl{tWo HuNdReD eIgHtY-fIvE times sEvEn HuNdReD nInEtY-sIx}}} \\
        & {\small \textbf{Correct:}} & \sethlcolor{white}{\footnotesize \texttt{\hl{226860}}}\\
        \redx{} & \textbf{GPT-4:} & \sethlcolor{pink}{\footnotesize \texttt{\hl{226960}}}
    \end{tabular}
    }
    \end{AIbox}
    \end{minipage}
    \hspace{0.5cm}
    \begin{minipage}[l]{0.33\textwidth}
    \vspace{1cm}
    \includegraphics[scale=0.65]{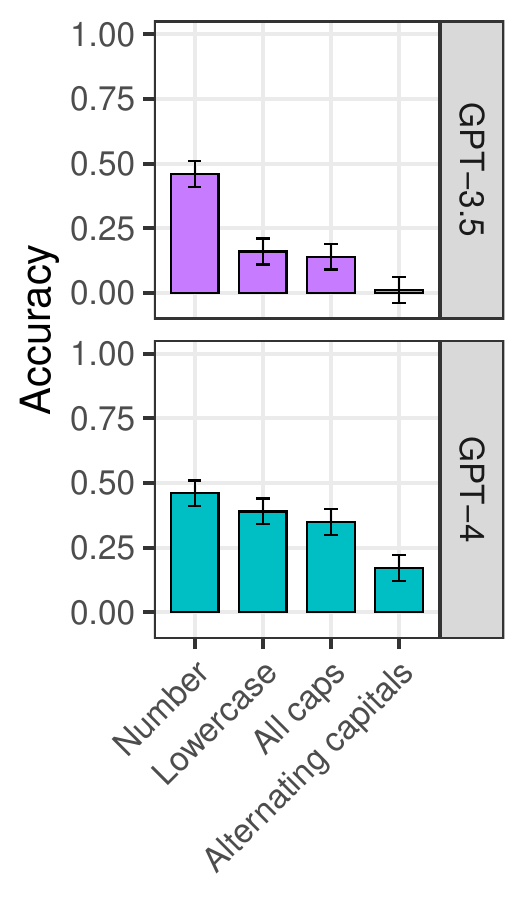}
    \end{minipage}
    \caption{\textbf{Left:} Example stimuli for multiplication problems with different types of input formatting. \textbf{Right:} Accuracy on multiplication using the different types of number formats. The error bars show one standard error.}
    \label{fig:multiplication_formats}
\end{figure}

\paragraph{Details:} We gave models a multiplication problem where both numbers being multiplied were random three-digit numbers. There were four conditions, corresponding to different ways of writing the numbers---using digits, lowercase words, all-caps words, or alternating capitals. See Figure~\ref{fig:multiplication_formats} for examples.
The values of the numbers were identical across the conditions.

Model accuracy varied substantially depending on how the numbers were formatted (Figure~\ref{fig:multiplication_formats}, right), and many of the differences between formats were statistically significant (specifically, all comparisons were significant except digits vs.\ lowercase for GPT-4, digits vs.\ all caps for GPT-4, and lowercase vs.\ all caps for both models; see Appendix~\ref{app:multsig}). This result illustrates how the models are, first and foremost, language models rather than math models or reasoning models; as a result, the same task expressed in different linguistic forms can be handled very differently by models.

\section{Discussion}

Our experiments highlight two scenarios where AI practitioners should be careful about using LLMs. First, we have shown that LLMs perform worse on rare tasks than on common ones, so we should be cautious about applying them to tasks that are rare in pretraining data. 
Second, we have shown that LLMs perform worse on examples with low-probability answers than ones with high-probability answers, so we should be careful about using LLMs in situations where they might need to produce low-probability text.
Overcoming these limitations is an important target for future work in AI.

Our experiments were inspired by a teleological analysis---an analysis aimed at understanding a system by understanding the problems that it was trained to solve. Our results therefore demonstrate the usefulness of the teleological approach as a way to illuminate important properties of a model. In the rest of the Discussion, we first compare the teleological approach to other approaches for understanding (Section~\ref{sec:discussion_understandnew}) or adversarially testing (Section~\ref{sec:discussionadversarial}) AI systems. 
We then describe how a teleological perspective can inform human-AI comparisons (Section~\ref{sec:discussion_comparehuman}). Finally, we discuss the ways in which scaling and advanced prompting techniques could improve model performance on the tasks we have investigated (Section~\ref{sec:promptvariations}).

\subsection{How to understand a new type of intelligence}\label{sec:discussion_understandnew}

To understand the benefits of the teleological approach, it is helpful to compare it to a more prevalent approach that we call \textbf{desideratum-focused evaluation}. In desideratum-focused evaluation, the evaluator starts by specifying some desideratum---a property or capability that they wish for models to have. The desideratum might be the ability to perform certain tasks that are targeted by a benchmark \cite[e.g.,][]{wang2019superglue,srivastava2022beyond}, or it might be a more fine-grained property such as the avoidance of a particular type of error \citep[e.g.,][]{ribeiro2020beyond,dou2022gpt}. Crucially, this approach is model-agnostic: what we want from a model is not driven by that model's nature. Therefore, the desideratum-driven approach runs the risk of missing model-specific properties, properties that one would not think to check for based on a generic characterization of the behavior that we want. In contrast, the teleological approach starts with an analysis of the model and is therefore well-suited for capturing the ways in which a model's behavior is influenced by its nature. 

As an example, a desideratum-focused approach to evaluating rot-13 would be likely to miss models' sensitivity to answer probability because probability is not part of how humans usually discuss this task. 
In more detail, we can characterize the desideratum as the ability to decode the sort of rot-13 text that users are likely to provide to the model. A standard way to test for such an ability would be to encrypt many sentences from a natural corpus and then evaluate whether models can decrypt them. Since low-probability sentences are by definition rare in natural corpora, such an evaluation would overlook models' sensitivity to sentence probability.
Rather than solely evaluating on a general test set, we might instead test for finer-grained desiderata---an approach akin to writing unit tests when testing software. We might, for example, write tests that check whether the model works correctly on letters that wrap around the end of the alphabet and on non-alphabetic characters. However, it might never occur to us to include a unit test that checks whether the program is invariant to string probability.
Taking a teleological perspective helps us to identify important tests such as this one that we might otherwise have missed.

To be clear, we are not arguing against testing for desiderata---such testing is necessary for answering important questions about whether models are doing what we want. 
Rather, we are arguing that such testing should be informed by a teleological analysis because this analysis can reveal non-obvious ways that models are likely to deviate from the desired behavior. 
Without a teleological analysis, we are likely to miss some potential failure modes, in which case our tests for desiderata will not actually be evaluating what we want them to evaluate---e.g., a generic rot-13 benchmark might lead us to falsely conclude that models have mastered rot-13 when in fact their rot-13 ability has important shortcomings that were simply not tested for.
For discussion of the challenges of ensuring that evaluations test what they are intended to, see \citet{lipton2018troubling}, \citet{linzen2020accelerate},\citet{raji2021ai},\citet{rodriguez2021evaluation}, and \citet{hutchinson2022evaluation}.

One reason why the teleological approach can reveal points that are otherwise overlooked is that it leverages more information than other approaches. 
That is, standard evaluation approaches treat LLMs as impenetrable black boxes. The teleological approach instead recognizes that, although the inner workings of LLMs are indeed difficult to understand, there is nonetheless a lot that we do know about these models:
we know their architecture, and we know what they were trained to do. We can use this knowledge to reason about how models are likely to approach a task, providing us with testable hypotheses about their behavior.\footnote{In the terminology of Marr's levels of analysis \citep{marr1982vision}, LLMs are black boxes at the algorithmic level but not at the implementation level or the computational level. Because the different levels of analysis influence each other, we can use what we know about computation and implementation to infer likely properties of the unknown algorithmic level.} To be sure, many prominent LLMs are proprietary ones for which many training details are not public, but---as we have shown---we still have enough general information about these systems to make predictions about their behavior.

A final benefit of the teleological approach is that it reveals unifying principles behind \textit{why} systems do what they do \citep{shepard1987toward,marr1982vision,anderson1990adaptive,griffiths2020understanding}. 
Other approaches, such as the desideratum-driven approach, instead tend to provide lists of scores without a unifying characterization of why the model displays the behavioral patterns that it does. 
Having a set of deeper explanatory principles is useful because it enables us to make more general predictions about the sorts of scenarios that models are likely to handle well or poorly. 

Besides the desideratum-based approach, there are two other recently-proposed approaches that are more similar to our proposal, but there are still important differences.
First, \citet{liang2022holistic} and \citet{holtzman2023generative} have called for holistic approaches that characterize LLMs by providing generalizable descriptions of the behavioral properties of LLMs (e.g., what general abilities do they have, and what general types of errors do they make?) This approach therefore focuses on \textit{what} models do. Second, \citet{binz2023using} argue for an approach based on experimental psychology: treat LLMs as if they are participants in an experiment, thereby applying controlled tests that illuminate the particular strategies that they use---i.e., this approach analyzes \textit{how} models do what they do. The teleological approach, like these other two approaches, emphasizes searching for general principles that define a model, but it differs in that it pursues this goal through an analysis of \textit{why} rather than \textit{what} or \textit{how}. 
These approaches can be fruitfully combined: a teleological analysis can be a useful source of hypotheses about the strategies that models will adopt and the behaviors that will result---indeed, this is how most of our experiments proceeded in this paper: we started by using \textit{why} to make predictions about \textit{how}, then used those predictions to inform predictions about \textit{what}, and then experimentally tested the \textit{what} predictions.

\subsection{Adversarial testing of AI systems}\label{sec:discussionadversarial}

Our evaluation strategy falls within the adversarial tradition of AI research---a line of work based on identifying situations where AI systems make mistakes so that we can understand them better and make them more robust (\citealt{barreno2006can}, \citealt{szegedy2014intriguing}; see \citealt{xu2020adversarial} for a review). 
There are several distinct strategies for identifying situations in which AI systems perform poorly. Many of the best-known strategies can be placed into one of two categories:

\vspace{-3pt}
\begin{enumerate}
    \item \textbf{White-box attacks:} Use knowledge of the model's architecture and parameters to identify input perturbations that break the model. 
    \item \textbf{Black-box attacks:} Analyze the model's input-output behavior (but not its internal workings) to identify types of examples on which it tends to make mistakes.
\end{enumerate}

\vspace{-3pt}

\noindent
A prominent example of a white-box attack is adversarial examples in computer vision where an image classifier's prediction is changed by applying a small perturbation that is derived through a process similar to backpropagation \citep{szegedy2014intriguing}; in one famous example, a classifier initially classified an image of a panda correctly, but after a perturbation that was imperceptible to humans, it then confidently classified the image as a gibbon \citep{goodfellow2015adversarial}. Black-box attacks have recently been used as a way to break game-playing AI systems, where an adversarial policy is trained on observations of the system's game play in order to identify strategies that defeat the system \citep{wang2023adversarial}.

The approach that we use is distinct from both the white-box strategy and the black-box strategy because the way that we identify challenging scenarios does not involve access to the model at all. Instead, we analyze the pressures that have shaped the system in order to make hypotheses about situations that will challenge it; in our case, this strategy led us to push LLMs into low-probability spaces where we expected they would perform poorly. That is, rather than analyzing the box (whether it be a white box or a black box), we are analyzing the factory in which the box was produced. 
This approach is reminiscent of prior challenge sets, such as the HANS dataset \citep{mccoy2019right}, that were created by identifying types of examples that are rare in common training sets.

One final strategy for predicting which examples will be challenging is to use an abstract notion of example complexity.
For instance, one might hypothesize that long sentences will be harder to process than short ones, or that multi-hop reasoning questions will be harder to answer than single-hop ones.
A downside of this approach is that the definitions of complexity that we focus on may not correspond to the factors that cause difficulty for a given model. The teleological approach provides a potential way around this shortcoming because it helps us identify what notion of difficulty is relevant for the model at hand. 
For example, if we approached the task of counting from first principles, a reasonable hypothesis would be that the main source of difficulty would be the size of the number being counted to. As our teleologically-informed experiments have shown, however, counting difficulty in language models is in fact driven more by the commonness of the number being counted to than by its size; e.g., models performed better at counting to 100 (a number that is used frequently) than counting to many numbers that are smaller but rarer, such as 59. 

In using a teleological adversarial strategy, we do not mean to dismiss the other approaches. The strategy that is most useful for a given project depends on that project's goals. For example, white-box and black-box attacks are useful when we intend to find model errors automatically, and a complexity-guided approach is useful when we have no knowledge of what the system was trained to do. The situation where the teleological perspective is most useful is when we seek a holistic understanding of a system for which we have at least some knowledge about the training task, training distribution, and model architecture.

\subsection{Comparing models to humans}\label{sec:discussion_comparehuman}

We have argued that, in order to understand LLMs, we should approach them on their own terms rather than evaluating them in the same ways that we test humans. But what if your goal is to compare LLMs to humans? Even in that case, we argue that the teleological perspective is important; much as it should inform testing for desiderata  (Section~\ref{sec:discussion_understandnew}), it should also inform comparisons to humans. Specifically, we argue that human-likeness can be better assessed by the two-step process in \ref{ex:2step} than the direct process in \ref{ex:1step}:

\vspace{-3pt}
\ex. \textbf{Sub-optimal approach for investigating human-likeness:} Ask ``is this model like a human?''\label{ex:1step}

\vspace{-12pt}
\ex. \textbf{Better approach for investigating human-likeness:}\label{ex:2step}
\a. First, use teleologically-motivated experiments to characterize the model in its own terms.
\b. Then, ask ``in what ways is this characterization like a human and unlike a human?''

There are several reasons why the direct approach in \ref{ex:1step} is sub-optimal.
First, it carries a risk of confirmation bias: many model behaviors are human-like in some ways but not in other ways, leading to a situation where an experiment's results could likely be interpreted as supporting whichever perspective the experimenter wants to find evidence for, whether it be human-likeness or human-unlikeness.
Second, the natural way to pursue path \ref{ex:1step} is to evaluate models by using tests developed for humans. Such an evaluation may overestimate similarities to humans because, for many tasks, there is only one way to be correct yet many ways to be incorrect. 
Thus, to the extent that models are accurate, they will likely be accurate in human-like ways; and to the extent that they make mistakes, a test that is designed for humans may only highlight the sorts of mistakes that humans are liable to make, omitting potential failure modes that are unique to models.
Third, if an experimenter starts with the implicit assumption that a model is the same sort of entity as a human, the experiment's results are unlikely to cause them to reassess that assumption because any errors can be interpreted as the model being only \textit{quantitatively} worse than a human, even if the correct conclusion is that the model is \textit{qualitatively} different.
The teleological approach in \ref{ex:2step} mitigates these risks by evaluating models in a way that is less biased by our view of human cognition, making it better equipped to reveal ways in which models might be qualitatively different from humans without removing the potential for also demonstrating similarities to humans.

One of the most influential proposals for investigating human-likeness is the Turing test \citep{turing1950computing}, in which models are evaluated on their ability to converse with a human such that an observer cannot reliably tell whether the computer is a computer or human. The teleological approach that we advocate can be viewed as a targeted way to propose queries for use in a Turing test. This more targeted approach can enable a Turing test to be more effective by revealing qualitatively non-human-like aspects of a model that would not be tested for in a model-agnostic Turing test---i.e., accessing the dotted region of the Venn diagram in Figure~\ref{fig:venn}. Turing originally referred to his test as the \textbf{imitation game}; since we propose to understand models by understanding the limitations that shape their behavior \citep{griffiths2020understanding}, this approach could instead be called the \textbf{limitation game}. 

In this work, we have focused on performing the first step of \ref{ex:2step}. Without evaluating humans on the same tasks and examples, we cannot make strong claims about whether the LLM properties that we have identified make LLMs qualitatively different from humans---the mere fact that models are making errors does not necessarily make them different from humans \citep{dasgupta2022language}, and when comparing types of intelligence one must be careful to evaluate them fairly to avoid penalizing one system for being limited by factors extraneous to the comparison at hand \citep{de2016we,firestone2020performance,lepori2022can,lampinen2022can}. Nonetheless, based on prior work studying human cognition, we suspect that most of the properties we have identified in Table~\ref{tab:listofembers} are present in humans to some extent but to a lesser degree than in LLMs. In the list below, we comment in more detail on several of the properties we have observed in LLMs:
\begin{itemize}
    \item \textbf{Sensitivity to task probability:} We have shown that LLMs achieve higher accuracy on common tasks than on rare tasks that are no more formally complex. Several studies have shown cases where humans performed better on a task they were trained on than on a similar but new task \citep{luchins1942mechanization,ho2023rational} or even on the same task with new items \citep{posner1968genesis}, showing that human performance is indeed influenced by experience with the task in question. However, we expect that humans are less sensitive to this factor than LLMs, in part because of the diversity of tasks that humans have to perform. In particular, we expect that, in comparison to LLMs, humans are better able to solve a task by using an abstract algorithm that can be equally well applied across the task variants that we studied (e.g., across different shifts in a shift cipher, or across different suffixes used in Pig Latin). If this view is accurate, humans would not show such stark differences across tasks as those we have found for LLMs. 
    \item \textbf{Sensitivity to sentence probability:} It is well-established in psycholinguistics that a human's ease of processing a sentence is influenced by sentence probability \citep{ehrlich1981contextual,smith2013effect}.  Nonetheless, we expect that humans are not as thoroughly influenced by string probability as LLMs are. In particular, we have shown that LLM performance is influenced substantially by probability even in deterministic tasks (e.g., Pig Latin decoding and acronym formation), and we expect that for such tasks humans would be better able to turn off their reliance on probability, yielding results that are more uniform across probability levels. Though general sentence probability may not play a large role in human performance on our tasks, there is one finding that we have attributed to probability that we expect would also occur in humans, namely the tendency to regularize an implausible output when there is a similar output that is much more plausible. \citet{gibson2013rational} showed that humans often perform such regularization when given a sentence with an implausible meaning; e.g., when presented with \textit{The mother gave the candle the daughter}, participants often interpreted it as \textit{The mother gave the candle \underline{to} the daughter.} Based on this finding, we think it is likely that, if humans were tested on our targeted rot-13 sentences, they would regularize the outputs in ways similar to the LLMs we studied (e.g., producing \textit{Are they now building a bridge of their own?} when the correct answer is \textit{Are they now building a bridge of their owl?}). 
    \item \textbf{Sensitivity to wording:} In Section~\ref{sec:multiplication}, we showed that models' accuracy on multiplication varied substantially based on the formatting of the numbers being multiplied. \citet{spelke2001language} showed that bilingual humans who were trained on math problems in one language could perform such problems more effectively in the language of instruction than in the other language that they were also fluent in. Thus, at a high level, it can be said that humans are also sensitive to the formatting of numbers used in math problems. Nonetheless, there are also important differences between our LLM results and \citeauthor{spelke2001language}'s human results: First, for the humans, the difference in ``formatting'' is a difference in the language used; in our LLM experiments, we instead varied capitalization and the use of digits vs.\ words. 
    We suspect that humans would be less sensitive to these particular perturbations than LLMs are. Second, though humans showed statistically significant differences between evaluation languages, these differences were generally less stark than the differences that we observed in LLM performance. 
\end{itemize}
In sum, it is reasonable to view our results as illuminating some ways in which LLMs are decidedly non-human-like, but establishing this conclusion with certainty would require further work that directly compares LLMs and humans on the same stimuli.

\subsection{Evaluating models fairly: Prompts, scaling, and calibration}\label{sec:promptvariations}

All of our experiments have used basic prompting, in which models are simply asked to provide the response to a query. For many of our tasks, it is likely that model performance could be increased by more advanced prompting techniques, such as chain-of-thought prompting \citep{wei2022chain} and step-by-step prompting \citep{kojima2022large}; for instance, \citeauthor{wei2022chain}'s chain-of-thought approach substantially improved performance on the task of last-letter concatenation, which is similar to our acronym task.
Similarly, it is also likely that performance on our tasks will improve as models are further scaled up.
However, it would not invalidate our conclusions if there are conditions in which LLM performance is greater than what we have observed. We do not claim to be highlighting fundamental incapabilities of LLMs but rather are claiming that some tasks and examples are harder for LLMs than others. Therefore, the existence of \textit{any} setting that yields the predicted performance differences supports our claims, even if other settings exist where LLM performance is at ceiling for both conditions being compared. 

\begin{figure}[t]
    \centering
    \includegraphics[scale=0.55]{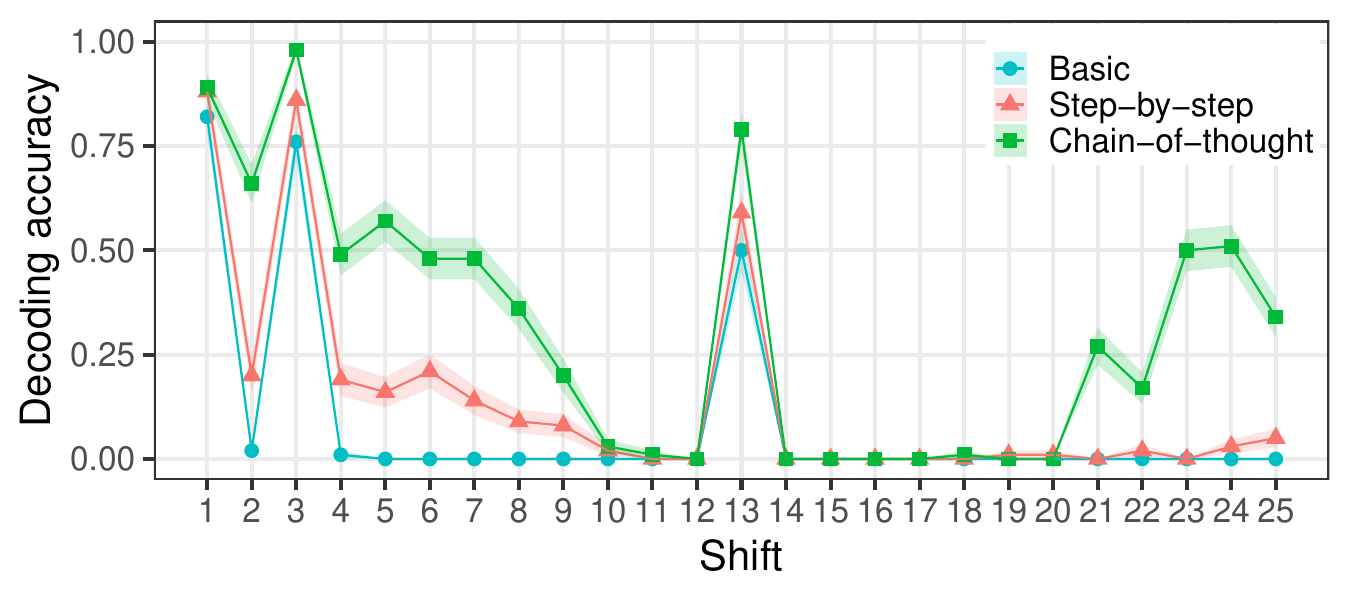}

    \vspace{0.5cm}
    
    \includegraphics[scale=0.55]{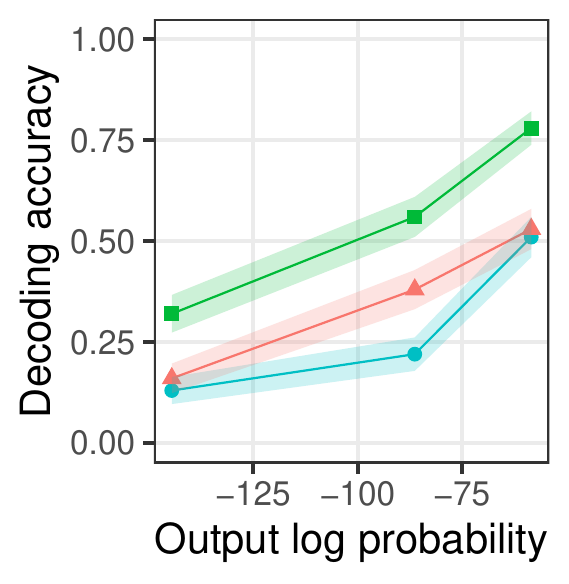} \hspace{0.6in}
    \includegraphics[scale=0.55]{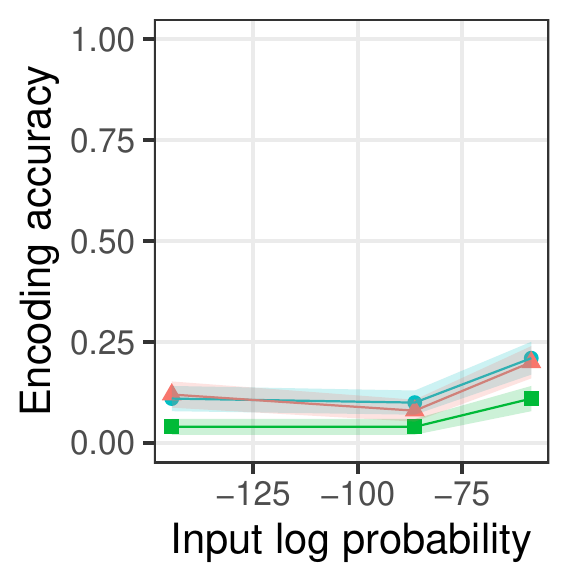}
    \caption{Comparing three prompting styles (basic, step-by-step, and chain-of-thought) on processing shift ciphers. In all plots, the intervals around the lines show one standard error. \textbf{Top:} Accuracy on decoding shift-cipher-encrypted text with different shifts. \textbf{Bottom left:} Influence of output probability when decoding. \textbf{Bottom right:} Influence of input probability when encoding.}
    \label{fig:cotstep}
\end{figure}

As a first step toward extending our results to other prompting techniques, we used chain-of-thought prompting and step-by-step prompting to evaluate GPT-3.5 and GPT-4 on shift ciphers (see Appendices~\ref{app:stepcotprompts} through \ref{sec:stepdecprompt} for details on the prompts). The results are in Figure~\ref{fig:cotstep} and show that, at least when decoding, these prompting techniques can indeed substantially increase performance (for encoding---shown in the bottom right of Figure~\ref{fig:cotstep}---they appear to hurt performance). 
Nonetheless, it is worth noting that the basic effects we have identified still hold (see Appendix~\ref{app:shiftsigpromptstyle} for significance tests): First, even with more advanced prompting, models show similar patterns of difficulty on different shift levels---e.g., showing a spike in accuracy at a shift of 13, and showing greater accuracy for rot-1 and rot-3 than rot-2 (Figure~\ref{fig:cotstep}, top). Second, models still perform better on high-probability sentences than low-probability ones, particularly when decoding (Figure~\ref{fig:cotstep}, bottom). Thus, while more advanced prompting techniques improve overall accuracy, they do not erase the ways in which models are influenced by their autoregressive roots. 

A similar conclusion applies to scaling. Across almost all tasks, we have observed that GPT-4 substantially outperforms GPT-3.5, showing that scaling can provide large benefits on the tasks we have studied---as it does for many other tasks \citep{kaplan2020scaling,rae2021scaling,srivastava2022beyond,wei2022emergent}. However, even though GPT-4 scored better than GPT-3.5, it still displayed the same qualitative trends of being sensitive to task probability and example probability. Therefore, like prompting techniques, it appears that scaling may improve overall performance but may not fully overcome the ways in which models are influenced by their nature.

In addition to prompt engineering and scaling, another way in which embers of autoregression could potentially be mitigated would be to ask the LLM how difficult the task at hand is for it. 
Ensuring that LLMs correctly report when they have difficulty is an ongoing area of research \citep{mielke2022reducing,kadavath2022language}; to the extent that an LLM is well-calibrated in this regard, its judgments could allow us to recognize when we should be cautious about its outputs (e.g., we might take a judgment that difficulty is high as a sign that we should verify the response with a human or pass the query to a symbolic interpreter that can robustly handle the task).
As an initial step toward investigating whether LLMs recognize that the tasks we have studied are difficult for them, we considered five queries for each of three tasks where we found LLMs to struggle:
\begin{itemize}
    \item Counting a list of 29 letters.
    \item Performing the article swapping task in the low-probability setting.
    \item Decoding a message written in rot-10, where the correct answer is a high-probability sentence.
\end{itemize}
For each of these cases, we tested GPT-4 via the online chat interface (accessed on September 20, 2023) on five examples. Of the fifteen total cases, there were two that GPT-4 answered correctly (both within the counting task); since we are most interested in cases where GPT-4 was incorrect, we regenerated those responses, which yielded an incorrect answer for both. Each of the fifteen queries was asked in a new chat session. After we received GPT-4's incorrect response, we then asked it, ``On a scale from 1 to 10---where 1 is easy and 10 is hard---how difficult was this question to answer for you?'' In all fifteen cases, GPT-4 responded that the question had a difficulty level of 1---i.e., very easy. 

Two examples of GPT-4's difficulty estimates are in Figure~\ref{fig:calibration}. In the first case, it states that the task is ``straightforward...for a computer program.'' In the second case, it refers to what its ``algorithms'' can perform easily. One possible interpretation of these responses is that GPT-4 is estimating difficulty based on a generic definition of computation, rather than by considering its own specific capacities and limitations. 
Perhaps, then, GPT-4 itself could benefit from the teleological perspective, as this perspective emphasizes approaching systems on their own terms.

\begin{figure}[t!h!]
    \centering
    \begin{AIbox}{Difficulty calibration}{0.95\textwidth}
    \parbox[t]{\linewidth}{

    {\small \underline{\textbf{Experimenter:}}}

    {\small
How many letters are in the following list? ``bbbbbbbbbbbbbbbbbbbbbbbbbbbbb''

\vspace{2mm}
{\small \underline{\textbf{GPT-4:}}}

The list provided contains \sethlcolor{pink}\hl{28} ``b'' letters. \textit{[Note: The correct answer is 29]}

\vspace{2mm}
{\small \underline{\textbf{Experimenter:}}}

On a scale from 1 to 10 - where 1 is easy and 10 is hard - how difficult was this question to answer for you?

\vspace{2mm}
{\small \underline{\textbf{GPT-4:}}}

For me, answering that question would be a 1 on the scale from 1 to 10 in terms of difficulty. Counting the number of characters in a string is a straightforward task for a computer program.

    }

\tcbline

        {\small \underline{\textbf{Experimenter:}}}

    {\small
Rot-10 is a cipher in which each letter is shifted 10 positions forward in the alphabet. For example, here is a message written in rot-10 along with the original text that it was created from:

\noindent
Rot-10 text: ``Cdki robo!''

\noindent
Original text: ``Stay here!''

\vspace{2mm}
\noindent
Here is another message in rot-10. Decode this message to produce the original text:

\noindent
Rot-10 text: ``Dryco gry exnobcdkxn dro boqsyx gsvv qod grkd S wokx pbyw drsc cswzvo zrbkco.''

\noindent
Original text:

\vspace{2mm}
{\small \underline{\textbf{GPT-4:}}}

To decode a ROT-10 cipher, you would shift each letter 10 positions backward in the alphabet. Numbers, spaces, and punctuation marks remain the same. 

\vspace{2mm}
\noindent
Here's how to decode your ROT-10 message:

\vspace{2mm}
\noindent
Rot-10 text: ``Dryco gry exnobcdkxn dro boqsyx gsvv qod grkd S wokx pbyw drsc cswzvo zrbkco.''

\vspace{2mm}
\noindent
1. For ``D'', shifting 10 positions backward gives \sethlcolor{pink}\hl{``A''}.

\noindent
2. For ``r'', shifting 10 positions backward gives \hl{``i''}.

\noindent
3. For ``y'', shifting 10 positions backward gives \hl{``m''}.

\noindent
4. For ``c'', shifting 10 positions backward gives \hl{``w''}.

\noindent
5. For ``o'', shifting 10 positions backward gives \sethlcolor{aigreen}\hl{``e''}.

\vspace{2mm} 
\noindent
Following this pattern for the entire sentence, the original text becomes:

\vspace{2mm} 
\noindent
Original text: ``\sethlcolor{pink}\hl{After you }\sethlcolor{aigreen}\hl{understand the}\sethlcolor{pink}\hl{ method used for }\sethlcolor{aigreen}\hl{this simple}\sethlcolor{pink}\hl{ message}\sethlcolor{aigreen}\hl{.}''

\vspace{2mm}
\noindent
I've shifted each letter 10 positions backward in the alphabet to decode the ROT-10 message into its original text.

\vspace{2mm}
\textit{[Note: The correct answer is ``Those who understand the region will get what I mean from this simple phrase.'']}

\vspace{2mm}
{\small \underline{\textbf{Experimenter:}}}

On a scale from 1 to 10 - where 1 is easy and 10 is hard - how difficult was this question to answer for you?

\vspace{2mm}
{\small \underline{\textbf{GPT-4:}}}

For me, the question is very straightforward to answer, so on a scale from 1 to 10, it would be a 1 in terms of difficulty. My algorithms are designed to perform tasks like decoding ROT ciphers relatively easily.

    }
    }
    \end{AIbox}
    \caption{Asking GPT-4 to rate the difficulty of queries. These responses are from GPT-4's online chat interface, accessed on September 20, 2023.}
    \label{fig:calibration}
\end{figure}

Our results have focused on ways in which the teleological perspective illuminates shortcomings of LLMs, but being critical of LLMs is not our goal. Instead, our goal is to promote a perspective that accurately captures the properties of LLMs, both positive properties and negative ones. 
Though in many cases this perspective has led us to point out model weaknesses, in some ways it makes LLMs \textit{more} impressive than they would otherwise seem. Once we fully recognize that LLMs are statistical next-word prediction systems, it becomes remarkable that they can perform rot-13 or acronym formation \textit{at all}, even if they do not perform these tasks perfectly.

\section{Related work}

We have argued that the problem an LLM must solve has three components: the training task, the training distribution, and the model architecture. For each of these components, we provide below a review of prior literature about ways in which a system is influenced by that factor. This review includes the identification of several additional embers of autoregression beyond those that we have focused on above. 

\subsection{Training task}

The idea that a system is crucially influenced by the function it performs has a long history in cognitive science \citep{marr1982vision,shepard1987toward,anderson1990adaptive}.
One important consequence of this idea is that two different systems that have the same goal might converge to similar strategies even if they are implementationally very different. This fact can help us to understand one system by leveraging what we know about a different system that achieves the same goal. For instance, \citet{grant2018recasting} gained insight into the gradient-based meta-learning approach MAML \citep{finn2017model} by demonstrating a relationship to hierarchical Bayesian modeling, and \citet{tenney2019bert} illuminated processing strategies of the LLM BERT \citep{devlin2019bert} by showing that, faced with the task of modeling natural language, BERT arrived at a similar analysis pipeline as NLP practitioners who had faced the same goal.
Such parallels can also be drawn in the opposite direction: instead of hypothesizing that two systems will work in similar ways because they have similar goals, we can infer that two systems may be shaped by similar goals based on the fact that they work in similar ways---a type of inference at the heart of work making inferences about the human brain based on similarities to computational models \citep{schrimpf2021neural,caucheteux2022brains}. Perhaps most relevant to our perspective is the argument from \citet{shanahan2022talking} that those who develop and use LLMs must continually remind themselves of what the LLMs were trained to do, in order to avoid anthropomorphizing these systems; avoiding this temptation was one of our main reasons for pursuing a teleological approach to understanding LLMs.

In addition to helping us understand systems, the importance of a system's goal is also helpful in work aimed at improving models. First, many authors have studied which (pre-)training objectives are most effective for producing desired behavior in a model
\citep{wang2019tell,raffel2020exploring,lin2021limitations}. 
Second, beyond differences in general training tasks, the influence of a system's goal in driving that system's nature is substantial enough that models can be given highly targeted training objectives that can override a model's inherent inductive biases, causing one type of system to behave like another type of system. In this way, it is possible to distill a symbolic hypothesis space \citep{mccoy2020universal} or a probabilistic model \citep{muller2022transformers,mccoy2023modeling} into a neural network, or to distill the inductive biases of one type of neural network into another \citep{abnar2020transferring}.

The next subsections discuss two specific ways in which the LLM training task of next-word prediction influences their behavior.

\subsubsection{Difficulty in tasks that depend on meaning}

Since LLMs are trained to manipulate sequences of words, they are primarily models of language rather than models of thought \citep{mahowald2023dissociating}. They do not directly manipulate ideas or structured world models; to the extent that such processing occurs in LLMs, it has to emerge indirectly. LLMs therefore sometimes struggle in contexts that could be naturally handled by using and updating a model of entities in the world; e.g., they sometimes fail to track entities in a discourse \citep{kim2023entity} and are sometimes insensitive to the ways that negation affects statements about entities and their properties \citep{ettinger2020bert}. Outside of particular linguistic phenomena, LLMs also displaying shortcomings in tasks that require knowledge of the causal properties and affordances of everyday objects \citep{yiu2023imitation,kosoy2023comparing}.

Beyond the realm of specific linguistic phenomena, a philosophical debate has arisen about whether LLMs have any form of understanding or meaning \citep{mitchell2023debate}.
Some have indeed argued that text-only LLMs are fundamentally incapable of capturing meaning, under various construals of meaning \citep{bender2020climbing,bisk2020experience}, while others have pointed to plausible ways in which text-only systems could be said to acquire meaning \citep{mollo2023vector,merrill2022entailment,li2021implicit,piantadosi2022meaning}.
We remain agnostic about whether LLMs truly capture meaning or only capture other properties that correlate with it; what we believe is clear is that meaning-sensitive tasks do not come naturally to systems trained solely on textual input, such that we can expect LLMs to encounter difficulty in handling these tasks.

\begin{figure}
    \centering
    \begin{AIbox}{Arithmetic puzzle}{0.9\textwidth}
    \parbox[t]{\linewidth}{

    {\small \underline{\textbf{Prompt:}}}

        {\small

If a pen and a pencil together cost \$2.50, and the pen costs \$2 less than the pencil, how much does the pen cost? First give your answer, and then explain your reasoning.

\bigskip
\noindent
{\small \underline{\textbf{GPT-4 output:}}}

\noindent
{\small 
\sethlcolor{pink}The pen costs \hl{\$1.25}.

\vspace{-2pt}
\bigskip
\noindent
Explanation:

\noindent
Let's assume the cost of the pencil is $x$ dollars.

\noindent
The cost of the pen would then be $x - 2$ dollars.

\vspace{-2pt}
\bigskip
\noindent
According to the problem, the pen and pencil together cost \$2.50:

\noindent
$x + (x - 2) = 2.50$

\noindent
$2x - 2 = 2.50$

\noindent
$2x = 4.50$

\noindent
$x = 2.25$

\vspace{-2pt}
\bigskip
\noindent
So the pencil costs \$2.25, and the pen, which costs \$2 less, would be:

\noindent
\hl{$2.25 - 2 = 1.25$}

\noindent
The pen costs \hl{\$1.25}.

}

    }

\tcbline

    {\small \underline{\textbf{Prompt:}}}

        {\small

If a pen and a pencil together cost \$2.50, and the pen costs \$2 less than the pencil, how much does the pen cost? First explain your reasoning, and then give your answer.

\bigskip
\noindent
{\small \underline{\textbf{GPT-4 output:}}}

\noindent
{\small 
To solve this problem, we can use a system of equations.

\vspace{-2pt}
\bigskip
\noindent
Let $P$ represent the cost of the pen and $C$ represent the cost of the pencil.

\vspace{-2pt}
\bigskip
\noindent
1) The first equation comes from the statement ``a pen and a pencil together cost \$2.50'':

\noindent
$P + C = 2.50$

\vspace{-2pt}
\bigskip
\noindent
2) The second equation comes from the statement ``the pen costs \$2 less than the pencil'':

\noindent
$P = C - 2$

\vspace{-2pt}
\bigskip
\noindent
First, we'll solve for one variable in terms of the other using one of the equations. The second equation already gives us $P$ in terms of $C$, so we can use it to substitute into the first equation:

\noindent
$P + C = 2.50$

\noindent
$(C - 2) + C = 2.50$

\noindent
$2C - 2 = 2.50$

\noindent
$2C = 4.50$

\noindent
$C = 2.25$

\vspace{-2pt}
\bigskip
\noindent
Now we know the cost of the pencil, $C$, is \$2.25. To find the cost of the pen, $P$, we can use the second equation $P = C - 2$:

\noindent
$P = 2.25 - 2$

\noindent
$P = 0.25$

\vspace{-2pt}
\bigskip
\noindent
\sethlcolor{aigreen}So, the pen costs \hl{\$0.25}.
}

    }
    
    }
    \end{AIbox}
    \caption{When given an arithmetic puzzle, GPT-4 gets the wrong answer if it is prompted to first provide its answer and then its reasoning (top), but it gets the right answer if it is prompted to provide the reasoning first and then the answer (bottom). These examples were produced by the online version of GPT-4 on September 17, 2023.}
    \label{fig:stuck}
\end{figure}

\subsubsection{Inability to modify text that has already been produced}

As we mentioned in the introduction, Section 8 of \citet{bubeck2023sparks} discusses a way in which LLMs are limited by the autoregressive nature of how they interact with language: LLMs produce text one word at a time. They are not explicitly trained to plan far ahead, and they are unable to alter text once they have produced it. These facts sometimes cause inaccuracies in model performance because LLMs will produce an error due to the inability to plan ahead and cannot later correct that error because they cannot change their previously-produced text. 
An example of this issue is in Figure~\ref{fig:stuck}.
\citet{ortega2021shaking} refer to this issue as self-delusion: models treat the text that they have previously produced as part of the input from which they should make future predictions, so errors in their previous output propagate forward into later outputs.

\subsection{Training distribution}

When evaluating any system that learns, it is important to consider what the system learns from \citep{pullum2002empirical,linzen2020accelerate,kim2022uncontrolled,hupkes2022state,piktus2023roots}. A wealth of papers have therefore highlighted ways in which language model performance is influenced by the model's training data. Some have shown ways in which the choice of dataset plays a role, both with respect to the type of data (e.g., Internet text vs.\ child-directed speech: \citealt{huebner2021babyberta}) and with respect to the way in which the dataset is curated or filtered \citep{gao2020pile,lee2022deduplicating}. 
On a more fine-grained level, several papers have connected model behavior to specific (pre-)training examples that have contributed to that behavior \citep{han2023understanding,wei2021frequency,razeghi2022impact,han2022orca,elazar2022measuring,mccoy2023raven}.
Finally, several abstract abilities of language models have been traced to properties of their training data; such capacities include cross-lingual abilities \citep{blevins2022language} and in-context learning \citep{chan2022data}.
Below, we highlight two particular properties of LLMs that connect directly to the training distribution.

\subsubsection{Societal biases and spurious correlations}

Language models are trained to capture the statistical properties of their training sets. At times, these statistics can have undesirable effects on model behavior. First, neural network models can absorb societal biases that are encoded in natural textual distributions \citep{bolukbasi2016man,caliskan2017semantics,bender2021dangers}, leading them to behave in bigoted ways such as producing toxic, prejudiced text \citep{gehman2020realtoxicityprompts,deshpande2023toxicity}. Second, neural networks are susceptible to using invalid heuristics that yield the correct answer most of the time in their training distribution but that are not valid strategies in the general case \citep{mccoy2019right,niven2019probing}. For example, when language models are given several sentences with sentiment labels and are then asked to label the sentiment of a new sentence, their answer is highly influenced by the labels of the examples they were shown: they are biased toward producing the label that is most common among the examples, relying too heavily on the invalid statistical cue of label frequency rather than the cues that truly indicate sentence sentiment \citep{zhao2021calibrate}.
Both of these phenomena---societal biases and spurious correlations---are driven by properties of the distribution that a model is trained on, and the fact that models are so sensitive to such properties has driven calls to curate models' training data more carefully \citep{rogers2021changing}.

\subsubsection{Idiosyncratic memorization}

We have documented above the fact that LLMs are biased toward sentences that have a high probability (Section~\ref{sec:outputprob}).
It is important to note that, from the perspective of an LLM, the probability of a sentence is determined by the particular dataset that the LLM was trained on. As a result, if the dataset contains some  sentences that are frequently repeated, the model is likely to memorize them---even if they are not important or high-probability sentences in the broader world.

\citet{mccoy2023raven} provide one example. They show that several language models trained on the Wikitext-103 dataset \citep{merity2016pointer}, when left to their own devices, have a tendency to generate sentences about a rowing competition called the Boat Race, such as the sentence in \ref{ex:boatrace}: 

\ex. The Boat Race is a side-by-side rowing competition between the University of Oxford (sometimes referred to as the ``Dark Blues'') and the University of Cambridge (sometimes referred to as the ``Light Blues'').\label{ex:boatrace}

The reason for this tendency is that the training set is a collection of Wikipedia articles, and it happens to contain 159 articles about the Boat Race (one article for each year that the event has been held: ``The Boat Race 1861'', ``The Boat Race 1862'', etc.) There are several sentences that are repeated word-for-word across these articles, so it is no surprise that models have memorized these frequent sentences, even though they are not particularly notable sentences outside the context of this dataset.

\subsection{Architecture}

There are several perspectives in cognitive science which emphasize that, in order to understand an information-processing system, it is important to consider the computational resources that the system has at its disposal. Because the human brain is a type of neural network (albeit a type that is very different from the networks used in LLMs), many of these perspectives are couched in discussions of how our understanding of higher cognition should be informed by the fact that the mind emerges from the brain, as opposed to being explained solely using symbolic theories that are not rooted in neural networks; for an overview of perspectives on this topic, see \citet{smolensky1988proper}. 
For our purposes, the perspectives that are particularly noteworthy are those saying that, even when the goal is to understand higher-level cognitive functions, it remains important to view the mind through the lens of the neural architecture that it emerges from---where this architecture is sometimes taken to be the primary consideration for cognitive theories \citep{mcclelland1986appeal} and is sometimes taken to be something that should be unified with higher-level symbolic analyses \citep{smolensky2006harmony,smolensky2022neurocompositional}. Outside of debates about the role of neural networks, our perspective overlaps with resource rationality \citep{lieder2020resource}---the idea that cognition should be analyzed as the rational use of limited resources to accomplish a task. This view is a modification of more traditional rational analysis \citep{anderson1990adaptive}, which focuses on the task that one aims to perform but not on the resources that one has access to; since we consider not just training task but also model architecture, our approach is closer to a resource-rational view than a more traditional rational view.

Many lines of research in AI can be understood as ways to understand and improve models based on the limitations that their architectures impose. For instance, new modeling developments are typically motivated by insights about the limitations of previous architectures: RNNs \citep{elman1990finding}, unlike preceding systems, were able to process variable-length sequential data; LSTMs \citep{hochreiter1997long} overcame the vanishing gradient problem inherent in non-gated RNNs; Transformers \citep{vaswani2017attention} afforded more efficient training than LSTMs; and current research now seeks to overcome the context-length limitations of standard Transformers by enabling them to efficiently handle longer contexts \citep{Beltagy2020Longformer,ainslie2023colt5}. Below we specify two embers of autoregression that can be attributed to model architecture.

\subsubsection{Sensitivity to tokenization and other aspects of input formatting}

All models that take in data need to have that data represented in some way. The choice of input format can have important implications for model behavior.
For instance, \citet{pinker1988language} provided an influential early criticism of neural-network modeling, focused on the model of \citet{rumelhart1986learning}, which was trained to take in an English verb and produce its past-tense form. Some of their points identify ways in which the model's input format makes it fundamentally incapable of handling certain cases; for instance, it does not distinguish between different words that are pronounced the same, meaning that it cannot capture cases where two identically-pronounced words have different past tenses (e.g., \textit{break} and \textit{brake} have the same pronunciation, yet their past tenses are pronounced differently: \textit{broke} vs.\ \textit{braked}). 

In more recent models, the way in which the input is formatted continues to be important. Current models of language typically use subword tokenization schemes \citep{sennrich2016neural} in which each word is encoded as a sequence of one or more tokens, each of which can be made of one or more contiguous letters. One consequence of this choice is that some languages are more expensive to process than others due to differences in how straightforwardly this approach can handle a language's writing system \citep{ahia2023all}; another consequence is that models can be brittle to perturbations (e.g., typos) that cause a word to be broken into unfamiliar token sequences, whereas approaches that do not rely on subword tokens can be more robust to such perturbations \citep{salesky2021robust}.
The fact that our tasks were challenging for GPT models may also relate to their tokenization scheme: though GPT-3.5 and GPT-4 do encode the spellings of their tokens (see Section~\ref{sec:tasks}), this information may be less readily accessible to them than it would be to models in which characters are fully listed in the input, which would be a source of difficulty in tasks that involve character-level manipulations (e.g., rot-13 decoding).

\subsubsection{Limited compositionality and systematicity}\label{sec:systematicity}

\citet{fodor1988connectionism} and \citet{marcus1998rethinking} both presented influential arguments for why neural networks, due to the way they operate, are likely to struggle with the related phenomena of compositionality (combining familiar structures in novel ways) and systematicity (handling related stimuli in unified ways). 
Compositionality and systematicity have continued to be important targets of neural network evaluation \citep{lake2018scan,hupkes2020compositionality,kim2020cogs}. Although models have shown some impressive advances in evaluations of compositionality \citep[e.g.,][]{zhou2023leasttomost}, fully compositional behavior continues to elude them in at least some settings \citep{dziri2023faith}, motivating proposals that aim to enhance neural networks in ways that support symbolic abstractions \citep[e.g.,][]{schlag2019enhancing,webb2021emergent}.

We do not claim that neural networks are fundamentally incapable of compositional or systematic behavior. On the contrary, theoretical work has proven that neural networks are capable of capturing these properties \citep{smolensky1990tensor}.  Our claims are not therefore about what is possible; rather, they are about what is straightforward. Even though neural networks \textit{can} behave compositionally and systematically, it is not straightforward for them to do so, so it can be expected that their handling of these phenomena will encounter some difficulties.

\subsection{Beneficial embers of autoregression}

In discussing the ways in which models are shaped by the problem they solve, we have mostly discussed negative consequences, because we believe that the positive consequences are already well-recognized; indeed, they are some of the reasons why certain tasks, training distributions, and architectures have become prevalent.

First, the next-word prediction objective provides a major benefit by virtue of its generality \citep{malach2023auto}: almost any task can be cast in terms of next-word prediction, such that training a model on this task will also give it training on a wide range of other tasks \citep{radford2019language}. For instance, predicting what words will come after the phrase \textit{In summary} requires a model to summarize the preceding text, giving it experience in the task of summarization. 

Second, Internet text is useful as a training distribution because it gives models exposure to a wide range of types of examples. As a result, very few types of input are out-of-distribution for models, mitigating the ways in which neural networks are brittle to out-of-distribution examples.

Finally, the use of a neural network  as the model architecture has several positive attributes compared to a symbolic approach, including similarity-based generalization, multiple constraint satisfaction, graceful degradation, and content-addressable memory \citep{mcclelland1986appeal}, as well as the ability to learn efficiently from naturalistic data \citep{lecun2015deep}.

\section{Conclusion}

Recent paper titles have made many statements about what language models are: 
\begin{itemize}
    \item ``Language Models are Unsupervised Multitask Learners'' \citep{radford2019language}
    \item ``Language Models are Few-Shot Learners'' \citep{brown2020language}
    \item ``Language Models are General-Purpose Interfaces'' \citep{hao2022language}
    \item ``Language Models are Multilingual Chain-of-Thought Reasoners'' \citep{shi2023language}
    \item ``Language Models are Open Knowledge Graphs'' \citep{wang2020language}
\end{itemize}
We should absolutely recognize and explore these advanced properties. Nonetheless, we should also remember a simpler fact: language models are...language models!
That is, they are statistical next-word prediction systems. Recognizing this fact highlights some important properties: for instance, as we have shown, language models have greater difficulty with infrequent tasks than frequent ones, even when comparing two tasks that seem equally complex to a human; and they have greater difficulty on examples with low-probability answers than high-probability ones, even when the task is deterministic. 
Both of these properties---as well as several others that we have discussed above---can be attributed to the way that LLM training focuses on the statistics of word sequences. In sum, our efforts to understand language models should consider both their sparks of artificial general intelligence and their embers of autoregression.

\section*{Limitations}

Throughout the paper, we have noted several limitations of the general approach we have proposed as well as the specific experiments we have run. Here we collect the most important limitations. First, the teleological perspective that we have argued for can be incomplete. Therefore, we do not claim that teleological analysis should be the only way that the field studies LLMs; rather, it should be one of many perspectives that can inform analyses. Second, the teleological perspective can be misleading: our analyses of a model's goal may be incorrect. For this reason, it is critical that teleologically-derived predictions should not be assumed to be correct but should instead be tested empirically, as we have done.

An important limitation of our specific experiments is that the models we tested---which we selected for the reasons given in Section~\ref{sec:models}---are proprietary ones for which many details are not publicly available (e.g., the precise model architecture and training data). We have nonetheless shown that the information we do have is sufficient to reason teleologically about these systems and to identify important sources of difficulty for them.  An additional limitation is that the tasks we investigated are relatively simple ones that do not have substantial practical utility. We chose these tasks because our goal was to understand the general principles that drive LLM performance, and for the purpose of understanding a system it is useful to use simple tasks so that we can reason clearly about what capabilities those tasks require \citep{efrat2022lmentry}. We leave for future work an investigation of whether and how the effects that we have identified also influence model performance in more practical settings.

\section*{Acknowledgments}

We thank Robert Frank, Tania Lombrozo, Dakota McCoy, Paul Smolensky, and Cameron Turner for helpful discussion. We also thank \citet{bubeck2023sparks} for making their LaTex code publicly available, as our paper uses many aspects of their formatting. Any errors are our own. 
RTM was supported by the National Science Foundation SBE Postdoctoral Research Fellowship under Grant No.\ 2204152, SY was supported by the Harold W. Dodds Fellowship from Princeton University, and MH was supported by the US Department of Defense through a National Defense Science and Engineering Graduate Fellowship. 
Any opinions, findings, and conclusions or recommendations expressed in this material are those of the authors and do not necessarily reflect the views of the National Science
Foundation or the US Department of Defense.

\newpage
\bibliographystyle{aaai24}
\bibliography{main}

\newpage
\appendix
\addtocontents{toc}{\protect\setcounter{tocdepth}{1}}

\section{Dataset details}\label{sec:datasets}

\subsection{Sample sizes}\label{app:samplesize}

Table~\ref{tab:samplesize} specifies how many examples there were in each dataset that we created for evaluating models.

\begin{table}[]
    \centering
    \begin{tabular}{p{3.3cm}l} \toprule
        Task & Sample size \\ \midrule
        Article swapping & 100 \\
        Reversal & 100 \\
        Counting & See caption \\
        Acronyms & 1000 \\
        Linear functions & 100 \\
        Multiplication & 100 \\
        Sorting & 100 \\
        Keyboard cipher & 100 \\
        Shift cipher & 100 \\
        Pig Latin & 100 \\
        Birthdays & 100 \\ \bottomrule
    \end{tabular}
    \caption{Sample sizes for our datasets. Each number shows the sample size for one condition, corresponding to a single point in each of our line plots or a single bar in each of our bar plots; e.g., for reversal, we had 3 conditions, yielding a total of 300 examples. For counting, there were different sample sizes for different settings. The results in the top right of Figure~\ref{fig:counting_freq_magnitude} use a sample size of 1200 per bar; the plots in the middle and bottom of that figure use a sample size of 30 per point; and the results in Figure~\ref{fig:counting_inp} use a sample size of 3000 per bar.}
    \label{tab:samplesize}
\end{table}

\subsection{Shift cipher sentences}\label{app:rot13_sentences}

On June 16, 2023, we downloaded the 50 most recent new articles from Global Voices\footnote{\url{https://globalvoices.org/}} (spanning from June 2, 2023, to June 16, 2023), to minimize the chances that models would have seen these particular sentences---though it would not cause problems for our conclusions if they had because we intended for these sentences to be high-probability for the models. We selected Global Voices because its content has a permissive license that allows sharing and modification, with attribution.\footnote{\url{https://globalvoices.org/about/global-voices-attribution-policy/}} A full list of the articles we used is at \url{https://github.com/tommccoy1/embers-of-autoregression/blob/main/global_voices_articles.md}.
We then sentence-tokenized these articles using the NLTK sentence tokenizer \citep{bird2009nltk}. For each sentence, we used GPT-2 \citep{radford2019language} to find the maximum perplexity of any token in that sentence and ranked sentences in order of increasing maximum perplexity; the goal was to find sentences in which no individual words were overly surprising. We then selected the first 100 sentences in this list that met the following criteria:
\begin{itemize}
    \item Did not include proper names (except for very common ones such as the names of countries)
    \item Did not discuss violence, death, or sex.
    \item Did not have grammatical errors.
\end{itemize}
The resulting 100 sentences served as the high-probability sentences for being the output of shift cipher decryption or the input for shift cipher encryption.

To produce medium-probability sentences, we started with the high-probability sentences and then modified each one in a way that reduced its probability. 
Specifically, we tokenized the sentence using the RoBERTa-large tokenizer  and then masked one token at a time, from left to right. For each masked token, we sampled a replacement token from RoBERTa-large \citep{liu2019roberta}, where the replacement token had to meet the following criteria:
\begin{itemize}
    \item Must be within the top 95\% of the probability mass
    \item Must not be within the top 90\% of the probability mass
    \item Must be the same number of characters as the token being replaced
\end{itemize}
The first two criteria were intended to find replacements that were low-probability but not so low-probability as to be ungrammatical. The last criterion served to preserve the length of the sentence. If there were no tokens that fit all criteria, the token was left unchanged. Once a new token was sampled, the sentence was changed to incorporate it before the next token's replacement was sampled. The full sentence was required to have the same number of subword tokens as the original, measured using the tokenizer that is used by both GPT-3.5 and GPT-4.\footnote{The tokenizer is \texttt{cl100k\_base} from the \texttt{tiktoken} package: \url{https://github.com/openai/tiktoken}} This procedure did not always produce grammatical sentences; to ensure grammaticality, for each sentence, we produced 20 candidates (ranked by perplexity) and manually chose the highest-perplexity one that was grammatical. In a few cases, we had to generate more than 20 candidates or had to put together parts of multiple candidates. 

To create each low-probability sentence, we started with a medium-probability sentence and then shuffled all of its words except the first word and last word. For this purpose, words were defined as whitespace-delimited units. The first word and last word were left in place to avoid unusual capitalization or punctuation.

To create the targeted sentences which differ from the high-probability ones by just one or two characters, we started with the high-probability sentences and then manually replaced one word in each one with another word. The new word had to have a Levenshtein edit distance \citep{levenshtein1966binary} of at most 2 from the original word (where one edit is defined as a single-character insertion, deletion, or substitution). The new word was selected to be one that was grammatical yet unlikely in the given context. In the codebase, these sentences are referred to as \textit{adversarial}.

The prompts used for rot-13 (Appendices~\ref{sec:prompt_rot13_enc} and \ref{sec:prompt_rot13_dec}) included one in-context example that clarified two potential points of confusion:
\begin{itemize}
    \item It indicated that capitalization was retained in the output.
    \item It indicated that non-letter characters were to be left unchanged.
\end{itemize}

\noindent
In all conditions, there were 100 examples. Table~\ref{tab:dataset_stats} shows some statistics for these examples that were used in our regression tests. The tokens were counted using the tokenizer used by GPT-3.5 and GPT-4. The log probabilities were estimated as the log probability assigned by GPT-2 to the sentences.

\renewcommand{\arraystretch}{2}
\begin{table*}[]
    \centering
    \begin{tabular}{p{3cm}cccccc} \toprule
        & \multicolumn{1}{C{1.3cm}}{Input chars.} & \multicolumn{1}{C{1.3cm}}{Output chars.} & \multicolumn{1}{C{1.3cm}}{Input tokens} & \multicolumn{1}{C{1.3cm}}{Output tokens} & \multicolumn{1}{C{1.3cm}}{Input logprob} & \multicolumn{1}{C{1.3cm}}{Output logprob} \\ \midrule
       \raggedright Rot-13 decryption: High-probability & 98.7 & 98.7 & 46.4 & 19.7 & -302.6 & -58.4 \\
       \raggedright Rot-13 decryption: Medium-probability & 98.7 & 98.7 & 46.8 & 19.7 & -304.5 & -86.3 \\
       \raggedright Rot-13 decryption: Low-probability & 98.7 & 98.7 & 46.8 & 19.7 & -306.9 & -144.5 \\ \midrule
       \raggedright Pig Latin decryption: High-probability & 118.3 & 88.1 & 40.8 & 17.6 & -256.3 & -62.5 \\
       \raggedright Pig Latin decryption: Medium-probability & 118.3 & 88.1 & 40.8 & 17.6 & -256.6 & -86.9 \\
       \raggedright Pig Latin decryption: Low-probability & 118.3 & 88.1 & 40.8 & 17.6 & -258.2 & -132.3 \\ \midrule
       \raggedright Reversal decryption: High-probability & 98.7 & 98.7 & 20.0 & 19.7 & -140.6 & -58.4 \\
       \raggedright Reversal decryption: Medium-probability & 98.7 & 98.7 & 20.1 & 19.7 & -149.7 & -86.3 \\
       \raggedright Reversal decryption: Low-probability & 98.7 & 98.7 & 20.1 & 19.7 & -148.4 & -144.5 \\
       \bottomrule
    \end{tabular}
    \caption{Average dataset statistics for some of the datasets discussed in Appendix~\ref{sec:datasets}. For the encryption versions of the decryption tasks, the numbers are the same, but the input and output values are swapped. For the decryption datasets, the intent was to vary output log probability yet keep all other factors similar; for encryption, the goal was the opposite.}
    \label{tab:dataset_stats}
\end{table*}

\renewcommand{\arraystretch}{1.0}

\subsection{Reversal sentences and keyboard cipher sentences}

For the reversal task and keyboard cipher task, we used the same sentences as for shift ciphers (Appendix~\ref{app:rot13_sentences}). As with the shift ciphers, there were 100 sentences per condition.

\subsection{Pig Latin sentences}

The Pig Latin sentences were generated with the same basic technique as the rot-13 sentences (Appendix~\ref{app:rot13_sentences}). However, Pig Latin is more constrained than rot-13 because Pig Latin is mainly a spoken phenomenon, so when it is written down there can be some ambiguity due to the ways in which English spelling does not match English pronunciation. E.g., should the Pig Latin for \textit{hour} be \textit{ouray} or \textit{ourhay} or \textit{houray}? 
Therefore, we excluded any sentences with words that created any such ambiguities. This included abbreviations, numbers, and words (such as \textit{hour}) for which the initial consonant cluster in writing does not match well with the initial consonant cluster in speech---such mismatches were automatically flagged using the CMU Pronouncing Dictionary\footnote{\url{http://www.speech.cs.cmu.edu/cgi-bin/cmudict}} and then manually checked. Thus, to generate Pig Latin sentences, we followed the same procedure as for rot-13 but discarded any examples that failed to meet these criteria, and sampled new examples until we had the requisite 100 examples per condition. All inputs and outputs were lowercased to avoid ambiguity about how to deal with capitalization (e.g., in Pig Latin, should \textit{Latin} become \textit{atinLay} or \textit{Atinlay} or \textit{atinlay}?). There were 100 examples per condition.

The prompts for the Pig Latin tasks (Appendices~\ref{app:pigenc_prompt} and \ref{app:pigdec_prompt}) include an in-context example that clears up several points of confusion:
\begin{itemize}
    \item It illustrates how to handle punctuation.
    \item It shows not to break words on apostrophes.
    \item It illustrates the handling of both consonant-initial and vowel-initial words (a distinction that is also described in words in the prompt).
\end{itemize}

\subsection{Article swapping sentences}

The article swapping sentences were generated with the same basic technique as the rot-13 sentences (Appendix~\ref{app:rot13_sentences}), but with additional constraints added. First, each sentence had to contain at least one article (\textit{a}, \textit{an}, or \textit{the}), since the task becomes trivial if there are no articles to be moved. Second, the sentence cannot start or end with an article because that could create issues when attempting to swap (e.g., if we are swapping each article with the previous word, and if the sentence starts with an article, then that initial article will not have anything to swap with). There were 100 examples per probability level.

\subsection{Acronyms}\label{app:acronym_data}

Our initial word list was created by extracting all seven-letter words in the CMU Pronouncing Dictionary that were two tokens long under the tokenizer used by GPT-3.5 and GPT-4; the token count had to be two both when the word was preceded by a space and when it was not. This word list had two versions---one in all lowercase letters and one in all capital letters. The lists differed because the number of tokens is often different between lowercase and uppercase versions of a word. Each of these word lists was ordered from highest to lowest probability, where the probability in question was the probability that GPT-2 assigned to the word in the context \textit{The word is ``WORD''}---chosen to provide a context in which any word should fit reasonably well, and using quotation marks to clearly indicate the word's boundaries. 

Each word list was then split into bins by probability. Certain words had to be excluded because no acronyms could be formed from them (e.g., due to a lack of words starting with X, no output could contain an X). After removing such words, there were 8203 lowercase words and 2244 all-caps words. These were split into bins as shown in Table~\ref{tab:vocab_bins}; there was some space left between bins to make them more distinct, and we also left out the words at the end of each list since their low-probability under GPT-2 might reflect that something is unusual about them. 

When varying the input probability, the output words were fixed across conditions---i.e., spelling the same word but from input words of differing probabilities. When comparing first-letter and second-letter acronyms, the same output words were used in the two conditions.
In all comparisons (varying output probability, varying input probability, and comparing first-letter vs.\ second-letter acronyms), each word in the $n^{\text{th}}$ example had the same split point as the corresponding word in the $n^{\text{th}}$ example in all other conditions, where the \textit{split point} is the index of the break in between the two tokens within the word; we controlled for this factor in case different split points lead to different difficulty levels. Within these constraints, examples were generated by first sampling an output word and then, for each letter in that word, sampling an input word that would yield that letter. We generated 1000 examples per condition.

To determine the log probabilities of the output words (for purposes of plotting accuracy vs.\ log probability, or for purposes of statistical tests), we used the log probability that GPT-2 assigned to the whole sequence \textit{The word is ``WORD''} minus the log probability that it assigned to \textit{The word is ``}; this gives the probability of the word plus the closing quotation mark conditioned on \textit{The word is ``}. We included the closing quotation mark because some effects of the word's probability might manifest in the following quotation mark (as the indicator that the word has ended). 
To compute the log probability of the input, we used this same approach to get the log probability of each input word, and we then summed the log probabilities of all the input words to yield the log probability of the whole input.

\begin{table}[]
    \centering
    \begin{tabular}{ccc} \toprule
         & Start & End \\ \midrule
        Lowercase bin 1 & 0 & 999 \\
        Lowercase bin 2 & 1500 & 2499 \\
        Lowercase bin 3 & 3000 & 3999 \\
        Lowercase bin 4 & 4500 & 5499 \\
        Lowercase bin 5 & 6000 & 6999 \\ \midrule
        Capital bin 1 & 0 & 399 \\
        Capital bin 2 & 440 & 839 \\
        Capital bin 3 & 880 & 1279 \\
        Capital bin 4 & 1320 & 1719 \\
        Capital bin 5 & 1760 & 2159 \\ \bottomrule
    \end{tabular}
    \caption{Acronym vocabulary binning.}
    \label{tab:vocab_bins}
\end{table}

\subsection{Counting}\label{app:counting_details}

In the case of counting characters, we used lowercase letters as the common characters and emojis as the rare characters. In the case of counting words, we extracted all words from the CMU Pronouncing Dictionary that fit the following criteria:
\begin{itemize}
    \item Must be 1, 2, or 3 tokens long.
    \item Must have the same number of tokens whether preceded by a space or not.
    \item Must appear at least 20 times in the Wikitext-103 corpus \citep{merity2016pointer}.
\end{itemize}

\noindent
We then sorted all these words by the probability assigned to them by GPT-2 in the context \textit{The word is ``WORD''}, as described in Appendix~\ref{app:acronym_data}. Then, we split the resulting words into the highest-probability 8000 words and the rest (this divided the list roughly in half, as there were about 16,000 words meeting our criteria). To create our candidate high-probability words, we took the 150 highest-probability words for each token count of 1, 2, or 3, after manually excluding profanity and words that could conceivably be considered not words (e.g., because they are names); since we were asking models to count words, we wanted to avoid an interpretation where some ``words'' were ignored on the grounds that they were not in fact words. We used the same procedure to create our list of rare words, except that in this case we used the 150 lowest-probability words of each token length. This gave us a list of 450 high-probability words and 450 low-probability words.

For each integer from 1 to 100, we created 30 examples in each condition. To create a character example for count $n$, we uniformly sampled a character from the possible candidates and then repeated it $n$ times (that is, within each character example, all characters were the same). To create a word example, we sampled $n$ words without replacement; the words were matched for token patterns across common/rare conditions (e.g., if the ``common words'' example had a word with 3 tokens of length 2, 5, and 7, the ``rare words'' example would also have a word in that position with the same token pattern). We sampled words without replacement so that, within each word sequence, there were no duplicated words, to avoid any ambiguity about whether we intended for the count to include duplicates or not.

The frequency of each number was estimated by counting its occurrences in the first one billion words of the C4 corpus \citep{raffel2020exploring}. In the experiment where we compared common and rare numbers, a number $n$ was categorized as common if it occurred at least twice as many times as each of $n-2$, $n-1$, $n+1$, and $n+2$; this yielded twelve common numbers (20, 30, 40, 45, 50, 60, 70, 75, 80, 90, 95, 100). We then used $n-2$, $n-1$, $n+1$, and $n+2$ as rare numbers. We generated 100 examples for each common number and 25 examples for each rare number; this setup guaranteed that the common-number and rare-number conditions were matched for average magnitude of the counts.

\subsection{Sorting}

For sorting both words and numbers, the input was a list of length 10 to 20 inclusive; the list could not contain repeats. Each example was generated by first selecting a length (by uniform sampling) and then randomly selecting (without replacement) list elements from the set of candidates. For sorting words, the candidates were the 450 common words used in the counting task (Appendix~\ref{app:counting_details}). For sorting numbers, the candidates were the integers from 1 to 10,000 inclusive. We generated 100 examples for each condition.

\subsection{Three-digit multiplication}

We generated 100 multiplication problems in which both numbers being multiplied were three-digit integers. The two numbers were sampled uniformly from the range [100,999]. The same set of 100 problems was used across all methods of formatting the text.

\subsection{Linear functions}\label{app:lineardescription}

We generated 100 examples for all conditions.
For the initial test, in which we compared $(9/5)x + 32$ to $(7/5)x + 31$ using the same set of inputs, we sampled 100 input values uniformly from the integers 0 to 999 inclusive. For the test in which we instead held the outputs constant, we sampled 100 output values from the integers 0 to 999 inclusive and then, for each function, used the integer input value---if there was one that satisfied our constraints---which yielded that output value after the function had been applied and the output had been rounded to the nearest integer. If either function did not have a suitable integer input value for a given output, we rejected that output and sampled another one.
For the out-of-distribution tests, we added two additional constraints:
\begin{itemize}
    \item The first number that was sampled (i.e., the Celsius input in the initial version of the test, or the Fahrenheit output in the inverse version) had to be in the range 501 to 999, inclusive.
    \item Neither number could be evenly divisible by 10.
\end{itemize}
We chose these constraints based on a corpus analysis of the C4 dataset \citep{raffel2020exploring} in which we aimed to determine the types of temperatures that are frequently present with conversions in an Internet corpus. In this analysis, we identified all sentences that contained both the word ``fahrenheit'' preceded by a number and the word ``celsius'' preceded by a number, or the characters ``°f'' preceded by a number and the characters ``°c'' preceded by a number, and assumed that the two numbers were a Fahrenheit and Celsius temperate being presented as conversions of each other. (We ignored case when searching for these strings). 
Of these 10,000 occurrences, 99.8\% are excluded by applying our two constraints when the first constraint is applied to the Celsius temperature, and 99.7\% are excluded by applying the constraints when the first constraint is applied to the Fahrenheit temperature. Thus, by enforcing these constraints, we can yield only temperatures of a type that tend to occur very rarely with a conversion in natural Internet text, reducing---but certainly not removing---the risk that models might have seen and copied it.

\subsection{Birthdays}\label{app:birthdayexampledetails}

We first extracted all name/birth date pairs from the WikiBio dataset \citep{lebret2016wikibio}, restricting the set to names containing exactly two words and containing only the 26 basic letters of the Roman alphabet. 
We then counted how many times each name occurred in the first 100 million lines (approximately 3.6 billion words) of the C4 corpus \citep{raffel2020exploring}. Using this frequency information, we created four sets of names. The first set contained names mentioned 2,000 times or more in our sample of C4; the second set contained names mentioned between 200 and 1,999 times inclusive; the third set contained names mentioned between 20 and 199 times inclusive; and the fourth set contained names mentioned between 5 and 19 times inclusive. We only included names that were unambiguous on Wikipedia (i.e., when the name is searched for, Wikipedia goes directly to the person's page, without redirecting to a disambiguation page, and the top of the resulting page does not include a link to a disambiguation page); we checked for ambiguity by manually searching for each candidate name on Wikipedia until we had gathered sufficiently many unambiguous ones. We also verified that, for each person we included, Wikipedia included that person's birthday, and the birthday listed by Wikipedia matched the one listed in the WikiBio dataset. Each set contains 100 names, and we matched the birth dates across sets (i.e., the $n^{\text{th}}$ names across all sets have the same birth dates; e.g., in each of the four sets, the eighteenth person was born on September 27, 1972).

\section{Prompts}\label{app:prompts}

\subsection{Shift ciphers: Encryption}\label{sec:prompt_rot13_enc}

Rot-13 is a cipher in which each letter is shifted 13 positions forward in the alphabet. For example, here is a message and its corresponding version in rot-13:

\noindent
Original text: ``Stay here!''

\noindent
Rot-13 text: ``Fgnl urer!''

\bigskip
\noindent
Here is another message. Encode this message in rot-13:

\noindent
Original text: ``To this day, we continue to follow these principles.''

\noindent
Rot-13 text:

\subsection{Shift ciphers: Decryption}\label{sec:prompt_rot13_dec}

Rot-13 is a cipher in which each letter is shifted 13 positions forward in the alphabet. For example, here is a message written in rot-13 along with the original text that it was created from:

\noindent
Rot-13 text: ``Fgnl urer!''

\noindent
Original text: ``Stay here!''

\bigskip
\noindent
Here is another message in rot-13. Decode this message to produce the original text:

\noindent
Rot-13 text: ``Gb guvf qnl, jr pbagvahr gb sbyybj gurfr cevapvcyrf.''

\noindent
Original text:

\subsection{Comments on step-by-step and chain-of-thought prompts}\label{app:stepcotprompts}

We used an initial informal exploration of several options to select the specific step-by-step and chain-of-thought prompts that we used. 
For step-by-step prompting, we compared two approaches: adding the sentence \textit{Make sure to think step by step} into the middle of the prompt, or adding the sentence \textit{Let's think step by step} at the end of the prompt. We found the former to be more successful, so we used it for step-by-step prompting (Appendices~\ref{sec:stepencprompt} and \ref{sec:stepdecprompt}). For chain-of-thought prompting, we tried the following options:
\begin{itemize}
    \item Convert one word at a time (e.g., given \textit{Fgnl urer!}, first encode \textit{Fgnl} and then \textit{urer!})
    \item Convert one letter at a time, each on its own numbered line of the generation (there were three variants: (i) do not include spaces; (ii) leave a blank line for a space; or (iii) leave a numbered line for a space exactly as if it were a letter).
    \item Convert one word at a time, but within the word, break it down into letters.
    \item Move the entire sentences back one position in the alphabet at a time. E.g., given \textit{Fgnl urer!}, first shift it to \textit{Efmk tqdq!}, then \textit{Delj spcp!}, etc.\
\end{itemize}
We were expecting the last option to perform the best, but we instead found that one of the letter-based variants performed the best, so we used that (Appendices~\ref{sec:chainencprompt} and \ref{sec:chaindecprompt}).

\subsection{Shift ciphers step-by-step: Encryption}\label{sec:stepencprompt}

Rot-13 is a cipher in which each letter is shifted 13 positions forward in the alphabet. For example, here is a message and its corresponding version in rot-13:

\noindent
Original text: ``Stay here!''

\noindent
Rot-13 text: ``Fgnl urer!''

\bigskip
\noindent
Here is another message. Encode this message in rot-13. Make sure to think step by step. On the last line, write the words ``Rot-13 text:'' followed by the encoded message:

\noindent
Original text: ``But this time, there may also be another reason.''

\subsection{Shift ciphers step-by-step: Decryption}\label{sec:stepdecprompt}

Rot-13 is a cipher in which each letter is shifted 13 positions forward in the alphabet. For example, here is a message written in rot-13 along with the original text that it was created from:

\noindent
Rot-13 text: ``Fgnl urer!''

\noindent
Original text: ``Stay here!''

\bigskip
\noindent
Here is another message in rot-13. Decode the message to produce the original text. Make sure to think step by step. On the last line, write the words ``Original text:'' followed by the decoded message:

\noindent
Rot-13 text: ``Ohg guvf gvzr, gurer znl nyfb or nabgure ernfba.''

\subsection{Shift ciphers chain-of-thought: Encryption}\label{sec:chainencprompt}

Rot-13 is a cipher in which each letter is shifted 13 positions forward in the alphabet. For example, here is a message to be encoded:

\noindent
Original text: ``Stay here!''

\bigskip
\noindent
To encode this message, we shift each letter 13 positions forward:

\noindent
1. S -> F

\noindent
2. t -> g

\noindent 
3. a -> n

\noindent 
4. y -> l

\noindent5.   ->  

\noindent6. h -> u

\noindent7. e -> r

\noindent8. r -> e

\noindent9. e -> r

\noindent10. ! -> !

\bigskip
\noindent
Therefore, the rot-13 text is: ``Fgnl urer!''

\bigskip
\noindent
Here is another message. Encode the message one letter at a time. On the last line, write the words ``Rot-13 text:'' followed by the encoded message:

\noindent
Original text: ``But this time, there may also be another reason.''

\subsection{Shift ciphers chain-of-thought: Decryption}\label{sec:chaindecprompt}

Rot-13 is a cipher in which each letter is shifted 13 positions forward in the alphabet. For example, here is a message written in rot-13:

\noindent
Rot-13 text: ``Fgnl urer!''

\bigskip
\noindent
To decode this message, we shift each letter 13 positions backward:

\noindent1. F -> S

\noindent2. g -> t

\noindent3. n -> a

\noindent4. l -> y

\noindent5.   ->  

\noindent6. u -> h

\noindent7. r -> e

\noindent8. e -> r

\noindent9. r -> e

\noindent10. ! -> !

\bigskip
\noindent
Therefore, the original text is: ``Stay here!''

\bigskip
\noindent
Here is another message in rot-13. Decode the message one letter at a time. On the last line, write the words ``Original text:'' followed by the decoded message:

\noindent
Rot-13 text: ``Ohg guvf gvzr, gurer znl nyfb or nabgure ernfba.''

\subsection{Reversal: Encryption}

Reverse the following sequence of words. Punctuation marks should be moved along with the words they are attached to; e.g., the reversed version of ``Good morning, everyone!'' would be ``everyone! morning, Good'': ``But this time, there may also be another reason.''

\subsection{Reversal: Decryption}

Reverse the following sequence of words. Punctuation marks should be moved along with the words that they are attached to; e.g., the reversed version of ``everyone! morning, Good'' would be ``Good morning, everyone!'': ``letter. sons, may another also be there with Yet''

\subsection{Pig Latin: Encryption}\label{app:pigenc_prompt}

There is a secret way of talking called Pig Latin in which you move the first consonant cluster of each word to the end of the word, and then add -ay to the end of the word. If the word starts with a vowel, you simply add -ay to the end without otherwise changing the word. For example, below is a sentence in English and its equivalent in Pig Latin:

\noindent
English: ``the frogs aren't noisy.''

\noindent
Pig Latin: ``ethay ogsfray aren'tay oisynay.''

\bigskip
\noindent
Write this sentence in Pig Latin:

\noindent
English: ``turkey almost exclusively relies on energy imports as its domestic output covers only three percent of its energy needs.''

\noindent
Pig Latin:

\subsection{Pig Latin: Decryption}\label{app:pigdec_prompt}

There is a secret way of talking called Pig Latin in which you move the first consonant cluster of each word to the end of the word, and then add -ay to the end of the word. If the word starts with a vowel, you simply add -ay to the end without otherwise changing the word. For example, below is a sentence in Pig Latin and its equivalent in English:

\noindent
Pig Latin: ``ethay ogsfray aren'tay oisynay.''

\noindent
English: ``the frogs aren't noisy.''

\bigskip
\noindent
Convert this sentence from Pig Latin to English:

\noindent
Pig Latin: ``urkeytay almostay exclusivelyay eliesray onay energyay importsay asay itsay omesticday outputay overscay onlyay eethray ercentpay ofay itsay energyay eedsnay.''

\noindent
English:

\subsection{Article swapping: Decoding, swap with previous}

Repeat the sentence listed as Input 2 below, but every time there is an article (``the'', ``a'', or ``an''), swap it with the previous word. For example, if the input were Input 1, you should reply with Output 1:

\noindent
Input 1: ``When we boarded train the, we saw strange a sight.''

\noindent
Output 1: ``When we boarded the train, we saw a strange sight.''

\bigskip
\noindent
Input 2: ``Well, it was not exactly planned from beginning the.''

\noindent
Output 2:

\subsection{Article swapping: Decoding, swap with next}

Repeat the sentence listed as Input 2 below, but every time there is an article (``the'', ``a'', or ``an''), swap it with the following word. For example, if the input were Input 1, you should reply with Output 1:

\noindent
Input 1: ``When we the boarded train, we a saw strange sight.''

\noindent
Output 1: ``When we boarded the train, we saw a strange sight.''

\bigskip
\noindent
Input 2: ``Well, it was not exactly planned the from beginning.''

\noindent
Output 2:

\subsection{Article swapping: Encoding, swap with previous}

Repeat the sentence listed as Input 2 below, but every time there is an article (``the'', ``a'', or ``an''), swap it with the previous word. For example, if the input were Input 1, you should reply with Output 1:

\noindent
Input 1: ``When we boarded the train, we saw a strange sight.''

\noindent
Output 1: ``When we the boarded train, we a saw strange sight.''

\bigskip
\noindent
Input 2: ``Well, it was not exactly planned from the beginning.''

\noindent
Output 2:

\subsection{Article swapping: Encoding, swap with next}

Repeat the sentence listed as Input 2 below, but every time there is an article (``the'', ``a'', or ``an''), swap it with the following word. For example, if the input were Input 1, you should reply with Output 1:

\noindent
Input 1: ``When we boarded the train, we saw a strange sight.''

\noindent
Output 1: ``When we boarded train the, we saw strange a sight.''

\bigskip
\noindent
Input 2: ``Well, it was not exactly planned from the beginning.''

\noindent
Output 2:

\subsection{Acronyms}

What sequence of letters is created when you combine the first letters of the words in the sequence ``chevron outback nomadic ferrous entitle splurge surfeit''? Write your answer in capital letters, with no spaces or punctuation.

\subsection{Counting: Words}

How many words are in the following list? ``stilted hyperbolic secret charlatan normative malevolent flippant lawless gullible flagrant albatross purposeful''

\subsection{Counting: Letters}

How many letters are in the following list? ``zzzzzzzzzzzzzz''

\subsection{Counting: Emojis}

How many tortoise emojis are in the following list? ``\tortoise{}\tortoise{}\tortoise{}\tortoise{}\tortoise{}\tortoise{}\tortoise{}\tortoise{}\tortoise{}\tortoise{}\tortoise{}\tortoise{}''

\subsection{Sorting words: Alphabetical order}

Sort the following list of words in alphabetical order: ``illogical, hard, scoundrel, over, sequester, irrefutable, lost, anathema, raucous, wondrous''

\subsection{Sorting words: Reverse alphabetical order}

Sort the following list of words in reverse alphabetical order: ``illogical, hard, scoundrel, over, sequester, irrefutable, lost, anathema, raucous, wondrous''

\subsection{Sorting numbers: Ascending order}

Sort the following list of numbers in ascending order: ``1292, 3475, 8774, 7076, 3695, 6441, 6298, 9762, 1916, 6988, 8084''

\subsection{Sorting numbers: Descending order}

Sort the following list of numbers in descending order: ``1292, 3475, 8774, 7076, 3695, 6441, 6298, 9762, 1916, 6988, 8084''

\subsection{Three-digit multiplication: Number}

Evaluate the mathematical expression in Question 2. For example, if you were asked Question 1, you would answer with Answer 1.

\noindent
Question 1: 831 times 447

\noindent
Answer 1: 371457

\bigskip
\noindent
Question 2: 285 times 796

\noindent
Answer 2:

\subsection{Three-digit multiplication: Lowercase}

Evaluate the mathematical expression in Question 2. For example, if you were asked Question 1, you would answer with Answer 1.

\noindent
Question 1: eight hundred thirty-one times four hundred forty-seven

\noindent
Answer 1: 371457

\bigskip
\noindent
Question 2: two hundred eighty-five times seven hundred ninety-six

\noindent
Answer 2:

\subsection{Three-digit multiplication: All caps}

Evaluate the mathematical expression in Question 2. For example, if you were asked Question 1, you would answer with Answer 1.

\noindent
Question 1: EIGHT HUNDRED THIRTY-ONE times FOUR HUNDRED FORTY-SEVEN

\noindent
Answer 1: 371457

\bigskip
\noindent 
Question 2: TWO HUNDRED EIGHTY-FIVE times SEVEN HUNDRED NINETY-SIX

\noindent
Answer 2:

\subsection{Three-digit multiplication: Alternating capitals}

Evaluate the mathematical expression in Question 2. For example, if you were asked Question 1, you would answer with Answer 1.

\noindent
Question 1: eIgHt HuNdReD tHiRtY-oNe times fOuR hUnDrEd FoRtY-sEvEn

\noindent 
Answer 1: 371457

\bigskip 
\noindent 
Question 2: tWo HuNdReD eIgHtY-fIvE times sEvEn HuNdReD nInEtY-sIx

\noindent
Answer 2:

\subsection{Linear function}

Below is a number. Multiply it by 9/5 and then add 32. Your answer should be a single number:

\noindent
Input: 328

\noindent
Answer:

\subsection{Linear function with rounding}

Below is a number. Multiply it by 9/5 and then add 32. Your answer should be a single number rounded to the nearest integer:

\noindent
Input: 78

\noindent
Answer:

\subsection{Keyboard cipher: Basic}\label{app:promptkeyboardbasic}

The keyboard cipher is a cipher where each letter is replaced with the letter to the right of it on a QWERTY keyboard. If the letter does not have a letter to its right, then you should wrap around to the start of the row; e.g., ``p'' maps to ``q''. For example, here is a message to be encoded:

\noindent
Original message: ``Hello world!''

\bigskip
\noindent
To encode this message, we replace each letter with the one to the right of it on a QWERTY keyboard:

\noindent
1. The first word is ``Hello''

\noindent
\hspace{0.1cm} * H -> J
 
\noindent
\hspace{0.1cm}  * e -> r
 
\noindent
\hspace{0.1cm}  * l -> a
 
\noindent
\hspace{0.1cm}  * l -> a
 
\noindent
\hspace{0.1cm}  * o -> p
 
\noindent
\hspace{0.1cm}  So ``Hello'' becomes ``Jraap''
 
\noindent
2. The second word is ``world!''

\noindent
\hspace{0.1cm}  * w -> e
 
\noindent
\hspace{0.1cm}  * o -> p
 
\noindent
\hspace{0.1cm}  * r -> t
 
\noindent
\hspace{0.1cm}  * l -> a
 
\noindent
\hspace{0.1cm}  * d -> f
 
\noindent
\hspace{0.1cm}  * ! -> !
 
\noindent
\hspace{0.1cm}  So ``world!'' becomes ``eptaf!''
 
\bigskip 
\noindent
Therefore, the message in the keyboard cipher is: ``Jraap eptaf!''

\bigskip 
\noindent
Here is another message. Encode the message in the keyboard cipher, word by word and letter by letter. On the last line, write the words ``Keyboard cipher:'' followed by the encoded message:

\noindent
Original message: ``But this time, there may also be another reason.''

\subsection{Keyboard cipher: Keyboard provided}\label{app:promptkeyboardprovided}

The keyboard cipher is a cipher where each letter is replaced with the letter to the right of it on a QWERTY keyboard. If the letter does not have a letter to its right, then you should wrap around to the start of the row; e.g., ``p'' maps to ``q''. For reference, here are the three rows of a QWERTY keyboard:

\noindent
q w e r t y u i o p

\noindent
a s d f g h j k l

\noindent
z x c v b n m

\bigskip
\noindent
For example, here is a message to be encoded:

\noindent
Original message: ``Hello world!''

\bigskip
\noindent
To encode this message, we replace each letter with the one to the right of it on a QWERTY keyboard:

\noindent
1. The first word is ``Hello''

\noindent
\hspace{0.1cm} * H -> J
 
\noindent
\hspace{0.1cm}  * e -> r
 
\noindent
\hspace{0.1cm}  * l -> a
 
\noindent
\hspace{0.1cm}  * l -> a
 
\noindent
\hspace{0.1cm}  * o -> p
 
\noindent
\hspace{0.1cm}  So ``Hello'' becomes ``Jraap''
 
\noindent
2. The second word is ``world!''

\noindent
\hspace{0.1cm}  * w -> e
 
\noindent
\hspace{0.1cm}  * o -> p
 
\noindent
\hspace{0.1cm}  * r -> t
 
\noindent
\hspace{0.1cm}  * l -> a
 
\noindent
\hspace{0.1cm}  * d -> f
 
\noindent
\hspace{0.1cm}  * ! -> !
 
\noindent
\hspace{0.1cm}  So ``world!'' becomes ``eptaf!''
 
\bigskip 
\noindent
Therefore, the message in the keyboard cipher is: ``Jraap eptaf!''

\bigskip 
\noindent
Here is another message. Encode the message in the keyboard cipher, word by word and letter by letter. On the last line, write the words ``Keyboard cipher:'' followed by the encoded message:

\noindent
Original message: ``But this time, there may also be another reason.''

\subsection{Keyboard cipher: Correspondences provided}\label{app:promptkeyboardcorrespondences}

The keyboard cipher is a cipher where each letter is replaced with the letter to the right of it on a QWERTY keyboard. If the letter does not have a letter to its right, then you should wrap around to the start of the row; e.g., ``p'' maps to ``q''. For reference, here are the correct replacements for all letters:

\noindent
a -> s

\noindent
b -> n

\noindent
c -> v

\noindent
d -> f

\noindent
e -> r

\noindent
f -> g

\noindent
g -> h

\noindent
h -> j

\noindent
i -> o

\noindent
j -> k

\noindent
k -> l

\noindent
l -> a

\noindent
m -> z

\noindent
n -> m

\noindent
o -> p

\noindent
p -> q

\noindent
q -> w

\noindent
r -> t

\noindent
s -> d

\noindent
t -> y

\noindent
u -> i

\noindent
v -> b

\noindent
w -> e

\noindent
x -> c

\noindent
y -> u

\noindent
z -> x

\bigskip
\noindent
For example, here is a message to be encoded:

\noindent
Original message: ``Hello world!''

\bigskip
\noindent
To encode this message, we replace each letter with the one to the right of it on a QWERTY keyboard:

\noindent
1. The first word is ``Hello''

\noindent
\hspace{0.1cm} * H -> J
 
\noindent
\hspace{0.1cm}  * e -> r
 
\noindent
\hspace{0.1cm}  * l -> a
 
\noindent
\hspace{0.1cm}  * l -> a
 
\noindent
\hspace{0.1cm}  * o -> p
 
\noindent
\hspace{0.1cm}  So ``Hello'' becomes ``Jraap''
 
\noindent
2. The second word is ``world!''

\noindent
\hspace{0.1cm}  * w -> e
 
\noindent
\hspace{0.1cm}  * o -> p
 
\noindent
\hspace{0.1cm}  * r -> t
 
\noindent
\hspace{0.1cm}  * l -> a
 
\noindent
\hspace{0.1cm}  * d -> f
 
\noindent
\hspace{0.1cm}  * ! -> !
 
\noindent
\hspace{0.1cm}  So ``world!'' becomes ``eptaf!''
 
\bigskip 
\noindent
Therefore, the message in the keyboard cipher is: ``Jraap eptaf!''

\bigskip 
\noindent
Here is another message. Encode the message in the keyboard cipher, word by word and letter by letter. On the last line, write the words ``Keyboard cipher:'' followed by the encoded message:

\noindent
Original message: ``But this time, there may also be another reason.''

\subsection{Birthdays}

On what date was Michelle Pfeiffer born? You should format your answer as ``Month Day, Year''; e.g., ``July 19, 1743.''

\section{Evaluation details}

\subsection{Shift ciphers and reversal}
We ignored quotation marks at the start and end of the answer, but otherwise the model's output had to exactly match the ground truth in order to be counted as correct.

\subsection{Pig Latin}
We ignored quotation marks at the start and end of the answer. 
When the task was decryption, we accepted any answer that was a possible decryption of the input, even if it was different from the intended one. The reason to make this allowance is that Pig Latin decryption can be ambiguous; e.g., \textit{identstray} could correspond to \textit{tridents} or \textit{strident}. We also ignored case; since we lowercased the text in the input, we decided not to penalize models for adding capitalization since inserting proper capitalization could reasonably be assumed to be part of the intended decoding process.

\subsection{Acronyms}

We ignored quotation marks, spaces, and capitalization in the answers. Models often placed their answer within a larger sentence (e.g., \textit{The sequence of letters created when combining the first letters of the words in the given sequence is ``CONCORD''.}); thus, we manually inspected answers containing spaces to extract just the answer, in order to avoid penalizing models for including this extra material. 

\subsection{Sorting}

We ignored quotation marks in the answers. The model's output list had to exactly match the correct output to be counted as correct. Sometimes the model placed the list within a larger sentence (e.g., \textit{The list of numbers in descending order is...}); in such cases, we extracted the list from the larger sentence.

\subsection{Counting, linear functions, and three-digit multiplication}

A model was considered correct if it got exactly the right number, or incorrect otherwise. We allowed models to use any wording that communicated the correct number (e.g., just saying the number, vs. producing it within a longer sentence; or writing large numbers with or without commas); we allowed for such cases by manually checking any answers that could not be parsed as a number.

\subsection{Birthdays}

For each model output, we first tried to automatically parse it into a date. If we could not parse it automatically, we checked it manually and converted it into a date if the answer mentioned a date. To be counted as correct, a model had to mention exactly the correct date (day, month, and year).

\section{Statistical tests}

\subsection{General notes}\label{app:stats_general}

In our comparisons across conditions (e.g., across output probability levels), we aimed to control our datasets for other variables (e.g., input probability). However, it was not always possible to exactly match these other variables. Therefore, here we run logistic and linear regressions that include predictors for our target variable as well as potential confounding variables, as a way to test whether the target variables have a statistically significant effect even when other factors are taken into account.
All statistical tests were run in R \citep{rpackage2022} using the \texttt{glmnet} package \citep{friedman2010regularization,tay2023elastic}. Plots were generated using \texttt{ggplot2} \citep{wickham2016ggplot2}. 
In some cases, various predictors of interest are correlated, so we report variance inflation factors (VIFs) as a measure of the extent to which this correlation is problematic. We deem the VIF to be excessive if it is above a threshold of 5. 
In all cases, we z-scored continuous or integer predictors due to large differences in scale, and we ran separate statistical tests for the two LLMs (GPT-3.5 and GPT-4). Unless otherwise stated, the response variable was a binary variable that was 1 if the GPT model produced the correct answer or 0 otherwise.

In some of the logistic regressions, the response variable was 0 for all examples in a particular condition, which creates issues for logistic regression due to complete separation. To address this issue, we  augmented the logistic regression with the weak Bayesian prior proposed by \citet{gelman2008weakly}, implemented in the R package \texttt{arm}.\footnote{\url{https://cran.r-project.org/package=arm}} The cases where we used this Bayesian augmentation are noted below.

\subsection{Shift ciphers}

\subsubsection{Comparing task variants}\label{app:shiftsig2vs13}

To compare rot-13 and rot-2, we used a logistic regression with predictors of shift (a binary variable that could either be a shift of 2 or 13), the number of characters in the sentence (which was the same for the input and the output), the number of subword tokens in the input, the number of subword tokens in the output, the log probability of the input (estimated using GPT-2), and the log probability of the output (estimated using GPT-2):

\begin{align}
    \text{correct}~\sim~& \text{shift} + \text{input\_nchars} + \text{input\_ntokens} + \text{output\_ntokens} \\ & + \text{input\_logprob} + \text{output\_logprob} 
\end{align}
In these tests, we used the weak Bayesian prior (Appendix~\ref{app:stats_general}) to address complete separation issues.

There was a statistically significant effect of shift for GPT-4 decoding ($p < 10^{-7}$), GPT-3.5 decoding ($p < 0.01$), and GPT-4 encoding ($p < 0.01$), but not GPT-3.5 encoding ($p = 0.29$), likely due to floor effects in that case (GPT-3.5's accuracy on rot-2 encoding was 0.00, and its accuracy on rot-13 encoding was 0.02).
In all cases, the shift predictor had a negligible VIF (less than 1.2) meaning that collinearity is not a concern.

\subsubsection{Input probability}\label{app:shift_input_prob_stats}

To investigate the effects of input probability, we used a logistic regression with predictors for input and output log probability (estimated using GPT-2), the number of subword tokens in the input and output, and the number of characters in the input (which was the same as the number of characters in the output): 
\begin{align}
    \text{correct} \sim \text{input\_logprob} + \text{output\_logprob} + \text{input\_ntokens} + \text{output\_ntokens} + \text{input\_nchars}  
\end{align}

\noindent
This test was applied to the cases where LLMs were asked to encode sentences of varying levels of probability.
The predictor of interest, input log probability, had a statistically significant effect for GPT-4 ($p < 0.05$) but not GPT-3.5 ($p = 0.61$). The VIF was small ($< 1.4$) for input log probability in both cases.

We then re-ran this analysis but with Levenshtein edit distance rather than correctness as the dependent variable; accordingly, we used a linear regression rather than a logistic regression. Input probability now had a statistically significant effect for both GPT-3.5 ($p < 0.01$) and GPT-4 ($p < 0.01$). For both models, the VIF for input log probability was less than 2.0.

\subsubsection{Output probability}\label{app:shiftsigdec}

To investigate the effects of output probability, we used the same variables as for input probability (Appendix~\ref{app:shift_input_prob_stats}).
Output probability had a highly statistically significant effect for both GPT-3.5 ($p < 10^{-4}$) and GPT-4 ($p < 10^{-7}$). The VIF for output log probability was small ($< 1.6$) in both cases.

\subsubsection{Step-by-step and chain-of-thought prompting}\label{app:shiftsigpromptstyle}

We applied the same statistical tests described above to the shift cipher results obtained with GPT-4 using step-by-step and chain-of-thought prompting. 
Output log probability had a significant effect in all three cases ($p < 10^{-7}$ for basic and step-by-step prompting; $p < 10^{-8}$ for chain-of-thought prompting). When having the response variable be a binary variable indicating whether the model was correct or not, input log probability had a significant effect for basic prompting ($p < 0.05$) and chain-of-thought prompting ($p < 0.05$), but not step-by-step prompting ($p = 0.059$). However, when we switched to the more sensitive metric of edit distance from the correct answer, we found a significant effect for both basic prompting ($p < 0.01$) and step-by-step prompting ($p < 0.05$), and it came close to significance for chain-of-thought prompting ($p = 0.066$). 
Finally, when comparing rot-13 to rot-2, there was a significant effect of the task for both encoding and decoding and for all prompt styles, except for the case of GPT-4 with chain-of-thought prompting. For encoding, we had $p < 0.01$ for basic prompting, $p < 0.01$ for step-by-step prompting, and $p < 0.05$ for chain-of-thought prompting. For decoding, we had $p < 10^{-7}$ for basic prompting, $p < 10^{-4}$ for step-by-step prompting, and $p = 0.22$ for chain-of-thought prompting.
Although the difference between rot-13 and rot-2 was no longer significant with chain-of-thought prompting, note that there are still substantial accuracy differences across shift levels in the chain-of-thought setting (see Figure~\ref{fig:cotstep}, top). Rot-13 vs.\ rot-2 was just one comparison that we initially chose to represent the broader set of comparisons that could be performed. Other comparisons retain a large difference; e.g., if we compare rot-13 and rot-12 in the same way as we compared rot-13 and rot-2, there is  a significant effect of shift level for GPT-4 with chain-of-thought prompting ($p < 10^{-4}$). For all the cases discussed in this paragraph, the VIF was small (less than 2.0).

\subsection{Reversal}

\subsubsection{Input probability}\label{app:revsigenc}

We used a logistic regression with the same variables as used for shift ciphers (Appendix~\ref{app:shift_input_prob_stats}. Input probability did not have a statistically significant effect for GPT-3.5 ($p = 0.22$) or GPT-4 ($p = 0.80$). The VIF for input log probability was modest for both models (1.82 for GPT-3.5 and 2.01 for GPT-4), meaning that collinearity is not a serious concern for these results.

\subsubsection{Output probability}\label{app:revsigdec}

We used a logistic regression with the same variables as used for shift ciphers (Appendix~\ref{app:shift_input_prob_stats}. Output probability had a statistically significant effect for both GPT-3.5 ($p < 10^{-8}$) and GPT-4 ($p < 10^{-11}$). For GPT-3.5, the VIF for output log probability was modest (1.68), meaning that collinearity is not a major concern. 
The VIF for GPT-3.5 was 1.76, and for GPT-4 it was 3.33, meaning that both were within the range we considered acceptable.
We therefore conclude that collinearity is unlikely to be a serious problem for drawing conclusions from these tests.

\subsection{Pig Latin}

\subsubsection{Comparing task variants}\label{app:pigsigtask}

To compare Pig Latin and Boar Etruscan, we used a logistic regression with predictors of task (a binary variable that could either be Pig Latin or Boar Etruscan), the number of characters in the input, the number of characters in the output, the number of subword tokens in the input, the number of subword tokens in the output, the log probability of the input (estimated using GPT-2), and the log probability of the output (estimated using GPT-2):

\begin{align}
    \text{correct}~\sim~&\text{task} + \text{input\_nchars} + 
    \text{output\_nchars} +
    \text{input\_ntokens} + \text{output\_ntokens} \\ &+ \text{input\_logprob} + \text{output\_logprob} 
\end{align}
In these tests, we used the weak Bayesian prior (Appendix~\ref{app:stats_general}) to address complete separation issues.

There was a statistically significant effect of task for GPT-3.5 encoding ($p < 0.01$), GPT-4 encoding ($p < 0.001$), and GPT-4 decoding ($p < 0.05$), but not GPT-3.5 decoding ($p = 0.08$).
In all cases, the shift predictor had a small VIF (less than 1.3) meaning that collinearity is not a concern.

To compare different variants of Pig Latin, we used a similar logistic regression as the one used for Pig Latin vs.\ Boar Etruscan, except instead of a binary categorical variable for task we used a continuous variable of task probability (which we defined as equal to the proportion of Pig Latin instances in C4 that used the Pig Latin variant in question). We did not use the Bayesian prior in this case because there was no complete separation.
For encoding text into Pig Latin, we found a statistically significant effect of task for both GPT-3.5 ($p < 10^{-6}$) and GPT-4 ($p < 10^{-8}$). For decoding text from Pig Latin, neither model showed a significant effect of task ($p = 0.27$ for GPT-3.5, and $p = 0.11$ for GPT-4). In all cases, the VIF for the task probability variable was small (less than 1.2), meaning that collinearity was not a concern for these regressions.

\subsubsection{Input probability}\label{app:pigsigenc}

We used a logistic regression with the same variables as used for shift ciphers (Appendix~\ref{app:shift_input_prob_stats}) except with a predictor added for the number of characters in the output (output\_nchars) because for Pig Latin (unlike for shift ciphers) this quantity differs from the number of characters in the input. Input probability did not have a statistically significant effect for GPT-3.5 ($p = 0.70$) or GPT-4 ($p = 0.88$). The VIF for input log probability was modest for both models (1.88 for GPT-3.5 and 1.98 for GPT-4), meaning that collinearity is not a serious concern for these results.

\subsubsection{Output probability}\label{app:pigsigdec}

We used a logistic regression with the same variables as used for investigating input probability (Appendix~\ref{app:pigsigenc}). Output probability had a statistically significant effect for both GPT-3.5 ($p < 10^{-11}$) and GPT-4 ($p < 10^{-14}$). The VIF for output log probability was reasonably small for both models (2.14 for GPT-3.5 and 2.69 for GPT-4), meaning that collinearity is not a major concern for these results.

\subsection{Article swapping}\label{app:articleswap_significance}

\subsubsection{Main experiment}\label{app:articleswap_basic_stats}

We first analyzed the effect of example probability using a logistic regression with predictors for input and output log probability (estimated using GPT-2), the number of subword tokens in the input (which was the same as the number in the output), and the number of characters in the input (which was the same as the number in the output):
\begin{align}
    \text{correct} \sim \text{task} + \text{input\_nchars} + 
    \text{input\_ntokens} +  \text{input\_logprob} + \text{output\_logprob} 
\end{align}
However, with this model, input log probability and output log probability had excessives VIFs of over 10 for both GPT-3.5 and GPT-4. We therefore reran the test without input log probability, leaving just output log probability to stand for the overall example probability. In this case, probability had a significant effect for both GPT-3.5 ($p < 10^{-11}$) and GPT-4 ($p < 10^{-15}$), and its VIF was reasonable (1.69 for GPT-3.5 and 2.22 for GPT-4). 

\subsubsection{Expanded experiment}\label{app:article_swap_expanded_stats}

The tests run in the previous subsection were unable to determine whether the effect of probability was due to the probability of the input or the probability of the output. We ran another serious of experiments to attempt to disentangle these effects. In the article swapping task that we have focused on, the input is an ungrammatical sentence that becomes grammatical when each article is swapped with the word before it. We now consider three additional framings of this task (based on which direction the swap occurs in and whether it is the input or output that is ungrammatical), to give four total task versions, illustrated in Table~\ref{tab:detailedswap}. 
These variants yield widely varying accuracies.
The direction of the swap (next vs.\ previous) does not appear to be the main driver of the differences in accuracy because, for both models, the ``previous'' direction yields both the highest-accuracy setting and the lowest-accuracy setting. 
The input also does not appear to be the main driver of accuracy differences: if it were, we would expect the highest accuracies in the conditions with a grammatical input, but in fact those conditions get the lowest accuracies.
What does appear to matter most is the output: the conditions with the grammatical outputs yield the highest accuracies for both models.
To be sure, this is not the only factor influencing the results: it appears the the ``ungramm.-P'' sentences are harder to process than the ``ungramm.-N'' ones, since for both encoding and decoding the condition with ``ungramm.-P'' has lower accuracies than the one with ``ungramm.-N.'' However, given that the most extreme differences in accuracy seem to be driven by the type of output that is produced, we conclude that output probability (rather than input probability) is mainly responsible for the highly significant effect of probability that we observed in Appendix~\ref{app:articleswap_basic_stats}.

\begin{table}[t]
    \centering
    \resizebox{\textwidth}{!}{
    \begin{tabular}{p{3.2cm}p{3.5cm}ccccc} \toprule
        & & & & & Accuracy: & Accuracy: \\
        Task version & Example & Input & Output & Direction & GPT-3.5 & GPT-4 \\ \midrule
        
        Encoding:\newline Swap with next & \raggedright I will not give up even for a moment. $\rightarrow$ I will not give up even for moment a. & Gramm. & Ungramm.-N & Next & 0.23 & 0.57 \vspace{3mm}\\ 

        Encoding:\newline Swap with previous & \raggedright I will not give up even for a moment. $\rightarrow$ I will not give up even a for moment. & Gramm. & Ungramm.-P & Previous & 0.01 & 0.05 \vspace{3mm}\\

        Decoding:\newline Swap with next & \raggedright I will not give up even a for moment. $\rightarrow$ I will not give up even for a moment. & Ungramm.-P & Gramm. & Next & 0.32 & 0.62 \vspace{3mm}\\
        
        Decoding:\newline Swap with previous & \raggedright I will not give up even for moment a. $\rightarrow$ I will not give up even for a moment. & Ungramm.-N & Gramm. & Previous & 0.65 & 0.83 \vspace{3mm}\\ \bottomrule

    \end{tabular}
    }
    \caption{Four variants of the article swapping task. \textit{Encoding} refers to a variant that converts a grammatical sentence to an ungrammatical one; \textit{decoding} is the reverse. \textit{Gramm.} and \textit{ungramm.} stand for \textit{grammatical} and \textit{ungrammatical}. In more detail, \textit{ungramm.-N} refers to an ungrammatical sentence created by starting with a grammatical sentence and swapping each article with the next word; \textit{ungramm.-P} is analogous but uses swapping with the previous word. \textit{Direction} refers to whether the task involves swapping articles with the next or previous word. The accuracies shown here are all based on the high-probability set of grammatical sentences.}
    \label{tab:detailedswap}
\end{table}

\subsection{Acronyms}\label{app:acronym_significance}

\subsubsection{Comparing task variants}\label{app:acronymsig1vs2}

We evaluated the effect of input probability using the logistic regression below. The task variable was a binary categorical variable that could either be the first-letter acronym task or the second-letter acronym task. We did not include predictors for example length because all examples had the same length, whether counted in characters or in subword tokens.
\begin{align}
    \text{correct} \sim \text{task} + \text{input\_logprob} + \text{output\_logprob}
\end{align}
The log probability of an individual word was as described in Appendix~\ref{app:acronym_data} (based on GPT-2's probability of the word in the context \textit{The word is ``WORD''}). Since the inputs were multiple words, their log probability was taken to be the sum of the log probabilities of the words in them. In these tests, we used the weak Bayesian prior (Appendix~\ref{app:stats_general}) to address complete separation issues.

The factor of interest was task, which had a significant effect for both GPT-3.5 ($p < 10^{-4}$) and GPT-4 ($p < 10^{-15}$). In both cases, the VIF for this factor was negligible (less than 1.02), showing that collinearity was not a concern.

\subsubsection{Input probability}\label{app:acronyminpsig}

We evaluated the effect of input probability using the logistic regression below:
\begin{align}
    \text{correct} \sim \text{input\_logprob} + \text{output\_logprob}
\end{align}
The factor of interest was input log probability, which had a significant effect for GPT-3.5 ($p < 10^{-5}$) but was not significant for GPT-4 ($p = 0.092$). In both cases, the VIF was negligible (less than 1.01), showing that collinearity was not a concern.

\subsubsection{Output probability}\label{app:acronymoutpsig}
To investigate the effect of output probability, we used the same logistic regression setup as for input probability (Appendix~\ref{app:acronyminpsig}).
Output probability had a highly significant effect for both GPT-3.5 ($p < 10^{-15}$) and GPT-4 ($p < 10^{-9}$). The VIF was negligible (less than 1.01) in both cases, so collinearity was not a concern.

\subsection{Counting}\label{app:counting_significance}

To investigate the effect of examples probability on counting success, we used logistic regressions with predictors for the magnitude of the number being counted to, the log probability of the input sequence, and the log probability of the output (i.e., the count):
\begin{align}
    \text{correct} \sim \text{magnitude} + \text{input\_logprob} + \text{output\_logprob} 
\end{align}

\noindent
For counting words, the input log probability was the mean of word log probabilities, where a word's log probability is the log probability assigned by GPT-2 to the sequence \textit{The word is ``WORD''} minus the log probability for \textit{The word is ``}. For characters, we counted how often each character appeared in the first 1 billion words of the C4 corpus \citep{raffel2020exploring}, and added 1 to each count (add-1 smoothing) to deal with the nonappearance of some emojis. The output log probability was estimated by counting how many times each number appeared in the first 1 million integers between 1 and 100 mentioned in the C4 corpus, then normalizing.

When the inputs were all common and the outputs could be any number from 1 to 100, output log probability had a significant effect in all cases (GPT-3.5 or GPT-4 processing characters or words; $p < 10^{-15}$ in all cases). Magnitude had a significant effect for both character-based conditions ($p < 10^{-15}$ for GPT-3.5; $p < 10^{-14}$ for GPT-4) but not for either word-based condition ($p = 0.56$ for GPT-3.5; $p = 0.07$ for GPT-4).  For all four conditions, the VIFs for both magnitude and output log probability are below 2.0, so we conclude that collinearity is not a concern.

We ran the same tests on the case where we performed a controlled comparison between high-probability counts and low-probability counts. Output log probability was also significant here ($p < 10^{-15}$ for all 4 cases), and the VIFs continued to be reasonable.

When we compared the counting of low-probability inputs with high-probability ones, we found that input log probability had a statistically significant effect for both models when counting words ($p < 0.05$ for both GPT-3.5 and GPT-4) and for counting characters ($p < 10^{-10}$ for GPT-3.5; $p < 10^{-6}$ for GPT-4). Note that, for GPT-3.5 counting words and for GPT-4 counting characters, the effect of input probability was in the opposite direction from what we expected: they scored higher when counting rare words than common ones. However, the effect size in these cases was small (Figure~\ref{fig:counting_inp}). The VIF for input log probability was below our threshold of 5 for all conditions.

\subsection{Sorting}\label{app:sortingsig}

To compare different sorting directions (alphabetical vs.\ reverse alphabetical, or ascending vs.\ descending) we used logistic regressions with predictors for the direction, the number of characters (which was the same for the input and output), the number of subword tokens (which was the same for the input and output), the input log probability, and the output log probability:
\begin{align}
    \text{correct} \sim \text{direction} + \text{nchars} + \text{ntokens} + \text{input\_logprob} + \text{output\_logprob}
\end{align}

\noindent
When we were investigating the sorting of words, the direction predictor was a binary variable that could take values of \textit{alphabetical} or \textit{reverse alphabetical}. 
When we were investigating the sorting of numbers, the direction predictor was a binary variable that could take values of \textit{ascending} or \textit{descending}. 
The input and output log probabilities were both estimated as the probability assigned by GPT-2 to the word sequence.

These regressions found that, for sorting words, direction had a significant effect for both GPT-3.5 ($p < 10^{-8}$) and GPT-4 ($p < 10^{-6}$). However, for sorting numbers, direction did not have a significant effect (for GPT-3.5, $p = 0.70$; for GPT-4, $p = 0.96$). The VIF for the task predictor was below our threshold of 5.0 for all cases, so we conclude that collinearity is not a major concern. 

\subsection{Linear functions}

\subsubsection{Comparing task variants}\label{app:linearsigtask}

To compare the two linear functions, we used a logistic regression with predictors for the task (a binary variable that could be the task $(9/5)x + 32$ or the task $(7/5)x + 31$), the continuous input value, and the continuous output value:
\begin{align}
    \text{correct} \sim \text{task} + \text{input} + \text{output}
\end{align}
In these tests, we used the weak Bayesian prior (Appendix~\ref{app:stats_general}) to address complete separation issues.
The factor of interest was task, which had a significant effect for both GPT-3.5 ($p < 0.001$) and GPT-4 ($p < 0.01$). In both cases, the VIF was negligible (less than 1.1), showing that collinearity was not a concern.

We repeated these tests for the setting in which the output (rather than the input) was controlled across tasks. The variable of task continued to have a significant effect for both GPT-3.5 ($p < 10^{-7}$) and GPT-4 ($p < 0.001$), and the VIF was still small for both tasks (less than 1.1).

\subsubsection{Comparing prompts}\label{app:linearpromptsig}

To compare the different prompts, we used a mixed-effects logistic regression with a fixed effect for prompt type (a 3-way contrast, sum coded) and a random intercept for item:
\begin{align}
    \text{correct} \sim \text{prompt} + \text{(1|item)}
\end{align}
We did not include effects for input or output since the by-item random intercept subsumes these factors.

For each LLM, we first tested whether there was a significant effect of prompt type by using a chi-square test to compare the full model described above with a null model that removed the fixed effect of prompt. These tests found a significant effect of prompt for both LLMs ($p < 10^{-12}$ for GPT-3.5; $p < 10^{-18}$ for GPT-4).
Since the chi-square test found a significant effect of prompt, we then tested which pairwise comparisons of prompt types were significant by applying the R package \texttt{multcomp} \citep{hothorn2008multcomp}, which controls for multiple comparisons, to the full model that had a fixed effect for prompt method. We found that, for both models, all three comparisons were significant, with $p < 0.01$ in all cases.

We repeated these tests on the setting where the values were constrained to be unusual ones (i.e., outside the standard range of temperatures, and not divisible by 10). The chi-square tests again showed a significant effect of prompt (with $p < 10^{-18}$ for both models), and again all pairwise comparisons were significant, with $p < 0.01$.

\subsection{Three-digit multiplication}\label{app:multsig}

To compare the four different ways of formatting the numbers in the prompt, we used a mixed-effects logistic regression with a fixed effect for the prompt format (a 4-way contrast, sum-coded) and a random intercept for item:
\begin{align}
    \text{correct} \sim \text{format} + \text{(1|item)}
\end{align}
The procedure for assessing the significance of differences between prompt formats was the same as the procedure for comparing linear function prompts (Appendix~\ref{app:linearpromptsig}). For each LLM, we first tested if the format had a significant effect by running a chi-square test that compared the full model to a null model that removed the fixed effect of format. For both LLMs, the chi-square test found a significant effect for format ($p<10^{-15}$ for GPT-3.5; $p<10^{-8}$ for GPT-4). Since the chi-square test found that the format had a significant effect, we tested which pairs of formats were significantly different from each other using the \texttt{multcomp} package. Many of the differences had a significant effect; see Table~\ref{tab:format_comps}.

\begin{table}[]
    \centering
    \begin{tabular}{ccc} \toprule
         & GPT-3.5 & GPT-4 \\ \midrule
       Number vs.\ lowercase  & *** & n.s. \\
       Number vs.\ all caps & *** & n.s. \\
       Number vs.\ alternating capitals & *** & *** \\
       Lowercase vs.\ all caps & n.s. & n.s. \\
       Lowercase vs.\ alternating capitals & * & *** \\
       All caps vs.\ alternating capitals & * & *** \\ \bottomrule
    \end{tabular}
    \caption{Statistical significance of the differences between LLM accuracy for three-digit multiplication when using different prompt formats. Key: n.s. = not significant; * = $p < 0.05$; ** = $p < 0.01$; *** = $p < 0.001$.}
    \label{tab:format_comps}
\end{table}

\subsection{Birthdays}\label{app:birthdaysig}

To investigate the effect of the frequency of a person being mentioned on model accuracy at naming the person's birthday, we ran logistic regressions with predictors for input log probability and output log probability:
\begin{align}
    \text{correct} \sim \text{input\_logprob} + \text{output\_logprob}
\end{align}

\noindent
The input probability (i.e., the probability of the name being asked about) was proportional to the number of occurrences of the name in the C4 corpus. The output probability (i.e., the probability of the date that was the answer) was proportional to the number of entries in the WikiBio dataset that were born on each date. For both models, input log probability had a highly statistically significant effect, with $p < 10^{-15}$. In both cases, the VIF of input log probability was negligible (less than 1.01).
 
\section{Estimating task probability}

To estimate the probability of task variants, we performed corpus analyses of the C4 corpus \citep{raffel2020exploring}, specifically the version hosted by Hugging Face Datasets \citep{lhoest2021datasets}.

It is challenging to precisely count how many occurrences of a task there are in a naturalistic corpus because task instances are typically not labeled as such. As a proxy for counting task instances, we counted phrases that are often used to invoke each task. Below we discuss precisely how we carried this out for each type of task, including discussion of the assumptions that underlie each investigation. The reliance on these (certainly imperfect) assumptions is a limitation of these analyses; the other main limitation is that, in many cases, we had to use manual analysis of data to ensure accuracy, but due to the large amount of data we did not scrutinize all examples carefully, meaning that some examples may have been missed or misclassified. Though these analyses are certainly imperfect, they are useful for giving a rough sense of how frequent each task variant is. 

\subsection{Shift ciphers}\label{app:shiftcounting}

We extracted all examples within the corpus that contained any of the following types of strings. The search was done in a case-insensitive way. These string templates are intended to capture the most frequent ways of referring to shift ciphers:

\begin{itemize}
    \item \textbf{shift cipher}
    \item \textbf{Caesar cipher} (another common term for a shift cipher)
    \item \textbf{rotN} (for all values of N from 1 to 25)
    \item \textbf{rot-N} (for all values of N from 1 to 25)
    \item \textbf{the next letter in the alphabet} (often used when discussing rot-1)
    \item \textbf{the previous letter in the alphabet} (often used when discussing rot-25)
\end{itemize}
We then manually checked all examples returned by this search. Many of the examples were not in fact discussing a shift cipher. For instance, there is a gene called ROT2 which should not be counted as a shift cipher; there is a scheme for encoding numbers called rot-5, which we also did not count because it is different from the text encoding scheme rot-5; and the phrase \textit{the next/previous letter in the alphabet} can be used in non-cipher-related contexts. We excluded such examples to give us only the examples that discuss shift ciphers. 
We then categorized which shift level was being discussed in each case (some examples were excluded because they did not mention a specific shift level). If a single document included several instances of the above phrases, we only counted that document once. In order to count an example as an instance of a shift, we did not require there to be examples of encoded text because checking for such examples would have further lengthened an already labor-intensive search. The resulting counts are shown in the top of Figure~\ref{fig:shifts}.

Instead of searching for phrases used to talk about shift ciphers, we considered searching for text encoded in shift ciphers by searching for the encoded forms of various common words. We attempted this approach but found that it ran into unexpected challenges. Words sometimes become other words when shifted, yielding far more false positives than we could manually check; yet we also could not simply exclude shifted candidates that become other words because often the false positives were strings that would not typically be recognized as a word; e.g., \textit{stop} gave us the false positive \textit{rsno}, which would be hard to exclude because it does not typically show up on word lists, yet which does occur in C4 as an abbreviation for the Royal Scottish National Orchestra.

The assumptions underlying this investigation include:
\begin{itemize}
    \item The distribution of shifts in C4 is similar to the distribution of shifts in the training data for GPT-3.5 and GPT-4.
    \item The texts returned by the key phrases described above are representative of the full set of texts mentioning shift ciphers.
    \item A shift's probability of being illustrated is proportional to its probability of being mentioned (this assumption relates to the fact that we did not require an example to be included, whereas models are most likely to only learn about a shift from cases with examples).
\end{itemize}

\subsection{Pig Latin}\label{sec:pigcounting}

We extracted all examples from the corpus that, when lowercased, contained the string \textit{pig latin}. We then manually examined each occurrence to categorize which variant of Pig Latin was used. We only counted examples which specified a variant, either by verbally describing the rule or by providing examples that made the rule clear. As with shift ciphers, we did not require the text to include a translated example to be counted. This search resulted in a count of 32 instances of the variant using \textit{-way}, 22 instances of the variant using \textit{-ay}, 12 instances of the variant using \textit{-yay}, and 9 instances of the variant using \textit{-hay}. 

This sample size is relatively small. However, we conducted a similar analysis on the Pile dataset \citep{gao2020pile} and got similar results: 68 instances of \textit{-way}, 46 instances of \textit{-ay}, 26 instances of \textit{-yay}, and 8 instances of \textit{-hay}. Thus, the rough frequencies of each variant seem reasonably reliable. Our results are also broadly similar to the counts found in the survey of \citet{vaux2003underdetermination}, with the major exception that they had many more respondents use \textit{-ay} than \textit{-way}; their population of respondents may have different Pig Latin preferences due to the difference in medium (online text vs.\ survey response) or due to time (their survey was collected in 2003, whereas most text in C4 and the Pile is more recent).

The following assumptions underlie this investigation:
\begin{itemize}
    \item The distribution of variants in C4 and the Pile is similar to the distribution of shifts in the training data for GPT-3.5 and GPT-4.
    \item The texts returned by searching for the phrase \textit{pig latin} are representative of the full set of texts discussing Pig Latin.
    \item A variant's probability of being illustrated is proportional to its probability of being mentioned.
\end{itemize}

\subsection{Sorting}\label{app:sortingfreq}

To estimate the relative frequency of alphabetical order and reverse alphabetical order in the C4 corpus, we counted the occurrences of the phrases \textit{in alphabetical order} and \textit{in reverse alphabetical order}. They occurred 95,492 times and 629 times, respectively, from which we estimate that alphabetical order is approximately 150 times as common as reverse alphabetical order. 
Similarly, to compare ascending order and descending order, we counted the occurrences of the phrases \textit{in ascending order}---which occurred 21,562 times---and \textit{in descending order}---which occurred 31,378 times. We thus estimate that descending order is approximately 1.5 times as common as ascending order.

These analyses assume that the frequency of these phrases is proportional to the frequency of each type of ordering. This assumption is likely imperfect; in particular, it might be that alphabetical order and ascending order are the default ways of ordering words and numbers, such that they are sometimes used without any explicit mention of the type of ordering that occurs. Nonetheless, these analyses provide one piece of evidence that alphabetical order is much more common than reverse alphabetical order in Internet text, but that ascending and descending order do not have as large of a difference in usage frequency.

\section{Spelling}\label{app:spelling}

To evaluate whether models have access to the spelling of their tokens, we asked them to spell 1,000 single-token words. We first produced a list of all words that appeared among the first 1 billion words in the Pile dataset that fulfilled two conditions. First, the word had to be a single token in the tokenization scheme used by GPT-3.5 and GPT-4, both when the word appeared after a space and when it did not. Second, the word had to consist entirely of lowercase letters from the Roman alphabet. We then randomly selected 1,000 words from this list of candidates. Models were asked to spell the word with a prompt of the following form:

\ex. Spell the word ``additional'' by listing its letters, in order, separated by spaces.

The inclusion of spaces in the output forces the model to break the token down into individual letters.

GPT-4 scored 99.9\% on this test. Its only error was producing \textit{b e c k e n d} when the correct answer was \textit{b a c k e n d}. When evaluated strictly, GPT-3.5 scored 98.1\%; of its 19 errors, 17 took the form of using hyphens rather than spaces (e.g., responding \textit{i-n-d-u-s-t-r-y} when the target response was \textit{i n d u s t r y}). We view these errors as orthogonal to our question of whether models encode the spelling of words---they reveal failure at following instructions rather than failure at spelling---so we counted GPT-3.5 as correct in these cases, yielding an accuracy of 99.8\%.

\end{document}